\documentclass[letterpaper]{article} 
\usepackage{aaai2026}  
\usepackage{times}  
\usepackage{helvet}  
\usepackage{courier}  
\usepackage[hyphens]{url}  
\usepackage{graphicx} 
\usepackage{natbib}  
\usepackage{caption} 
\usepackage{algorithm}
\usepackage{algorithmic}
\usepackage{xcolor}

\usepackage{newfloat}
\usepackage{listings}
\DeclareCaptionStyle{ruled}{labelfont=normalfont,labelsep=colon,strut=off} 
\floatstyle{ruled}
\newfloat{listing}{tb}{lst}{}
\floatname{listing}{Listing}
\usepackage{amsmath}
\usepackage{amsthm}
\usepackage{subcaption}
\usepackage{booktabs}
\usepackage{multirow}

\newcommand{\crossentropy}{\mathbb{C}}
\newtheorem{theorem}{Theorem}[section]

\newtheorem{lemma}[theorem]{Lemma}

\newtheorem{problem}[theorem]{Problem}
\newtheorem{definition}[theorem]{Definition}

\usepackage{amsmath,amsfonts,bm}

\def\eqref#1{equation~\ref{#1}}

\def\1{\bm{1}}

\def\vc{{\bm{c}}}

\def\vx{{\bm{x}}}

\def\vz{{\bm{z}}}

\DeclareMathAlphabet{\mathsfit}{\encodingdefault}{\sfdefault}{m}{sl}
\SetMathAlphabet{\mathsfit}{bold}{\encodingdefault}{\sfdefault}{bx}{n}

\def\gD{{\mathcal{D}}}
\def\gE{{\mathcal{E}}}

\newcommand{\E}{\mathbb{E}}

\newcommand{\R}{\mathbb{R}}

\usepackage{xspace}

\newcommand{\appendixBroaderImpact}{\appendixname~\ref{appx:sec:broader_impact}\xspace}
\newcommand{\appendixMvtec}{\appendixname~\ref{appx:sec:mvtec}\xspace}
\newcommand{\appendixProofs}{\appendixname~\ref{appx:proofs}\xspace}
\newcommand{\appendixLimitations}{\appendixname~\ref{appx:sec:limitations}\xspace}
\newcommand{\appendixConceptAcc}{\appendixname~\ref{sec:cf_disentanglement}\xspace}
\newcommand{\appendixMinEdits}{\appendixname~\ref{appx:sec:xad:min_edits}\xspace}

\newcommand{\appendixHyperparameterAnalysis}{\appendixname~\ref{appx:sec:hp_sensivity_analysis}\xspace}
\newcommand{\appendixADMethods}{\appendixname~\ref{appx:deep_ad_methods_details}\xspace}
\newcommand{\appendixDatasets}{\appendixname~\ref{appx:sec:datasets}\xspace}
\newcommand{\appendixMetrics}{\appendixname~\ref{appx:sec:metrics}\xspace}
\newcommand{\appendixHyperparameters}{\appendixname~\ref{appx:sec:hyperparameters}\xspace}

\newcommand{\appendixQuantitative}{\appendixname~\ref{appx:sec:full_quantitative_results}\xspace}
\newcommand{\appendixQualitative}{\appendixname~\ref{appx:sec:full_qualitative_results}\xspace}

\usepackage{adjustbox}
\usepackage{listings}
\usepackage{enumitem}
\usepackage[utf8]{inputenc} 
\usepackage[T1]{fontenc}

\title{Reimagining Anomalies: What if Anomalies Were Normal?}
\author {
    Philipp Liznerski\textsuperscript{\rm 1}\equalcontrib,
    Saurabh Varshneya\textsuperscript{\rm 1}\equalcontrib,
    Ece Calikus\textsuperscript{\rm 2},
    Puyu Wang\textsuperscript{\rm 1},
    Alexander Bartscher\textsuperscript{\rm 1},
    Sebastian Josef Vollmer\textsuperscript{\rm1,\rm3},
    Sophie Fellenz\textsuperscript{\rm 1},
    Marius Kloft\textsuperscript{\rm 1}
}
\affiliations {
    \textsuperscript{\rm 1}RPTU University Kaiserslautern-Landau, Germany \\
    \textsuperscript{\rm 2}Uppsala University, Sweden \\
    \textsuperscript{\rm 3}German Research Center for Artificial Intelligence (DFKI), Germany \\
}

\begin{document}

\maketitle

\begin{abstract}
Deep learning-based methods have achieved a breakthrough in image anomaly detection, but their complexity introduces a considerable challenge to understanding why an instance is predicted to be anomalous.
We introduce a novel explanation method that generates multiple alternative modifications for each anomaly, capturing diverse concepts of anomalousness.
Each modification is trained to be perceived as normal by the anomaly detector.
The method provides a semantic explanation of the mechanism that triggered the detector, allowing users to explore ``what-if scenarios.''
Qualitative and quantitative analyses across various image datasets demonstrate that applying this method to state-of-the-art detectors provides high-quality semantic explanations. 
\end{abstract}

\begin{links}
    \link{Code}{https://github.com/liznerski/counterfactual-xad}
    \link{AAAI}{https://aaai.org/aaai-publications} 
\end{links}

\section{Introduction}\label{sec:intro}
\begin{figure}[!ht]
    \centering
    \includegraphics[width=0.47\textwidth]{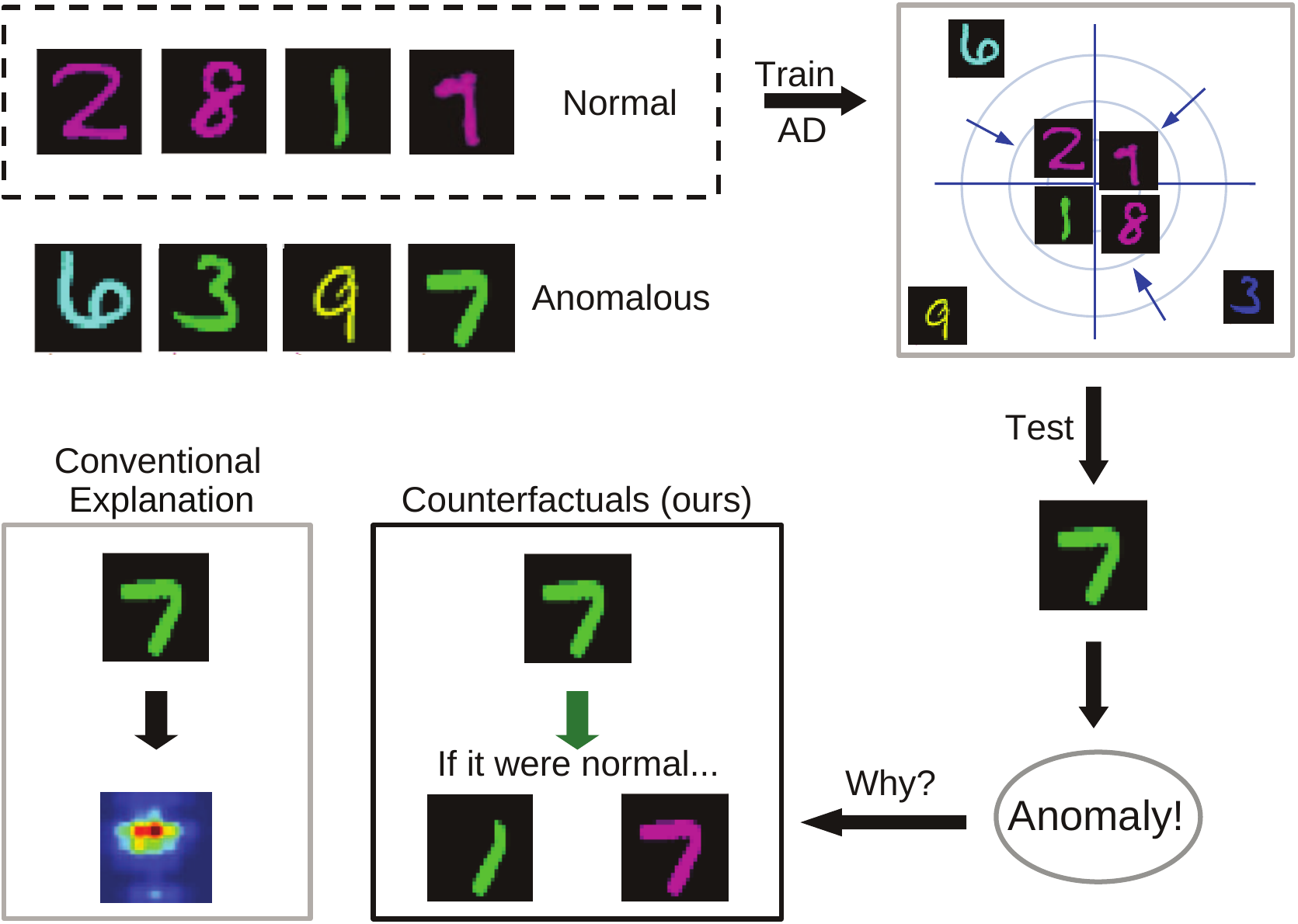}
    \caption{The figure illustrates the benefit of counterfactual explanation of anomaly detectors over traditional methods, using a dataset of handwritten digits in various colors. The normal data (top left) consist of pink digits and instances of the digit one in any color. An example anomaly---a green seven---is shown on the right. Conventional explanation methods localize the anomaly in the image and highlight it on a heatmap (bottom left). 
    In contrast, the proposed method transforms the anomaly into multiple counterfactuals, addressing the crucial question: ``How must the anomaly be altered to appear normal to the detector?"} 

    \label{fig:motivation}
\end{figure}

Anomaly detection identifies patterns that deviate from normal behavior, the so-called \emph{anomalies}. These anomalies can correspond to crucial actionable information in various domains such as medicine, manufacturing, and environmental monitoring \citep{chandola2009anomaly}.  

Recently, deep learning has shown tremendous success in anomaly detection (AD), reducing error rates to approximately $1\%$ in numerous image benchmarks \citep{reiss2021panda, ruff2021}.  
However, detectors based on deep learning lack the out-of-the-box interpretability of their traditional counterparts, making it difficult to understand the reasoning behind their predictions \citep{liznerski2021}. Their lack of transparency is particularly concerning in sectors where safety is crucial and in situations where building trust 
is essential \citep{samek2020toward}.  
Understanding modern anomaly detectors is a major challenge in contemporary AD and a necessary step before using AD in decision-making systems \citep{ruff2021}.

Although feature-attribution techniques such as anomaly heatmaps \citep{roth2022towards} have been explored, they do not explain the underlying semantics relevant to the decision-making of the detectors. 
In domains beyond AD, counterfactual explanation (CE) has emerged as a popular alternative. CE generates synthetic samples that change the model's prediction with minimal alterations to the original sample \citep{ghandeharioun_dissect_2021}. 

In this paper, we propose the use of CE to explain image anomaly detectors. 
While prevalent approaches identify anomalous regions within images, the presented technique generates a set of counterfactual examples for each anomaly, capturing diverse disentangled aspects (see Figure~\ref{fig:motivation}). 
Our goal is to explain why a detector flags an instance as anomalous. 
The framework provides semantic CEs that reveal when detectors rely on spurious or biased cues (\appendixBroaderImpact), e.g., labeling portraits as anomalous due to skin tone or background context. These insights support accountable and lawful deployment of AD systems.



\section{Related Work}

In the past decade, research has increased on improving the interpretability of neural networks. This increase is driven by the growing use of ML in decision-making systems, where transparency of predictions is crucial and even legally mandated in many countries \citep{neuwirth2022eu}.

\paragraph{Explanation of image AD.} 
Research in explainable image AD has primarily focused on feature attribution methods, pinpointing image areas that influence predictions. 
Feature attribution methods trace an importance score from the model output back to the pixels \citep{selvaraju2017grad,zhang2018top} or alter parts of the image and measure the impact on the model output \citep{fong2017interpretable, dhurandhar2018explanations}. Some of these approaches have been applied to AD \citep{liznerski2021,li2021cutpaste}. 
Several methods generate explanations using generative models or autoencoders, where the pixel-wise reconstruction error yields an anomaly heatmap \citep{bergmann2019mvtec, dehaene2020Iterative}. Others use fully convolutional architectures \citep{liznerski2021} or transfer learning \citep{defard2021padim, roth2022towards}. All these methods identify regions within an image that influence the detector's prediction; however, they do not explain the detectors at a higher semantic level \citep{adebayo2018sanity,varshneya2024interpretable}. 

\paragraph{Counterfactual explanation of neural networks on images.} 
CE methods \citep{guidotti2022counterfactual} identify the necessary changes in the input to alter the model prediction in a specific way. 
Such explanations can provide profound insights that enhance comprehension of model behavior and align more closely with human cognitive processes \citep{pearl2009causality}. Existing CE algorithms are designed primarily for supervised learning on tabular data \citep{wachter2017counterfactual,mothilal2020explaining,guidotti2022counterfactual}. A few studies have also explored the application of CE to image classification \citep{goyal_counterfactual_2019, ghandeharioun_dissect_2021,abid2022meaningfully,singla2023explaining}. DISSECT \citep{ghandeharioun_dissect_2021} is notable for its ability to generate multiple CEs with disentangled high-level concepts. 
Recent work started exploring CE for supervised image AD. 
Studies by \citet{sanchez2022healthy, wolleb2022diffusion, siddiqui2024vald, ahamed2024igconda, fontanella2024diffusion} utilize diffusion models guided by text prompts or learnable conditions to generate normal counterparts of abnormal medical images. 
These approaches rely on supervised learning, framing the AD problem as a classification task. They fine-tune pretrained diffusion models or use classifier-guidance, utilizing both normal and ground-truth anomalies. 
Although these approaches are promising for explaining model decisions with counterfactuals, they are applicable only to supervised settings, making them unsuitable for unsupervised AD models.

\paragraph{Counterfactual Explanations of AD.} 
Virtually all CE methods for AD have been applied to tabular data \citep{angiulli2023counterfactuals,datta2022framing,han2023achieving} and time series \citep{sulem_diverse_2022,cheng2022fine}. These methods use knowledge graphs or structural causal models to generate CEs for categorical features \citep{datta2022framing,han2023achieving} or take advantage of temporal aspects \citep{sulem_diverse_2022, cheng2022fine}. Some of these methods have been applied to fairness \citep{han2023achieving} and algorithmic recourse \citep{datta2022framing}. None of these methods is applicable to image data.
Very recently, \citet{ji2024arpro} introduced AR-Pro, a CE approach to explain anomalies in images. 
However, this method performs defect repair by modifying localized faulty regions with mask supervision, merely transforming defective regions into their normal versions. 
Our framework instead explains semantic anomalies without localization, identifying conceptual attributes (background, color, bias) that drive a detector’s decision.
As shown in \appendixMvtec, it also handles defect-type anomalies as a special case: on industrial data similar to AR-Pro it produces repair-like CEs but generalizes to multi-concept and semantic anomalies.

\section{Counterfactual Explanation of Image AD} \label{sec:method}
We formally present a novel framework for generating CEs of image AD. We first define the general setup and then explain how to use GANs and diffusion models to produce CEs. 
To the best of our knowledge, this approach is the first to explain semantic image AD using CE.

\subsection{What if the Anomaly were Normal?}
Our aim is to provide explanations for a given anomaly detector $\phi: \R^D \rightarrow [0,1]$  that maps an image $x \in \R^D$ to an anomaly score $\alpha \in [0,1]$.
We define a CE for the detector $\phi$ and perceived anomaly $\vx^* \in \R^{D}$ (i.e., $\phi(\vx^*) \gg 0$) as a modified sample $\bar{\vx}^*$ with $\phi(\bar{\vx}^*) \approx 0$ and $\lVert \bar{\vx}^* - \vx^* \rVert_1 \leq  \epsilon$ for an $\epsilon \geq 0$.
In other words, a CE must be normal according to $\phi$, while being minimally changed w.r.t.~the original anomaly $\vx^*$. Thus, CEs address the question: ``What if the anomaly $\vx$ were normal?'', explaining the behavior of the anomaly detector at a high semantic level. 

To produce such CEs for deep AD, we need to train a generator $G: \R^D \rightarrow \R^D$ to yield $G(\vx^*) = \bar{\vx}^*$. 
However, normal images can differ from anomalies in multiple ways, and thus multiple CEs may be required to adequately explain an anomaly. We want the generator to consider multiple categorical concepts $k \in \{1,\dots,K\}$. Thus, the generator is now of the form $G: \R^D \times \{1, \dots, K\} \rightarrow \R^D$ and is supposed to produce $G(\vx^*, k) = \bar{\vx}^*_k$ with $\lVert\bar{\vx}^*_k - \bar{\vx}^*_{k'} \rVert_1 \geq \epsilon'$ for any $k \neq k'$.

The same data $\lbrace (\vx_0, y_0), \dots, (\vx_n, y_n) \rbrace$ used to train $\phi$ can also train $G$. Note that in AD, training labels $y_i$ are typically unknown, and most samples are assumed normal.

\subsection{Deep Generative Models for CE of Image AD} \label{sec:ct_for_cf_on_ad}
In practice, it has been found beneficial to train the generator to produce sequences of CEs with increasing impact on a classifier's output \citep{ghandeharioun_dissect_2021}. The proposed approach is based on this idea. 
We modify the generator $G: \R^D \times [0, 1] \times \{1, \dots, K\} \rightarrow \R^D$ to consider a target score $\alpha$, aiming for the trained $G$ to produce a sample with an anomaly score of approximately $\alpha$. 

\begin{figure}[t]
    \centering
    \begin{subfigure}[b]{0.80\columnwidth}
        \centering
        \includegraphics[width=\textwidth]{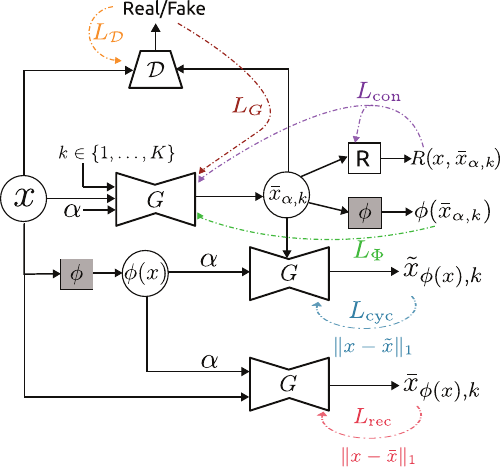}
        \caption{Training}
        \label{fig:concept_disentangle_training}
    \end{subfigure}
    \hspace{0.4em}
    \begin{subfigure}[b]{0.12\columnwidth}
        \centering
        \includegraphics[width=\textwidth]{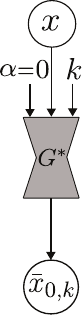}
        \vspace{3em}
        \caption{Test}
        \label{fig:concept_disentangle_inferece}
    \end{subfigure}
    \caption{Schematic overview of the proposed CE framework for explaining a black box image AD model $\phi$. (a) shows the training losses and their impact on the different models. (b) shows the inference, where the target anomaly score is zero. Gray nodes (i.e., the trained AD model $\phi$ and generator $G^*$) represent models that are not optimized. }
    \label{fig:concept_disentangle}
\end{figure}

\subsubsection{GANs for CE of image AD.}
Here, we first introduce a GAN-based \citep{goodfellow2020generative} model for producing CEs of image AD.
In particular, we train $G$ as a concept-disentangled GAN. We define a discriminator $\mathcal{D}: \R^D \rightarrow \R$ and a concept classifier $R: \R^D \times \R^D \rightarrow [0,1]^K$.
$\mathcal{D}$ is trained to distinguish between generated $\bar{\vx}_{\alpha, k} = G(\vx, \alpha, k)$ and true samples from the dataset, encouraging \emph{realistic} outcomes. $R$ classifies the concept $k$ for a sample $\bar{\vx}_{\alpha, k}$, encouraging the generated samples to be \emph{concept-disentangled} on a semantic level. Further losses encourage the generator to incur \emph{minimal changes} on the original sample $\vx$ and to yield target anomaly scores $\alpha$ (i.e., $\phi(\bar{\vx}_{\alpha, k}) \approx \alpha$). The proposed objective summarizes to
\begin{align} \label{eq:opt_system}
    &\min_{G, R} \max_{\mathcal{D}} \mathbb{E}_{\vx\sim P_X} \mathbb{E}_{\alpha, k} \big[ \lambda_{gan} \left(L_{\mathcal{D}}(\mathcal{D}) + L_{G}(G)\right) -  L_{\phi}(G) \nonumber  \\   
     &\lambda_\phi +  \lambda_{rec} \left(L_{rec}(G) +  L_{cyc}(G) \right) + \lambda_r L_{con}(G, R) \big], 
\end{align}
where $p_X$ denotes the training data distribution, and $\lambda_{gan}, \lambda_\phi, \lambda_{rec}$, and $\lambda_r$ are some constant factors. 
Figure~\ref{fig:concept_disentangle} shows a schematic overview of the framework. 

In the following, we will explain the various losses, starting with $L_{\phi}(G)$, which encourages for $\bar{\vx}_{\alpha, k}$ an anomaly score of $\alpha$. $L_{\phi}(G)$ can be any loss that measures the divergence of target anomaly scores $\alpha$ and true anomaly scores $\phi(\bar{\vx}_{\alpha, k})$; for example, the $L_2$ distance or the KL-divergence. We assume the detector $\phi$ to be bounded (w.l.o.g, $0 \leq \phi(\vx) \leq 1$). In our experiments, we found that a continuous binary cross entropy loss produces the best results: 
\begin{equation*}
L_{\phi}(G) = \alpha \log\big(\phi(\bar{\vx}_{\alpha, k})\big) + \big(1-\alpha\big) \log\big(1-\phi(\bar{\vx}_{\alpha, k})\big).
\end{equation*}

The losses $L_{\mathcal{D}}(\mathcal{D})$ and $L_{G}(G)$ can be any discriminative and generative GAN losses, respectively. We specifically experimented with spectral normalization \citep{miyato2018spectral} and the hinge loss \citep{miyato2018cgans}: 
\begin{equation*}
L_{G}(G) = - \min(0, -1 + \mathcal{D}(\bar{\vx}_{\alpha, k})),
\end{equation*}
\begin{equation*}
L_{\mathcal{D}}(\mathcal{D}) = \min(0, -1 + \mathcal{D}(\vx)) + \min(0, -1 -\mathcal{\mathcal{D}}(\bar{\vx}_{\alpha, k})).
\end{equation*}

The reconstruction loss makes $G$ recreate $\vx$ for every concept $k$, when conditioned on $\vx$ and its ``true'' score $\phi(\vx)$:
\begin{equation*}
L_{rec}(G) = \left\lVert \vx - G(\vx, \phi(\vx), k) \right\rVert_1.
\end{equation*}
This ensures that $G$ remains unchanged when the sample already has the targeted anomaly score, overall encouraging minimal changes. 

Similarly, the ``cycle consistency loss'' \citep{zhu2017unpaired} 
\begin{equation*}
L_{cyc}(G) = \left\lVert \vx - \tilde{\vx}_{\alpha, k} \right\rVert_1,
\end{equation*} where  $\tilde{\vx}_{\alpha, k} = G(\bar{\vx}_{\alpha, k}, \phi(\vx), k),$
encourages $G$ to recreate the sample $\vx$ when targeting its true anomaly score $\phi(\vx)$ and being conditioned on any generated sample $\bar{\vx}_{\alpha, k}$ based on $\vx$. It encourages minimal changes because the generator needs to be able to revert any change of $\vx$.

The concept loss drives $G$ to produce disentangled CEs:
\begin{equation*}
L_{con}(G, R) =  \crossentropy\left(k, R\big(\vx, \bar{\vx}_{\alpha, k}\big)\right) + \crossentropy\left(k, R\big(\bar{\vx}_{\alpha, k}, \tilde{\vx}_{\alpha, k} \big)\right),
\end{equation*}
where $\crossentropy$ denotes the cross entropy loss. 

In summary, the losses encourage the generated samples $\bar{\vx}_{\alpha, k}$ to be semantically distinguishable for different concepts $k$ while having an anomaly score of $\alpha$ according to $\phi$ and undergoing minimal changes with respect to the original $\vx$. This results in a disentangled set of $K$ counterfactual explanations for an anomaly $\vx^*$ with $\lbrace G(\vx^*, 0, 1), \dots, G(\vx^*, 0, K) \rbrace$. The generator can produce pseudo anomalies $G(\vx, \alpha, K)$ when $\phi(\vx) \approx 0$ and $\alpha \gg 0$, which help $G$ in learning how to turn anomalies into normal samples, when included in $L_\phi$.

\subsubsection{Diffusion Models for CE of Image AD.} 
This section proposes an approach for producing CEs of high-resolution anomaly detectors using state-of-the-art diffusion models. 
In particular, we incorporate DiffEdit \citep{couairon2023diffedit}. DiffEdit modifies the LAION-5B pre-trained text-conditional latent diffusion model known as Stable Diffusion \citep{rombach2022high} to semantically edit images. Let $A_{\gE}: \R^D \rightarrow \R^\Delta$ and $A_{\Omega}: \R^\Delta \rightarrow \R^D$ denote the encoder and decoder of the autoencoder used in Stable Diffusion. From a high-level perspective, the DiffEdit model can be defined as $\psi: \R^{\Delta \times T} \rightarrow \R^\Delta$ where $T$ denotes the output dimension of the word embedding model. 
For an image $\vx \in \R^D$, we retrieve a semantically modified version $\hat{\vx}$ controlled by the text prompt $t$ via $\hat{\vx} = A_{\Omega}(\psi(A_{\gE}(\vx), t))$. 
For more details, we refer to the paper by \citet{couairon2023diffedit}. 
We incorporate DiffEdit into the proposed framework by training the generator on its latent output.
That is, we redefine the generator $G(\vx, \alpha, k) = A_{\Omega}\left(G'(\psi(A_{\gE}(\vx), t), \alpha, k)\right)$ with $G': \R^\Delta \times [0, 1] \times \{1, \dots, K\} \rightarrow \R^\Delta$.
The text prompt $t$ is set to the normal class label (e.g., ``cat'' for cats being normal).
We train the generator $G$ (i.e., the parameters of $G'$) as before. 

\section{Theoretical Analysis}
The objective of the proposed method is intertwined with several interacting losses. 
Here, we provide a theoretical analysis on the performance of the optimization problem.
Let $V(\gD,G)=\lambda_{gan}\E_{\vx\sim p_X}\E_{\alpha,k}[L_\gD(\gD)]$ and $U(\gD,(G,R))=\E_{\vx\sim p_X}\E_{\alpha,k}[\lambda_{gan}L_G(G)-\lambda_{\phi}L_\phi(G)+\lambda_{rec}(L_{rec}(G)+L_{cyc}(G))+\lambda_rL_{con}(G,R)]$. $V$ trains $\gD$, and $U$ trains $G$ and $R$.
 \begin{definition}
     We say $(\gD^*, (G^*,R^*))$ is a Nash equilibrium of the system if $V(\gD,G^*) \le V(\gD^*,G^*)$ for any $\gD$ and $ U(\gD^*,(G^*,R^*)) \le U(\gD^*, (G,R) ) \text{ for any } G, R.$
 \end{definition}
\begin{theorem}\label{thm:nash}
Assume $G$ and $R$ have enough capacity. Let $(\gD^*, (G^*,R^*))$ be a Nash equilibrium of the system. 
(I) If $\lambda_\phi=\lambda_{rec}=\lambda_{con}=0$, then $\E_{\alpha,k}[p_{G^*(\alpha,k)} ]=p_X$ and  $V(\gD^*,G^*)=-2\lambda_{gan}$.
(II) If $\lambda_\phi=0$ and $\phi$ is nearly flat, then $\E_{\alpha,k}[p_{G^*(\alpha,k)} ] \approx p_X$ and V$(\gD^*,G^*) \approx -2\lambda_{gan}$. If we assume $\phi$ is flat, then $\E_{\alpha,k}[p_{G^*(\alpha,k)} ]=p_X$ and  $V(\gD^*,G^*)=-2\lambda_{gan}$.
\end{theorem}

Part (I) of Theorem \ref{thm:nash} shows that when training only with the modified GAN-based objectives $L_\gD$ and $L_G$, the generator indeed converges to the training data distribution $p_X$, similar to the original objective in \citet{goodfellow2020generative}. Part (II) shows that when including the losses that encourage minimal changes, we still obtain the distribution $p_X$ with flatness assumptions on the detector $\phi$. It follows that $L_\phi$ is the main antagonist that causes divergence from the training data distribution to produce reasonable CEs.

\begin{proof}[Proof of Theorem~\ref{thm:nash}] \appendixProofs provides the full proofs.  Here, we show very brief sketches.
The main idea for proving part (I) of the theorem is to divide the min-max problem into a min part and a max part and then analyze them separately. 
The maximizer $\gD^*$ of this problem has the explicit form $\gD^*(\vx)=1$ if $\E_{\alpha,k}[p_{G^*{\alpha,k}}](\vx)\le p_X(\vx)$ and $\gD^*(\vx)=0$ otherwise.  
Plugging this into $V(\gD^*,G^*)$, we get $V(\gD^*,G^*)\ge -2\lambda_{gan}$. 
According to the property of the Nash equilibrium, we know $\E_{\vx,\alpha,k}[L_G(G^*)]\le \E_{\vx,\alpha,k}[L_G(G)]$ holds for any $G$.  
Specifically considering $G$ as the ``ideal" generator with a density function $p_{G(\alpha,k)}=p_X$, we can establish $V(\gD^*,G^*) \le -2\lambda_{gan}.$
For part (II), the analysis of $V$ and $\gD^*$ are the same as for the max problem. For the min problem, from the flatness of $\phi$, we can show that 
\(\int    p_X(\vx)  \gD^*(\vx)- \E_{\alpha,k}[p_{G^*(\alpha,k)}](\vx)\gD^*( \vx ) d\vx  \le \epsilon.  \)
Here, $\epsilon>0$ is proportional to the flatness of $\phi$. If $\phi$ is almost flat, $\epsilon\approx 0$ and then $V(\gD^*,G^*)\approx -2\lambda_{gan}$. Hence, $\E_{\alpha,k}[p_{G^*(\alpha,k)}]\approx p_X$.
If $\phi$ is flat, then $\epsilon=0$. It implies  $V(\gD^*,G^*)= -2\lambda_{gan}$  and $\E_{\alpha,k}[p_{G^*(\alpha,k)} ]=  p_X$. 
\end{proof} 

\begin{theorem}\label{thm:general}
Assume $G$ and $R$ have enough capacity, $\lambda_\phi>0$, and $\text{Prob}(\phi(\vx)=0 \cup \phi(\vx)=1)>0$, $\vx\sim p_X$. Let $(\gD^*,(G^*,R^*))$ be a Nash equilibrium of the system. Then $p_{G^*(\alpha,k)}\neq p_X$ for any $\alpha,k$ and $V(\gD^*,G^*)\neq -2\lambda_{gan}$.
\end{theorem}
Theorem \ref{thm:general} shows that $L_\phi$ indeed causes a divergence from the training data distribution, implying that the generator learns to map samples to anomalous data regimes when $\alpha > 0$. Empirically, our experiments show that with $\alpha=0$, the generator consistently creates normal samples. 

\begin{proof}[Proof of Theorem~\ref{thm:general}] We prove the theorem by contradiction. For a Nash equilibrium $(\gD^*,(G^*,R^*))$, it holds $U(\gD^*,(G^*,R^*))\le U(\gD^*,(G,R))$ for any $G,R$. We show that if $(\gD^*,(G^*,R^*))$ is a Nash equilibrium with $p_{G^*(\alpha,k)}= p_X$ and $V(\gD^*,G^*)= -2\lambda_{gan}$, then there exists a generator $G'$ such that $U(\gD^*,(G',R^*))<U(\gD^*,(G^*,R^*))$. This violates $U(\gD^*,(G^*,R^*))\le U(\gD^*,(G',R^*))$. Hence, $p_{G^*(\alpha,k)}\neq p_X$ and $V(\gD^*,G^*)\neq -2\lambda_{gan}$. We choose $G'$ as satisfying $\phi(G'(\vx,\alpha,k))=\alpha$ and being Lipschitz continuous w.r.t.~the first and second argument. By noting that $\E_{\vx,\alpha,k}L_{\phi}(G^*)=-\infty$ when $\text{Prob}(\phi(\vx)=0 \cup \phi(\vx)=1)>0$ and $U(\gD^*,(G',R^*))$ is uniformly bounded, we conclude $U(\gD^*,(G',R^*))<U(\gD^*,(G^*,R^*))$.  \end{proof}

\section{Experiments}\label{sec:exp}
We empirically assess the capabilities of CEs for deep AD, providing qualitative and quantitative evidence of the superiority of the proposed CEs over traditional methods. 

\paragraph{Deep Anomaly Detection Methods.}
We specifically study three state-of-the-art \emph{semantic image AD} methods. 
\emph{DSVDD} \citep{ruff2018deep} trains a neural network to enclose the (mostly normal) unlabeled training data by a minimal volume hypersphere with the distance to the sphere's center becoming the anomaly score. \citet{hendrycks2019deep} showed that Outlier Exposure (OE)---using a large unstructured collection of natural images as example anomalies during training---consistently outperforms previous AD methods, while still being unsupervised. A neural network learns to differentiate normal data from OE samples with a \emph{Binary Cross Entropy (BCE)} loss. \citet{liznerski2022exposing} introduced \emph{HSC} as a modification of the DSVDD loss to enable it to take advantage of OE. Since the CE generator requires bounded anomaly scores, we slightly adjust some of the objectives without impacting the performance. A more detailed description can be found in \appendixADMethods.

\paragraph{Datasets.}
We evaluate on \emph{MNIST} \citep{mnist}, \emph{Colored-MNIST}, \emph{CIFAR-10} \citep{krizhevsky2009learning}, and \emph{GTSDB} \citep{Houben2013gtsdb}. Furthermore, we introduce ImageNet-Neighbors (INN), a subset of ImageNet-1k~\citep{ILSVRC15} designed for AD tasks. INN comprises multiple AD setups; in each setup, one ImageNet-1k class is considered normal, and the ten most semantically similar classes, based on the Wu-Palmer similarity metric~\citep{wu1994verb}, are defined as ground-truth test anomalies. We follow previous work for using disjoint OE data for the OE-based AD methods. Details are in \appendixDatasets.

\paragraph{Experimental Setup.} 
Following previous work on semantic image AD \citep{ruff2018deep, golan2018deep, hendrycks2019using, tack2020csi, ruff2021, liznerski2022exposing}, we convert the multi-class classification datasets into AD benchmarks. This is achieved by defining a subset of the classes to be normal and using the remaining classes as ground-truth anomalies during testing. When only one class is considered normal, this approach is known as one vs.~rest. 
We also explore a variation in which multiple classes are normal. This emulates a multifaceted normal class that includes different notions of normality. 
Finally, we consider the INN setup where we have particular ground-truth anomalies per normal class. For INN, we generate CEs with diffusion models, while for all other datasets we use the pure GAN-based model. 

Overall, we consider over $80$ different AD setups. Details of the specific AD setups are provided in \appendixQuantitative. Our quantitative analysis reports results averaged over all scenarios and multiple seeds per dataset. Detailed quantitative results for each scenario are in \appendixQuantitative and further qualitative results in \appendixQualitative. Hyperparameters and architectures are explained in detail in \appendixHyperparameters, and there is a small hyperparameter sensitivity analysis in \appendixHyperparameterAnalysis. 

\subsection{Qualitative Results\label{sec:qualitative_results}}
We present qualitative examples of CEs, demonstrating the benefit of using CE for semantic image AD over traditional explanation methods. 
We refer to \appendixQualitative for more examples using further normal scenarios with similar findings.

\subsubsection{CEs Explain why Images are Predicted Anomalous} \label{sec:visual_examples}

\begin{figure}[h]
    \centering
     \begin{subfigure}{0.145\textwidth}
        \includegraphics[width=\textwidth]{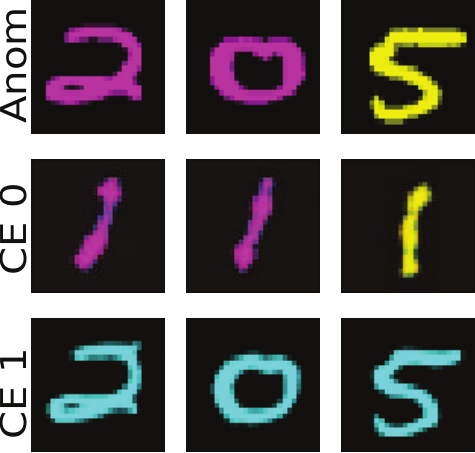}
        \caption{BCE (OE)}
    \end{subfigure}
    \hspace{3pt}
    \begin{subfigure}{0.145\textwidth}
        \includegraphics[width=\textwidth]{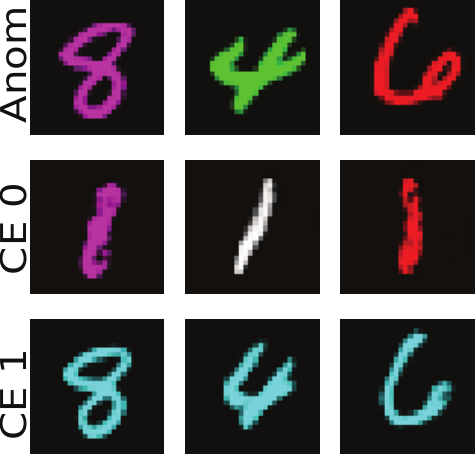}
        \caption{HSC (OE)}
    \end{subfigure}
    \hspace{3pt}
    \begin{subfigure}{0.145\textwidth}
        \includegraphics[width=\textwidth]{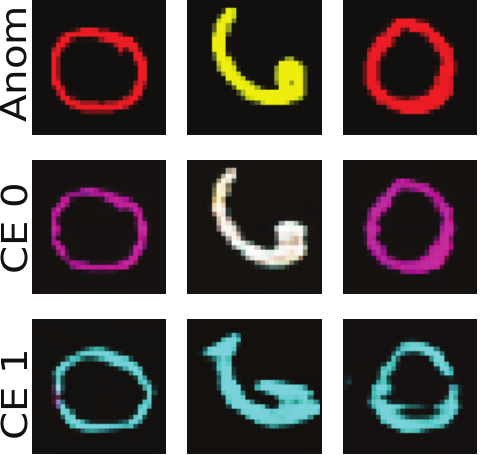}
        \caption{DSVDD}
    \end{subfigure}
    \caption{Examples of CEs for Colored-MNIST, with cyan digits and the digit one serving as the normal class. The first row shows anomalous images and the next two rows their corresponding CEs using two concepts.
    The CEs of BCE and HSC appear normal and realistic for each concept.} 
    \label{fig:cf_colored_mnist}
\end{figure}

\paragraph{Colored-MNIST.} Figure \ref{fig:cf_colored_mnist} shows CEs when the normal class is formed from instances of the digit one and digits colored cyan. 
We observe that the CEs generated to explain the BCE detector align well with our expectation. 
The proposed method transforms the anomalies into ones without changing the color, or their color is changed to cyan without changing the digit.
Both modifications are minimal alterations of the anomaly, transforming its appearance to normality in two distinct ways. 
The CEs of HSC also correspond to normal samples. However, in one case, both the color and the digit is changed, resulting in unnecessary changes. 
We found that this behavior represents a local optimum of the proposed objective, highlighting the inherent difficulty of the unsupervised generation of CEs. 
The CEs created to explain the DSVDD detector perform least effectively. 
They appear normal for one concept but often fail for the other. This behavior may be attributed to DSVDD's limited ability to detect anomalies. 
See \appendixLimitations for details. 

\begin{figure}[!ht]
    \centering
    \begin{subfigure}{0.145\textwidth}
        \includegraphics[width=\textwidth]{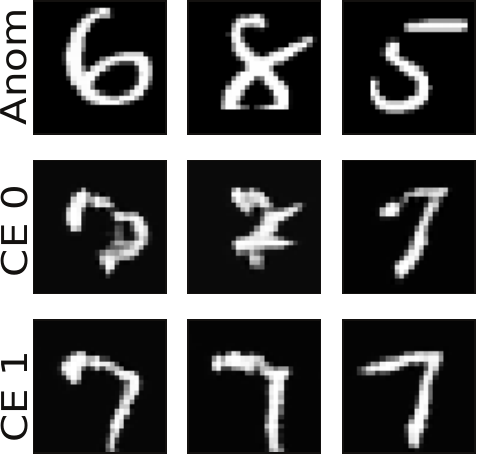}
        \caption{BCE (OE)}
    \end{subfigure}
    \hspace{3pt}
    \begin{subfigure}{0.145\textwidth}
        \includegraphics[width=\textwidth]{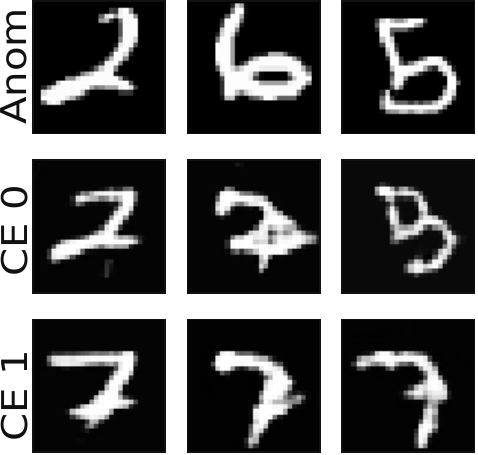}
        \caption{HSC (OE)}
    \end{subfigure}
    \hspace{3pt}
    \begin{subfigure}{0.145\textwidth}
        \includegraphics[width=0.994\textwidth]{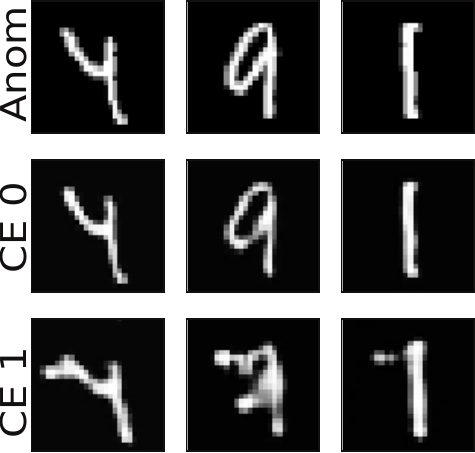}
        \caption{DSVDD}
    \end{subfigure}
    \caption{CEs for MNIST with seven as the normal class. The first row shows anomalies, the other two rows CEs using two different concepts. CEs of BCE and HSC are variations of seven and thus represent intuitive counterfactuals.}
    \label{fig:cf_mnist_single}
\end{figure}

\paragraph{MNIST.} In Figure \ref{fig:cf_mnist_single}, a seven is considered normal. The CEs of BCE and HSC are meaningful: the anomalies are transformed into plausible variations of seven. 
Since the normal class consists of a single, monochromatic digit, the generator primarily learns to manipulate the presence or absence of the horizontal bar characteristic of a seven to distinguish between concepts. 
Therefore, the CEs indicate that the detectors do not heavily rely on the horizontal bar to rate anomalousness properly.

\begin{figure}[!ht]
    \centering
    \begin{subfigure}{0.145\textwidth}
        \includegraphics[width=\textwidth]{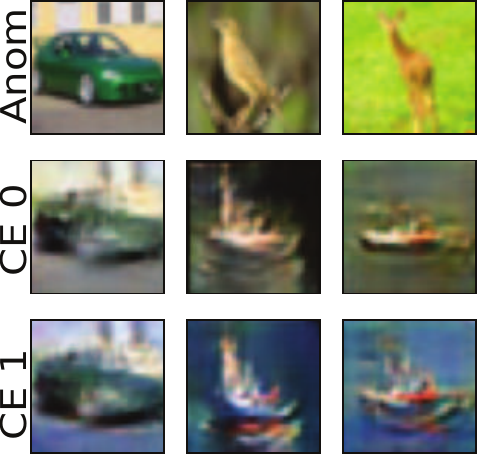}
        \caption{BCE (OE)}
    \end{subfigure}
    \hspace{3pt}
    \begin{subfigure}{0.145\textwidth}
        \includegraphics[width=\textwidth]{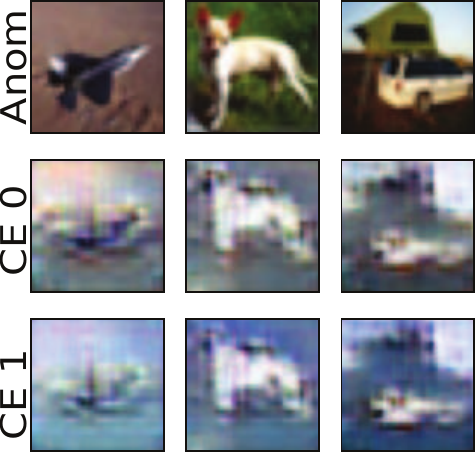}
        \caption{HSC (OE)}
    \end{subfigure}
    \hspace{3pt}
    \begin{subfigure}{0.145\textwidth}
        \includegraphics[width=\textwidth]{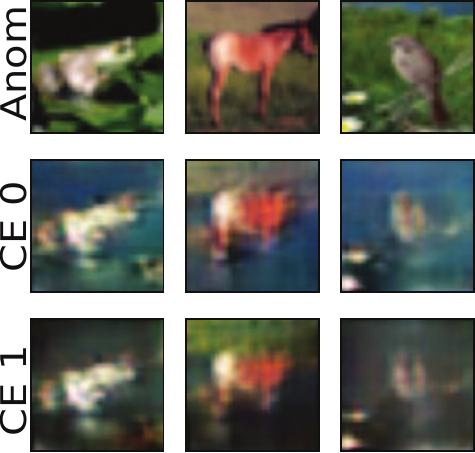}
        \caption{DSVDD}
    \end{subfigure}
    \caption{Example CEs for CIFAR-10 when ships are normal. The rows show anomalies and CEs for two concepts, respectively. The CEs of BCE display normal ships, varying the background for successful disentanglement while keeping the object's color to avoid unnecessary changes. }
    \label{fig:cf_cifar_single}
\end{figure}

\paragraph{CIFAR-10.} 
Especially for BCE, the CEs in Figure \ref{fig:cf_cifar_single} represent intuitive normal samples (ships) that retain the anomalous object's color to incur minimal changes on the anomaly. As there is only one single normal class, the CEs primarily disentangle the concepts by changing the background. Ships are typically depicted floating on water, which may vary in color. 
The CEs reveal that HSC and DSVDD predominantly focus on the background to detect anomalies, as all CEs are perceived as normal (see Section \ref{sec:cf_are_normal}), although the CEs' foregrounds are often similar to the original anomaly.  
This aligns with prior findings on background exploitation in CIFAR-10 \citep{ding2024improving}.

\begin{figure}[ht]
    \centering
    \begin{subfigure}{0.145\textwidth}
        \includegraphics[width=\textwidth]{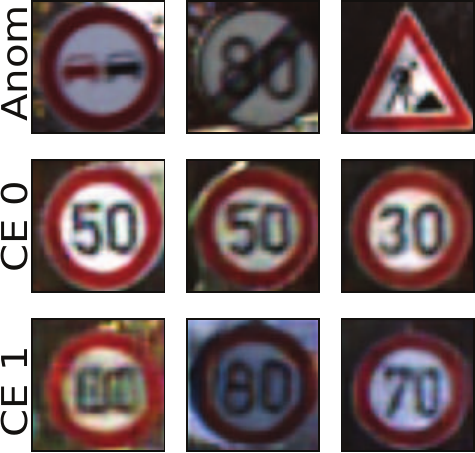}
        \caption{BCE (OE)}
    \end{subfigure}
    \hspace{3pt}
    \begin{subfigure}{0.145\textwidth}
        \includegraphics[width=\textwidth]{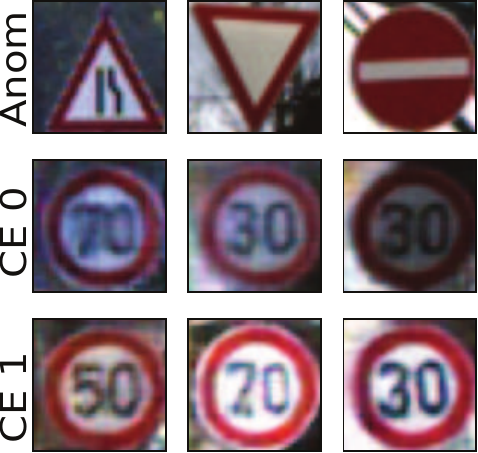}
        \caption{HSC (OE)}
    \end{subfigure}
    \hspace{3pt}
    \begin{subfigure}{0.145\textwidth}
        \includegraphics[width=\textwidth]{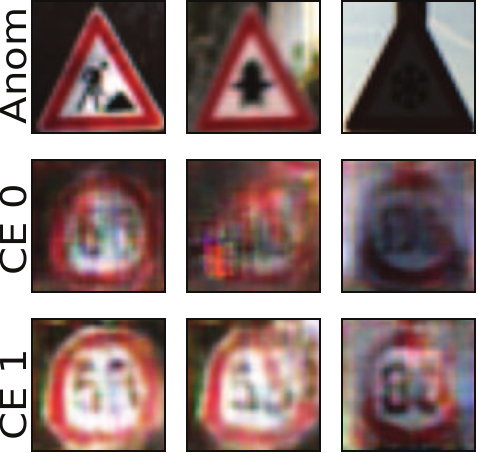}
        \caption{DSVDD}
    \end{subfigure}
    \caption{CEs for GTSDB with speed signs being normal. The first row shows anomalies, the other rows present disentangled CEs, which appear as different normal speed signs.}
    \label{fig:cf_gtsdb}
\end{figure}

\paragraph{GTSDB.}
Figure \ref{fig:cf_gtsdb} shows the proposed CEs when speed signs are taken as a normal class. 
The CEs of BCE and HSC show well-disentangled normal traffic signs, obtained from anomalous ones.
For instance, the CE of BCE changes the ``80km/h restriction ends'' sign into a ``80km/h limit'' sign---a minimal intervention to make the sample appear normal.
Note that all triangular anomalies become circles. The CEs show that the shape is an important feature for the detector to rate anomalousness. 

\begin{figure}[!ht]
    \centering
    \begin{subfigure}{0.22\textwidth}
        \includegraphics[width=\textwidth]{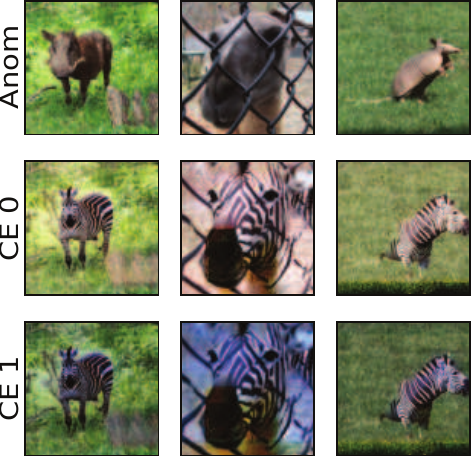}
        \caption{BCE (OE)}
    \end{subfigure}
    \hspace{6pt}
    \begin{subfigure}{0.216\textwidth}
        \includegraphics[width=\textwidth]{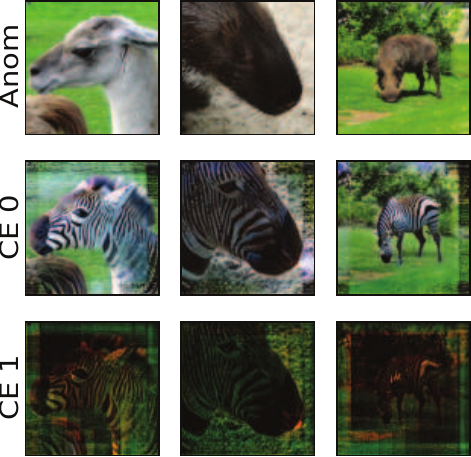}
        \caption{HSC (OE)}
        \label{fig:cf_inn_single_hsc}
    \end{subfigure}
    \caption{Examples of CEs for INN, where zebras are considered normal. The first row shows anomalous images, the other two rows present CEs using two different concepts. 
    }
    \label{fig:cf_inn_single}
\end{figure}

\paragraph{ImageNet-Neighbors.} 
Figure \ref{fig:cf_inn_single} shows CEs for the INN dataset when zebras are normal. 
The ground-truth anomalies are ``similar'' animals, ranging from horses and boars to armadillos. 
The CEs depict zebras while keeping the general pose and background of the anomalous animal. For disentanglement, the CEs vary the color scheme, which apparently the detectors perceive as normal. The CEs for the second concept for HSC are dark and, while still showing zebras, perturb the image with green and orange patterns. Interestingly, our experiments show that HSC assigns lower anomaly scores to the CEs for this second concept. 

\begin{figure}[h]
\centering
\includegraphics[width=0.99\columnwidth]{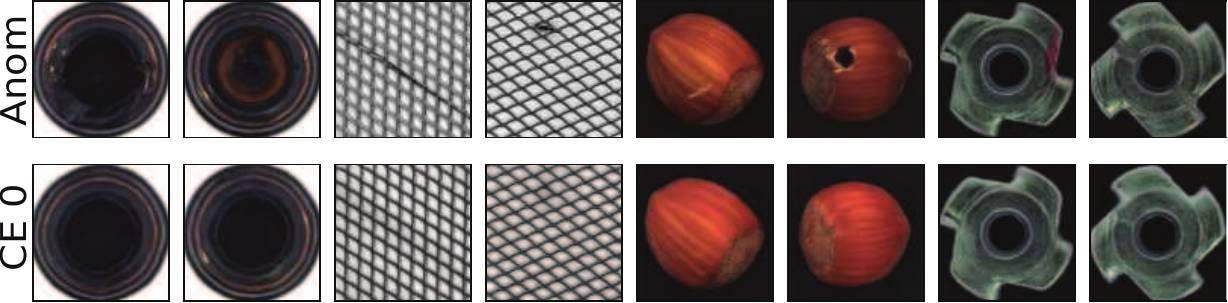}
\caption{CEs of the BCE detector for MVTec-AD and ImageNet-21k as OE. The top row shows anomalies as defects. The bottom row shows corresponding CEs, resembling normal, healthy images.}
\label{fig:mvtec}
\end{figure}

\paragraph{MVTec-AD.}
\emph{Our experiments focus on semantic image-AD} \citep{ruff2021} \emph{rather than low-level AD} where anomalies are defects instead of out-of-class. Figure~\ref{fig:mvtec} shows CEs for such a scenario with the MVTec-AD dataset \citep{bergmann2019mvtec}. 
While the CEs are of good quality and correct, they do not provide valuable insight, as here the behavior of an anomaly detector is less opaque and requires no counterfactual explanation. We include further reasoning for this and more results in \appendixMvtec.

\subsubsection{CEs Explain why Images are Predicted Anomalous---\textit{even when Feature Attribution Fails}. } \label{sec:cf_vs_fcdd}
\begin{figure}[h]
    \centering
    \begin{subfigure}{0.17\textwidth}
        \includegraphics[width=\textwidth]{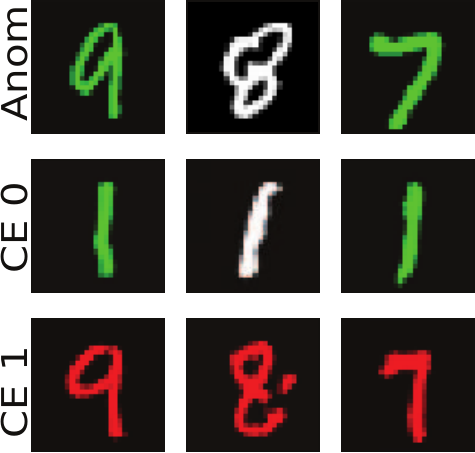}
        \caption{Counterfactuals}
    \end{subfigure}
    \hspace{5pt}
    \begin{subfigure}{0.17\textwidth}
        \includegraphics[width=\textwidth]{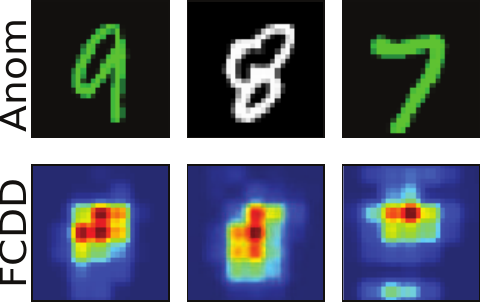}
        \vspace{1.6em}  
        \caption{FCDD's Heatmaps} 
    \end{subfigure}
    \caption{The first row shows anomalies from Colored-MNIST, with red digits and the digit one forming the normal class. The other rows show (a) corresponding CEs and (b) anomaly heatmaps from FCDD. The CEs explain the detector that perceives anomalies turned red or into one as normal, while heatmaps just highlight the difference to one.}
    \label{fig:cf_vs_fcdd}
\end{figure}

We demonstrate the advantage of the proposed CEs over conventional explanations that attribute features. 
Figure~\ref{fig:cf_vs_fcdd} shows (a) CEs generated with our method and (b) heatmaps for the corresponding anomalies generated with FCDD \citep{liznerski2021}. 
FCDD's heatmaps explain only spatial aspects of the AD: It highlights the horizontal bar in digit seven, the circle in digit nine, and all of digit eight. These spatial aspects are also explained by the CEs created for the first concept, where anomalies are turned into the digit one. However, FCDD's heatmaps fail to identify the color as  anomalous, whereas our CEs capture this aspect with their second concept, where the anomalies are colored red, making them look normal. This shows that CEs can provide more holistic explanations. 

\subsection{Quantitative Results\label{sec:quantitative_results}}
This section presents a quantitative analysis of the CEs, assessing their normality, realism, and disentanglement in terms of metrics based on AuROC, FID, and accuracy. 
The metrics are described in detail in \appendixMetrics. 

\subsubsection{The CEs Appear as Normal.} \label{sec:cf_are_normal}
An important attribute for any CE in AD is that it must be perceived as normal by the anomaly detector. 
To evaluate this quality criterion, we compare the anomaly scores of the normal test samples with those of the generated CEs in terms of AuROC. Ideally, the AuROC should approach $50\%$, indicating that CEs and normal samples are indistinguishable. As shown in Table \ref{tbl:main_CF_AD_AuROC}, the AuROC is indeed very close to $50\%$ on CIFAR-10, GTSDB, and Colored-MNIST (here abbreviated as C-MNIST), underlining that the detector perceives the CEs as normal. Only on MNIST and INN, some of the CEs appear anomalous. This might be due to the enforced disentanglement that produces diverse samples despite a limited variety of possible normal variations. 

\begin{table}[ht]
\centering
\setlength\tabcolsep{3pt}
\renewcommand{\arraystretch}{1}
\begin{tabular}{@{}l|lccc@{}}
\toprule
&\multirow{2.5}{*}{Datasets} & \multicolumn{3}{c}{Methods}\\
\cmidrule{3-5}
& & BCE OE & HSC OE & DSVDD\\
\midrule
\multirow{3}{3.4em}{Single normal class} & MNIST & 72.0 $\pm$ \;4.0 & 80.8 $\pm$ \;5.3 & 75.2 $\pm$ 9.2 \\
                                         & CIFAR-10 & 47.5 $\pm$ 10.0 & 49.9 $\pm$ \;4.4 & 54.6 $\pm$ 3.4 \\
                                         & INN & 69.1 $\pm$ 18.1  & 67.9 $\pm$ 13.2  & $\times$  \\\midrule
\multirow{4}{3.4em}{Multiple normal classes}& C-MNIST & 55.6 $\pm$ 1.5 & 55.8 $\pm$ \;4.7 & 61.5 $\pm$ 4.3 \\
                                           & MNIST & 78.1 $\pm$ 4.1 & 82.1 $\pm$ \;3.8 & 73.4 $\pm$ 6.5 \\
                                           & CIFAR-10 & 49.0 $\pm$ 8.5 & 44.4 $\pm$ \;6.7 & 50.7 $\pm$ 3.3 \\
                                           & GTDSB & 50.2 $\pm$ 8.0 & 48.6 $\pm$ 14.4 & 53.1 $\pm$ 4.8 \\
\bottomrule
\end{tabular}
\caption{AuROC of normal test data vs.~CEs. The CEs appear entirely normal for values $\leq 50\%$.
}
\label{tbl:main_CF_AD_AuROC}
\end{table}

\subsubsection{The CEs are Realistic.} \label{sec:cf_are_realistic}
To assess the realism, we compute the FID \citep{heusel2017gans} between the CEs and normal test samples, and then normalize it by dividing by the FID between normal and anomalous test samples. 
The normalized FID$_{N}$ is $100\%$ if the CEs are as realistic as the anomalies. 
We found that a FID$_{N}$ of $50$ to $100\%$ is a reasonable target for expressive CEs. 
If the CEs became too similar to the normal data distribution, they would not be valid counterfactuals, as they would not retain non-anomalous features from the anomalies. 
Table \ref{tbl:main_FID_N} displays the normalized FID$_{N}$ scores. 
The CEs for BCE and HSC are mostly as realistic as the anomalies. 
On MNIST, INN, and Colored-MNIST, the CEs are even more realistic. 

\begin{table}[ht]
\centering
\setlength\tabcolsep{2.5pt}
\renewcommand{\arraystretch}{1}
\begin{tabular}{@{}l|lccc@{}}
\toprule
&\multirow{2.5}{*}{Datasets} & \multicolumn{3}{c}{Methods}\\
\cmidrule{3-5}
& & BCE OE & HSC OE & DSVDD\\
\midrule
\multirow{3}{3.4em}{Single normal class} & MNIST & 43 $\pm$ 8.1 & 68 $\pm$ 14.6 & 100 $\pm$ 8.8 \\
                                        & CIFAR-10 & 116 $\pm$ 20.8 & 300 $\pm$ 90.0 & 116 $\pm$ 12.0 \\
                                        & INN & 85.0 $\pm$ 28.6 & 85.4 $\pm$ 24.6  & $\times$  \\\midrule
\multirow{4}{3.4em}{Multiple normal classes}& C-MNIST & 56 $\pm$ 12.4 & 95 $\pm$ 30.5 & 83 $\pm$ 8.7 \\
                                        & MNIST & 78 $\pm$ 26.0 & 96 $\pm$ 25.0 & 100 $\pm$ 10.7 \\
                                        & CIFAR-10 & 103 $\pm$ 27.9 & 254 $\pm$ 69.7 & 110 $\pm$ 10.0 \\
                                        & GTDSB & 110 $\pm$ 101 & 95 $\pm$ 73.5 & 131 $\pm$ 118 \\
\bottomrule
\end{tabular}
\caption{FID$_{N}$ for the CEs. Most of the CEs are as realistic as the anomalies, which are also realistic since they follow the general data distribution (e.g., are digits for MNIST).}
\label{tbl:main_FID_N}
\end{table}

\subsubsection{The CEs Capture Multiple Disentangled Concepts.} \label{sec:cf_are_disentangled}
In \appendixConceptAcc, we report the accuracy of the concept classifier. 
The classifier almost always achieves an accuracy of more than 90\%.
Exceptions are for DSVDD and MNIST, HSC and MNIST, and BCE and CIFAR-10, where it scores roughly 80\%.
This provides evidence that, for each anomaly, our method generates concept-disentangled CEs.

\subsubsection{The CEs Are Minimal Modifications.} \label{sec:cf_min_edits_eval}
To assess the minimality of the modifications present in the CEs, we report the LPIPS \citep{zhang2018unreasonable} and MSE between test anomalies and their corresponding CEs in \appendixMinEdits.
The results indicate that the modifications are reasonably limited, as both metrics attain comparatively low values.

\subsection{The CEs Reveal a Classifier Bias in Deep AD} \label{sec:cf_for_biased_classifiers}
\begin{figure}[!ht]
    \centering
    \begin{subfigure}{0.17\textwidth}
        \includegraphics[width=\textwidth]{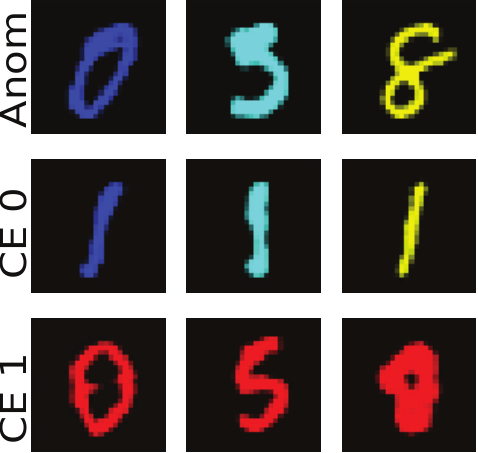}
        \caption{BCE with OE}
    \end{subfigure}
    \hspace{1em}
    \begin{subfigure}{0.17\textwidth}
        \includegraphics[width=\textwidth]{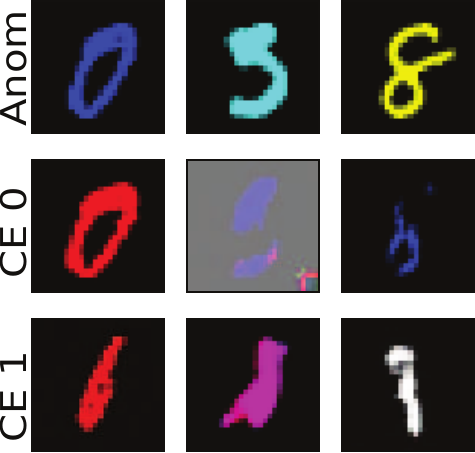} 
        \caption{BCE with blue}
    \end{subfigure}
    \caption{The first row shows anomalies for C-MNIST with red digits and the digit one as the normal class. The other rows show CEs of BCE trained with OE in (a) and a classifier trained with only blue anomalies in (b). The generator's inability to generate normal-looking CEs for non-blue anomalies suggests that the classifier (b) is biased.}
    \label{fig:cf_biased_clf}
\end{figure}

The hypothesis of a ``classification bias,'' suggesting supervised classifiers underperform when trained with limited and biased anomaly subsets \citep{ruff2020}, remains insufficiently investigated. 
To test this hypothesis, we train a supervised classifier on Colored-MNIST to distinguish between a normal set (red digits and the digit one) and a subset of the ground-truth anomalies (all blue anomalies). This simulates a realistic scenario in which one has no access to all variations of the ground-truth anomalies. 
The classifier bias becomes apparent as the AuROC of normal test vs.~ground-truth anomalies decreases from $98\%$ for BCE with OE (unsupervised) to $75\%$ for supervised BCE. 
Our CEs illuminate this phenomenon in Figure \ref{fig:cf_biased_clf}.
The CEs of the AD method in (a) indicate that anomalies should be transformed into red or digit one to appear normal. 
For the supervised classifier in (b), only for the blue anomalous zero, which is seen during training, the CEs show normal versions of the anomaly.  
For unseen anomalies, such as the yellow eight, the CEs do not show intuitive normal images. This suggests that the classifier is biased towards blue anomalies and fails to generalize to colors not present in the training data. 
The experiment underlines the need for specialized AD methods (e.g., using OE or semi-supervision) because they are less prone to bias.

\section{Conclusion}

This paper introduced a novel method that can interpret image anomaly detectors at a semantic level. This is achieved by modifying anomalies until they are perceived as normal by the detector, creating instances known as counterfactuals. 
We found that counterfactuals can provide a deeper, more nuanced understanding of image anomaly detectors, far beyond the traditional feature-attribution level. Extensive experiments across various image benchmarks and deep anomaly detectors demonstrated the efficacy of the proposed approach, particularly also where conventional techniques fail. 
This research marks a paradigm shift and a significant departure from the more superficial interpretation of anomaly detectors using feature attribution. 
This may be a substantial milestone in the pursuit of more transparent and accountable AD systems. 

\section*{Acknowledgments} 
MK and SF acknowledge support by the BMFTR award 01IS24071A, by the DFG through FOR 5359 (ID 459419731), TRR 375 (ID 511263698), SPP 2298 (Id 441826958), and SPP 2331 (441958259, 553345933, 466468799), and by the Carl-Zeiss Foundation through the initiatives AI-Care and Process Engineering 4.0. The work of PW is partially supported by the Alexander von Humboldt Foundation. SV acknowledges support by the BMFTR award 01IW23005.

\bibliography{aaai2026}

\appendix
\onecolumn
\setcounter{page}{1} 

\section{Broader Impact} \label{appx:sec:broader_impact}

As an explanation technique, our method naturally aids in making deep AD more transparent.
It may reveal biases in the model (see Section \ref{sec:cf_for_biased_classifiers}) and improve trustworthiness. 
For example, it may reveal a social bias when a portrait of a person is labeled anomalous due to race or gender. 
In this scenario, our method might generate CEs where merely the skin color has been changed. 
Applying our method can prevent a harmful deployment of such an AD model.

\section{Counterfactual Explanations of Defects} \label{appx:sec:mvtec}
In the main paper, we did not include full experiments on datasets such as MVTec-AD, where anomalies are subtle modifications of normal samples (e.g., cracked hazelnuts for healthy hazelnuts being normal) rather than being out of class. Such datasets are not interesting in the context of high-level explanations. Contrary to usual assumptions in AD, where anomalies are \emph{everything}, which is not normal, in MVTec-AD there is a very precise definition of anomalousness and only one specific way to turn anomalies normal (i.e., by removing the defect). CEs would not help in understanding the model. Hence, we focus on the well-established and important semantic image-AD setting.

\begin{figure}[h]
\centering

\begin{subfigure}{0.99\textwidth}
    \includegraphics[width=0.99\textwidth]{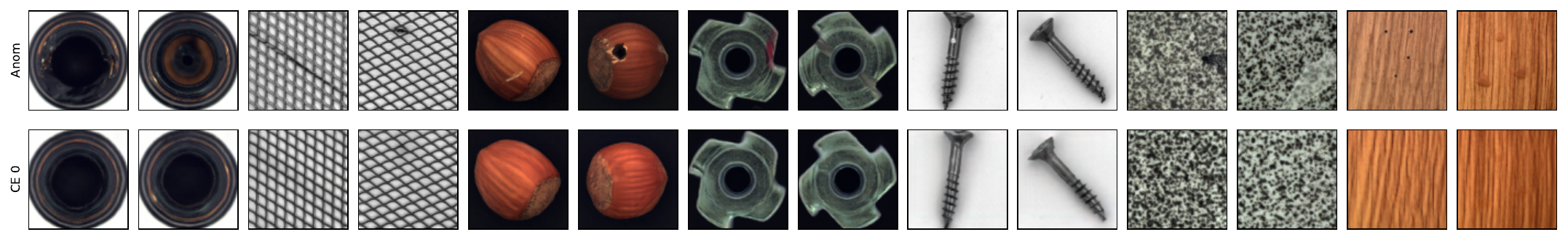}
\end{subfigure}

\caption{CEs for MVTec-AD and an anomaly detector trained with BCE and ImageNet-21k as OE. For each class, a different detector and CE generator was trained. The first row shows anomalies, the other corresponding CEs. }
\label{fig:appx_mvtec}
\end{figure}

To visualize why CEs are not a useful tool for explaining low-level AD, we trained our proposed method from scratch with a single concept on several classes of MVTec-AD. Figure \ref{fig:appx_mvtec} shows some generated CEs for the classes bottle, grid, hazelnut, metal nut, screw, tile, and wood. Mostly, the CEs are high-quality: realistic and normal. However, they do not help us to understand the behavior of the model. They simply show the sample with the defect removed, which is a trivial explanation of the anomaly but does not explain the anomaly detector. 

\section{Proof of the Theorems in the main paper} \label{appx:proofs}

\begin{problem}{(Counterfactual explanation of image AD)} \label{eq:minmax}
\begin{align*}
    \min_{G, R} \max_{\gD} \quad \E_{\vx\sim p_X} \E_{\alpha, k} \big[ \;\; &\lambda_{gan} \left(L_{\gD}(\gD) + L_{G}(G)\right) -  \lambda_\phi L_{\phi}(G) \nonumber  \\   
      +  &\lambda_{rec} \left(L_{rec}(G) +  L_{cyc}(G) \right) + \lambda_r L_{con}(G, R) \;\; \big], 
\end{align*}
with
\[ L_\gD(\gD)=  \min\big( 0,-1+ \gD(x) \big)   +   \min\big(0, -1-\gD( G(\vx,\alpha,k) )\big)   ,\]
 \[ L_G(G)= -   \min(0, -1+\gD(G(\vx,\alpha,k)))  ,\]
\[ L_{\phi}(G)=   \alpha \log( \phi(G(\vx,\alpha,k) ) + ( 1-\alpha)\log( 1-\phi( G(\vx,\alpha,k) ) )  ,\]
\[  L_{rec}(G)=    \|\vx- G(\vx,\phi(\vx),k) \|_1 ,\]
\[ L_{cyc}(G)=    \| \vx- G(\bar{\vx}_{\alpha,k},\phi(\vx),k) \|_1   \text{ with } \bar{x}_{\alpha,k}=G(\vx,\alpha,k) ,\]
\[ L_{con}(G,R)= \crossentropy\big(k,R(\vx,\bar{\vx}_{\alpha,k}) \big) + \crossentropy\big(k,R( \bar{\vx}_{\alpha,k}, G(\bar{\vx}_{\alpha,k},\phi(\vx),k) ) \big)  .\]
\end{problem}

Without loss of generality, we assume $\alpha$ is discrete. Denote
\begin{align*}
  V(\gD,G)&=\E_{\vx\sim p_X}\E_{\alpha,k}\big[ \lambda_{gan}L_\gD(\gD)\big],
\end{align*}
and
\begin{align*}
    U(\gD,(G,R)) = \E_{\vx\sim p_X}\E_{\alpha,k}\big[ \;\; &\lambda_{gan}L_G(G)-\lambda_{\phi}L_{\phi} (G) \\ 
    + &\lambda_{rec}(L_{rec}(G) + L_{cyc}(G)) + \lambda_{con} L_{con}(G,R) \;\; \big]. 
\end{align*}
We use $V(\gD,G)$ to train $\gD$ and use $U(\gD,(G,R))$ to train $(G,R)$. 

\begin{definition}
    $(\gD^*, (G^*,R^*))$ is a Nash equilibrium of System (\ref{eq:minmax}) if
\[ V(\gD,G^*) \le V(\gD^*,G^*) \text{ for any } \gD,\] 
\[ U(\gD^*,(G^*,R^*)) \le U(\gD^*, (G,R) ) \text{ for any } G, R.\]
\end{definition}
\noindent Let $p_{G(\alpha,k)}$ denote a density function of $G(\vx,\alpha,k)$ conditional on $k\in[K]$ and $\alpha\in[0,1]$ with $\vx\sim p_X$.
Let $\tilde{G}$ be the ideal vanilla generator if for any $\alpha,k$, it holds $p_{\tilde{G}_{\alpha,k}}=p_X$ and $\vx=\tilde{G}(\vx,\phi(\vx),k)$.   
We say $\tilde{G}(\vx,\cdot,k)$ is $\beta_{d}$-Lipschitz w.r.t the second argument if for any $x$
\[ \|\tilde{G}(\vx,\phi_1,k) -  \tilde{G}(\vx,\phi_2,k) \|_1 \le \beta_{d}| \phi_1 -\phi_2 |.\] 
We say $\phi(\cdot)$ is nearly flat if $|\phi(\vx_1)-\phi(\vx_2)|\le \beta_\phi\|\vx_1-\vx_2\|_1$ with $\beta_\phi$ being almost zero and say $\phi(\cdot)$ is flat if $\beta_\phi=0$.




\begin{lemma}\label{lem:V}
For any fixed generator $G$, $V(\gD,G)$ reaches its maximum in $1$ if $\E_{\alpha,k}[ p_{G(\alpha,k)}](\vx)  \le p_X (\vx) $ and otherwise in $0$. 
If $(\gD^*, (G^*,R^*))$ is a Nash equilibrium of System (\ref{eq:minmax}), then it holds $-\lambda_{gan} \ge V(\gD^*,G^*) \ge -2\lambda_{gan}$. In addition,  
\begin{align}\label{eq:Vgan}
    V(\gD^*,G^*)
    =- \lambda_{gan} \Big(1+ \int   &\mathbb{I}\big\{\E_{\alpha,k}[p_{G^*(\alpha,k)} ](\vx) \le  p_X(\vx)  \big\} \E_{\alpha,k}[p_{G^*(\alpha,k)} ](\vx) \nonumber\\ &+ \mathbb{I}\big\{\E_{\alpha,k}[p_{G^*(\alpha,k)} ](\vx) >  p_X(\vx)  \big\}  p_X(\vx) d\vx \Big),
\end{align}
and   \begin{align}\label{eq:V*}  
V(\gD^*,G^*) =- {\lambda_{gan}}  \Big(2+ \big(\int  \E_{\alpha,k} [p_{G^*(\alpha,k)}] (\vx) \gD^*(\vx ) -  p_X(\vx)  \gD^*(\vx)  d\vx \big)\Big).\end{align}

\end{lemma}
\begin{proof}
  For any fixed $G$,  from the definition of $V(\gD,G)$, we know
   \begin{align}\label{eq:V}
        &V(\gD,G)\nonumber\\
        &=
         \begin{aligned}[t]
        \lambda_{gan}\Big(\int p_X(\vx)    &\min\big( 0,-1+ \gD(\vx) \big)   d\vx
        +  \\&\frac{1}{K|\alpha|}\sum_{k=1}^K \sum_{\alpha=0}^1 \int p_X(z)     \min\big(0, -1-\gD( G(z,\alpha,k) )\big)   dz \Big)
        \end{aligned}
        \nonumber\\
        &=  
         \begin{aligned}[t]
        \lambda_{gan}\Big(\int  p_X(\vx)   &\min\big( 0,-1+ \gD(\vx) \big)   +  \\&\frac{1 }{K|\alpha|}\sum_{k=1}^K \sum_{\alpha=0}^1 p_{G(\alpha,k)} (\vx)  \min\big(0, -1-\gD(\vx )\big) d\vx \Big).
         \end{aligned}
    \end{align}
For a function $F(y)=a \big[ \min\big( 0,-1+ y \big)  \big]+ b\big[ \min\big(0, -1-y\big) \big]$ where $a,b\ge 0$, the maximizer of $F$ on $[0,+\infty]$ exists and is  reached in $y^*=1$ if $b\le a$, and is reached in $y^*=0$ else (the maximizer may not be unique). 

Setting $a=  p_X(\vx)$ and $b=\frac{1}{K|\alpha|}\sum_{k=1}^K \sum_{\alpha=0}^1 p_{G(\alpha,k)} (\vx)$, $V(\gD,G)$ reaches its maximum in $1$ if $\frac{1}{K|\alpha|}\sum_{k=1}^K\sum_{\alpha=0}^1 p_{G(\alpha,k)}(\vx)  \le p_X (\vx) $ (i.e., $\E_{\alpha,k}[p_{G(\alpha,k)} ] (\vx) \le  p_X (\vx)$ )  and otherwise in $0$. This completes the proof of the first part of the lemma.

\bigskip

Recall $(\gD^*, (G^*,R^*))$ is a Nash equilibrium. 
Plugging the value of $\gD^*$ back into $V(\cdot,G^*)$, we get
\begin{align*}
    &V(\gD^*,G^*)\nonumber\\
    &=
     \begin{aligned}[t]
    \frac{\lambda_{gan}}{K|\alpha|}\bigg(\int  &K|\alpha| \cdot p_X(\vx) \big[   \min \big( 0,-1 + \gD^*(\vx) \big)  \big] \\&+  \sum_{k=1}^K \sum_{\alpha=0}^1 p_{G^*(\alpha,k)} (\vx)\big[  \min\big(0, -1-\gD^*(\vx )\big) \big] d\vx \bigg)
    \end{aligned}
    \nonumber\\
    &=-\frac{\lambda_{gan}}{K|\alpha|}\bigg(\int   2\mathbb{I}\big\{\E_{\alpha,k}[p_{G^*(\alpha,k)} ](\vx) \le  p_X(\vx)  \big\} \sum_{k=1}^K \sum_{\alpha=0}^1 p_{G^*(\alpha,k)} (\vx)  \nonumber\\
    &\quad + \mathbb{I}\big\{\E_{\alpha,k}[p_{G^*(\alpha,k)} ](\vx) >  p_X(\vx)  \big\}  \big( K|\alpha| \cdot p_X(\vx) +  \sum_{k=1}^K \sum_{\alpha=0}^1 p_{G^*(\alpha,k)} (\vx)\big) d\vx \bigg)\nonumber\\
    &=
    \begin{aligned}[t]
    - \lambda_{gan} \bigg(1&+ \int  \mathbb{I}\big\{\E_{\alpha,k}[p_{G^*(\alpha,k)} ](\vx) \le  p_X(\vx)  \big\} \frac{1}{K|\alpha|}\sum_{k=1}^K \sum_{\alpha=0}^1 p_{G^*(\alpha,k)} (\vx)  \\&+\mathbb{I}\big\{\E_{\alpha,k}[p_{G^*(\alpha,k)} ](\vx) >  p_X(\vx)  \big\} p_X(\vx) d\vx \bigg)
    \end{aligned}
    \nonumber\\
    &=- 
    \begin{aligned}[t]
    \lambda_{gan} \Big(1&+ \int   \mathbb{I}\big\{\E_{\alpha,k}[p_{G^*(\alpha,k)} ](\vx) \le  p_X(\vx)  \big\} \E_{\alpha,k}[p_{G^*(\alpha,k)} ](\vx)\\  &+ \mathbb{I}\big\{\E_{\alpha,k}[p_{G^*(\alpha,k)} ](\vx) >  p_X(\vx)  \big\}  p_X(\vx) d\vx \Big),
    \end{aligned}
\end{align*}
which implies $-\lambda_{gan} \ge V(\gD^*,G^*) \ge -2\lambda_{gan}$.

Since we already showed $V(\gD,G)$ reaches its maximum in $1$ or $0$, it follows $\gD^*(\vx)\le 1$. Combining this observation with (\ref{eq:V}), we get
\begin{align*}
     V(\gD^*,G^*) 
        &= - \frac{\lambda_{gan}}{K|\alpha|}\sum_{k=1}^K \sum_{\alpha=0}^1\Big(\!\int  p_X(\vx)  (1- \gD^*(\vx))  +   p_{G^*(\alpha,k)} (\vx)( 1+\gD^*(\vx)) d\vx \Big) \nonumber\\
        &=- {\lambda_{gan}}  \Big(2+ \big(\int  \E_{\alpha,k} [p_{G^*(\alpha,k)}] (\vx) \gD^*(\vx ) -  p_X(\vx)  \gD^*(\vx)  d\vx \big)\Big).
\end{align*}
This completes the proof of the lemma.
\end{proof}

\bigskip

\paragraph{Proof of Theorem \ref{thm:nash}}
\begin{proof}[Proof of Theorem \ref{thm:nash}]
Note that we assume $G, R$ is given enough capacity. Let $\tilde{G}$ be the ``ideal" generator such that $p_X=p_{\tilde{G}(\alpha,k)}$, $\vx=\tilde{G}(\vx,\phi(\vx),k)$. 
Let $R^*$ be the ideal concept classifier that can predict the correct concept, then $L_{con}(G,R^*)=0$ for any $G$. 

\bigskip

\noindent\textbf{Case 1.} Let $\lambda_\phi=\lambda_{rec}=\lambda_{con}=0$; i.e., we only consider the problem
\[ \min_{G,R}\max_{\gD} \E_{\vx\sim p_X}\E_{\alpha,k}\big[\lambda_{gan} \big( L_\gD(\gD )+ L_G(G)\big)\big]. \]
Then, we are going to show $\E_{\alpha,k}[p_{G^*(\alpha,k)}]=p_X$ and $V(\gD^*,G^*)= -2\lambda_{gan}$.
From the Nash equilibrium, we have $  \E_{\vx\sim p_X}\E_{\alpha,k}\big[ L_G(G^*) \big]\le  \E_{\vx\sim p_X}\E_{\alpha,k}\big[  L_G(\tilde{G})\big]$. Note we shown that any maximizer $D^*(\vx)\le 1$, it follows 
\begin{align*}
    \frac{1}{K|\alpha|}\sum_{k,\alpha}\int   p_X(\vx)\gD^*(\tilde{G}(\vx,\alpha,k))  - p_X(\vx)\gD^*(G^*(\vx,\alpha,k))d\vx \le 0.
\end{align*}
Note $p_{\tilde{G}(\alpha,k)}=p_X$.  
By using integral transform, we have
\begin{align*}
&\frac{1}{K|\alpha|}\sum_{k,\alpha}\int    p_X(\vx) \gD^*(\vx)-  p_{G^*(\alpha,k)} \gD^*(\vx) d\vx \\&= \int  p_X(\vx) \gD^*(x)-  \E_{\alpha, k} [p_{G^*(\alpha,k)}](\vx) \gD^*(\vx)d\vx \le 0 .
\end{align*}
Plugging the above results back into (\ref{eq:V*}) in Lemma~\ref{lem:V} yields
\[ V(\gD^*,G^*) \le -2\lambda_{gan}. \]
Since (\ref{eq:Vgan}) implies $V(\gD^*,G^*) \ge -2\lambda_{gan}$, then it holds $V(\gD^*,G^*)= -2\lambda_{gan}$. Further, $ p_X =  \E_{\alpha, k} [p_{G^*(\alpha,k)}]$.

\bigskip

\noindent\textbf{Case 2.} If $\lambda_{\phi}=0$, then the optimization problem (\ref{eq:minmax}) becomes
\begin{align*}
\min_{G,R}\max_\gD \E_{\vx\sim p_X}\E_{\alpha,k}\big[ &\lambda_{gan}\big(L_\gD(\gD)+L_G(G)\big) + \lambda_{rec}(L_{rec}(G) + L_{cyc}(G)) + \\ &\lambda_{con} L_{con}(G,R)\big].
\end{align*}
From $U(\gD^*,G^*,R^*)\le U(\gD^*,\tilde{G},R^*)$, $\gD^*(\vx)\le 1$ and $L_{con}(G,R^*)=0$ for any $G$, we know
\begin{align*}
   & \frac{1}{K|\alpha|}\sum_{k,\alpha} \int    p_X(\vx) \lambda_{gan}\big( \gD^*(\tilde{G}(\vx,\alpha,k))-\gD^*( {G}^*(\vx,\alpha,k) \big) d\vx \nonumber\\
   &\le  \frac{1}{K|\alpha|}\sum_{k,\alpha} \int    p_X(\vx)\big(  \lambda_{rec} \big( \|  \vx- \tilde{G}( \vx,\phi(\vx),k) \|_1
   - \|  \vx- G^*( \vx,\phi(\vx),k) \|_1\\&\quad
   + \|  \vx- \tilde{G}(\tilde{G}(\vx,\alpha,k),\phi(\vx),k) \|_1
   -  \|  \vx- G^*(G^*(\vx,\alpha,k),\phi( \vx),k) \|_1  \big)\big)d\vx\nonumber\\
   &\le \frac{1}{K|\alpha|}\sum_{k,\alpha} \int    p_X(\vx)  \lambda_{rec}    \|  \vx- \tilde{G}(\tilde{G}(\vx,\alpha,k),\phi( \vx),k) \|_1 d\vx. 
\end{align*}
Since we assume the ideal generator $\tilde{G}$ is $\beta_{d}$-Lipschitz w.r.t the second argument with $\beta_{d}< \infty$ and $\phi$ is $\beta_\phi$-Lipschitz, it follows
\begin{align*}
    &\;\quad\big\|\vx-\tilde{G}\big(\tilde{G}(\vx,\alpha,k),\phi(\vx),k\big)\big\|_1
    \\&=\big\|\tilde{G}\big(\tilde{G}(\vx,\alpha,k),\phi(\tilde{G}(\vx,\alpha,k)),k\big)-\tilde{G}\big(\tilde{G}(\vx,\alpha,k),\phi(\vx),k\big)\big\|_1\nonumber\\
    &\le \beta_d \big| \phi(\tilde{G}(\vx,\alpha,k)) -\phi(\vx)  \big|\le \beta_d \beta_{\phi} \big\| \tilde{G}(\vx,\alpha,k) - \vx \big\|_1\nonumber\\&=\beta_d \beta_{\phi} \big\| \tilde{G}(\vx,\alpha,k) - \tilde{G}(\vx,\phi(x),k) \big\|_1\le  \beta_d^2  \beta_{\phi} \big|\phi(\vx) - \alpha\big|\le  \beta_d^2  \beta_{\phi}.
\end{align*}
Combining the above two observations together, we get
\begin{align*}
   \frac{1}{K|\alpha|}\sum_{k,\alpha} \int    p_X(\vx) \lambda_{gan}\big( \gD^*(\tilde{G}(\vx,\alpha,k))-\gD^*( {G}^*(\vx,\alpha,k) \big) d\vx  \le   \lambda_{rec}   \beta_d^2  \beta_{\phi}. 
\end{align*}
Note $p_{\tilde{G}_{\alpha,k}}=p_X$. By using integral transform, we have
\begin{align*}
   \frac{\lambda_{gan}}{K|\alpha|}\sum_{k,\alpha} \int    p_X(\vx)  \gD^*(\vx)-p_{G^*(\alpha,k)}(\vx)\gD^*( \vx ) d\vx  \le   \lambda_{rec}   \beta_d^2  \beta_{\phi}. 
\end{align*}
That is 
\begin{align*}
    \int    p_X(\vx)  \gD^*(\vx)- \E_{\alpha,k}[p_{G^*(\alpha,k)}](\vx)\gD^*( \vx ) d\vx  \le   \frac{\lambda_{rec}}{\lambda_{gan}}   \beta_d^2  \beta_{\phi}:=\epsilon. 
\end{align*}
Plugging this observation with (\ref{eq:V*}) in Lemma~\ref{lem:V}, we know
\begin{align*}
   V(\gD^*,G^*) &=-\lambda_{gan}(2+  \int   \E_{\alpha,k}[p_{G^*(\alpha,k)}](\vx)\gD^*( \vx )-  p_X(\vx)  \gD^*(\vx) d\vx ) \\&\le -\lambda_{gan}(2-\epsilon).
\end{align*} 
Combining this observation with (\ref{eq:Vgan})  in Lemma~\ref{lem:V}, we know
 $V(\gD^*,G^*)\in[-2\lambda_{gan}, -\lambda_{gan}(2-\epsilon)  ]$ and
 \[1-\epsilon \le \int \min\big\{ p_X(\vx), \E_{\alpha,k}[p_{G^*(\alpha,k)}](\vx) \big\}d\vx \le 1. \]
If $\phi$ is nearly flat with $\beta_\phi\approx 0$ such that $\epsilon\approx 0$, then it holds 
\[ V(\gD^*,G^*) \approx -2\lambda_{gan}, \]
and further
\[ \E_{\alpha,k}[p_{G^*(\alpha,k)}]\approx p_X .\]
If $\beta_\phi=0$, then 
\[ V(\gD^*,G^*) = -2\lambda_{gan}  \text{ and } \E_{\alpha,k}[p_{G^*(\alpha,k)}]= p_X .\]
The proof is completed. 
\end{proof}

\bigskip

\paragraph{Proof of Theorem \ref{thm:general}}
\begin{proof}[Proof of Theorem \ref{thm:general}]
For a Nash equilibrium $(\gD^*,(G^*,R^*))$, we are going to show that $\E_{\alpha,k}[p_{G^*(\alpha,k)}]= p_X$ and $V(\gD^*,G^*)= -2\lambda_{gan}$ cause a contradiction. Then $p_{G^*(\alpha,k)}\neq p_X$ and $V(\gD^*,G^*)\neq -2\lambda_{gan}$ for a Nash equilibrium.

Let $R^*$ be the ideal classifier with $L_{con}(G,R^*)=0$ for any $G$.
Let $G'$ be the generator satisfying $\phi(G'(\vx,\alpha,k))=\alpha$, $\|G'(\vx,\alpha_1,k)-G'(\vx,\alpha_2,k)\|_1\le \beta_1|\alpha_1-\alpha_2|$ and $\|G'(\vx,\alpha,k)-G'(\vx',\alpha,k)\|_1\le \beta_2\|\vx-\vx'\|_1$.   
We will show $U(\gD^*,(G^*,R^*))> U(\gD^*,(G',R^*))$ which violates the definition of a Nash equilibrium. 

From Lemma~\ref{lem:V}, we know $|\gD^*(\vx)|\le 1$. Therefore, 
\begin{align*}
\E_{\vx\sim p_X}\E_{\alpha,k}\big[ L_G(G') - L_G(G^*) \big]=   \int \big(p_X(\vx)-p_{G'(\alpha,k)}(\vx)\big)\gD^*(\vx)d\vx \in [-1,1].
\end{align*}
From the property of $G'$, it follows
\begin{align*}
&\E_{\vx\sim p_X}\E_{\alpha,k}\big[ L_{cyc}(G') -L_{cyc}(G^*)\big]\\&=   \int p_X(\vx) \big(\|\vx-G'(\vx,\phi(\vx),k)\|_1 -  \|\vx-G^*(\vx,\phi(\vx),k)\|_1\big) \le 0.
\end{align*}
Since $\phi(G'(\vx,\alpha,k))=\alpha$, it follows $f(\alpha)=\alpha\log(\phi(G'(\vx,\alpha,k)))+(1-\alpha)\log(1-\phi(G'(\vx,\alpha,k)))=0$ for $\alpha=0$ and $\alpha=1$ for any $k$. Hence, there exists a constant $0\le C<\infty$ such that 
\[ \E_{\vx\sim p_X}\E_{\alpha,k}\big[L_{\phi}(G')\big]\ge -C .\]
Note $p_{G^*(\alpha,k)}=p_X$. We know
\begin{align*}
  &\E_{\vx\sim p_X}\E_{\alpha,k}\big[\!  L_\phi(G^*)\big]\\&=\frac{1}{K|\alpha|}\sum_{k,\alpha}\int p_X(\vx)\big( \alpha\log( \phi(G^*(\vx,\alpha,k)) )+ (1-\alpha)\log(1-\phi(G^*(\vx,\alpha,k))) \big)d\vx\nonumber\\
    &=\frac{1}{K|\alpha|}\sum_{k,\alpha}\int p_X(\vx)\big(\alpha \log( \phi(\vx) )+ (1-\alpha)\log(1-\phi(\vx)) \big)d\vx.
\end{align*}
There exists a constant $M<\infty$ such $\E_{\vx\sim p_X}\E_{\alpha,k}\big[L_\phi(G^*)\big]\ge -M$ only if $\text{Prob}\big(\phi(\vx)=1 \cup \phi(\vx)=0\big)=0$ for every $\alpha$. That is, with zero probability, $\phi(\vx)=0$ or $\phi(\vx)=1$. However,  it violates the assumption of $\phi$. Hence, $L_\phi(G^*)=-\infty$. 

Combining the above two observations together, we know
\[ \E_{\vx\sim p_X}\E_{\alpha,k}\big[L_{\phi}(G^*) - L_\phi(G')\big] = -\infty. \]
Therefore,
\begin{align*}
 &\E_{\vx\sim p_X}\E_{\alpha,k}\big[  L_G(G') - L_G(G^*) +  L_{cyc}(G') -L_{cyc}(G^*) +  L_{\phi}(G^*) - L_\phi(G') \big] \\ &= -\infty,
\end{align*}
which implies
\[U(\gD^*,(G',R^*))< U(\gD^*,(G^*,R^*)).\]
This causes a contradiction from the fact that $U(\gD^*,(G^*,R^*))\le U(\gD^*,(G,R^*))$ for any $G$ since $(\gD^*,(G^*,R^*))$ is a Nash equilibrium of the system. Hence, $p_{G^*(\alpha,k)}\neq p_X$ and $V(\gD^*,G^*)\neq -2\lambda_{gan} $ is implied by (\ref{eq:Vgan}).  
This completes the proof.
\end{proof}

\section{Limitations} \label{appx:sec:limitations}
In the main paper, we demonstrated the efficacy of the proposed method to generate counterfactual explanations (CEs) for deep anomaly detection (AD). Here, we discuss a few limitations.
One challenge is that the ideal number of categorical concepts is problem-specific and cannot be automatically learned, though future work may solve this by comparing the models for different numbers of concepts. Additionally, the quality of the CEs may be affected when the AD model performs poorly. For example, DSVDD without OE \citep{ruff2018deep} tends to perform weakly on some image datasets, leading to less intuitive CEs that may collapse into mere reconstructions of the anomaly. This occurs when DSVDD struggles to detect anomalies, resulting in already low anomaly scores of anomalies and thus no clear incentive for the generator to transform the anomaly. Another challenge is that the generator may over-correct, making unnecessary changes and falling into a local optimum of the proposed objective. Balancing objectives in an unsupervised setting is challenging, especially given the limited diversity and quantity of normal samples. A potential improvement when using diffusion models is to explore prompt engineering or learning in an unsupervised manner, which could enhance the explainability of the CEs and address some of these challenges. Another direction for improvement would be leveraging advancements from the related field of adversarial examples \citep{mustafa2024non}.

\section{The CEs Capture Disentangled Concepts} \label{sec:cf_disentanglement}
Recall that the concept classifier predicts the concept of each CE (see Section \ref{sec:method}).
We present the accuracy of this concept classifier for the generated CEs in Table \ref{tbl:main_Concept_Acc}. 
Our models demonstrate a consistent ability to disentangle concepts effectively. 
Notably, disentanglement is effective even when just one class is considered normal. On CIFAR-10 the generator exploits the background, on INN the color scheme, and on MNIST it generates disentangled variants of digits. We hypothesize that this strong disentanglement is the reason for the CEs appearing less normal for MNIST.

\begin{table}[ht]
\centering
\caption{Concept classifier accuracy for the CEs.
}
\label{tbl:main_Concept_Acc}
\setlength\tabcolsep{3pt}
\renewcommand{\arraystretch}{1}
\begin{tabular}{@{}l|lccc@{}}
\toprule
&\multirow{2.5}{*}{Datasets} & \multicolumn{3}{c}{Methods}\\
\cmidrule{3-5}
& & BCE OE & HSC OE & DSVDD\\
\midrule
\multirow{3}{3.2em}{Single normal class} & MNIST & 94.3 $\pm$ 3.9 & 90.8 $\pm$ 4.8 & 77.5 $\pm$ 14.1 \\
                                        & CIFAR-10 & 93.0 $\pm$ 4.3 & 98.8 $\pm$ 3.2 & 97.1 $\pm$ 2.9 \\
                                        & INN & 97.0 $\pm$ 5.4 &   98.9 $\pm$ 1.1 & $\times$ \\\midrule
\multirow{4}{3.2em}{Multiple normal classes}& C-MNIST & 99.4 $\pm$ 1.3 & 98.9 $\pm$ 2.0 & 98.0 $\pm$ 3.0 \\
                                        & MNIST & 93.8 $\pm$ 5.1 & 85.7 $\pm$ 9.6 & 81.6 $\pm$ 11.3 \\
                                        & CIFAR-10 & 86.2 $\pm$ 7.5 & 98.9 $\pm$ 2.4 & 92.2 $\pm$ 4.2 \\
                                        & GTDSB & 98.8 $\pm$ 0.8 & 94.0 $\pm$ 8.4 & 93.4 $\pm$ 4.5 \\
\bottomrule
\end{tabular}
\end{table}

\section{The CE Framework Performs Minimal Edits} \label{appx:sec:xad:min_edits}
The main paper presented several qualitative examples of CEs, demonstrating that the modifications relative to the original anomalies are both meaningful and minimal. 
To quantitatively assess the minimality of these changes, we report the average MSE between test anomalies and their corresponding CEs in Table \ref{tbl:appx_MSE}, as well as the average LPIPS \citep{zhang2018unreasonable} in Table \ref{tbl:appx_LPIPS}. 
The results indicate that the modifications are reasonably limited, as both metrics attain comparatively low values.
In particular, the LPIPS scores---ranging between 0 and 1---typically fall between $0.15$ and $0.5$, which we consider to be a reasonable range for perceptually minimal changes.

\begin{table}[ht]
\centering
\caption{Mean squared error (MSE) between test anomalies and their CEs. Results are averaged over 4 random seeds and up to 20 normal definitions (see Appendix \ref{appx:sec:full_quantitative_results}). }
\label{tbl:appx_MSE}
\begin{adjustbox}{max width=.99\textwidth}
\setlength\tabcolsep{5pt}
\begin{tabular}{lcccccc}
\toprule
 & \multicolumn{2}{|c|}{Single normal classes} & \multicolumn{4}{|c|}{Multiple normal classes} \\
Method & MNIST & CIFAR-10 & C-MNIST & MNIST & CIFAR-10 & GTSDB \\
\midrule
BCE OE & 0.06 $\pm$ 0.019 & 0.02 $\pm$ 0.001 & 0.06 $\pm$ 0.008 & 0.03 $\pm$ 0.007 & 0.01 $\pm$ 0.002 & 0.08 $\pm$ 0.011 \\
HSC OE & 0.05 $\pm$ 0.016 & 0.03 $\pm$ 0.006 & 0.07 $\pm$ 0.006 & 0.02 $\pm$ 0.006 & 0.03 $\pm$ 0.004 & 0.07 $\pm$ 0.011 \\
DSVDD & 0.03 $\pm$ 0.018 & 0.02 $\pm$ 0.003 & 0.05 $\pm$ 0.007 & 0.02 $\pm$ 0.008 & 0.01 $\pm$ 0.001 & 0.06 $\pm$ 0.017 \\
\bottomrule
\end{tabular}
\end{adjustbox}
\end{table}

\begin{table}[ht]
\centering
\caption{Learned Perceptual Image Patch Similarity (LPIPS) \citep{zhang2018unreasonable} between test anomalies and their CEs. Results are averaged over 4 random seeds and up to 20 normal definitions (see Appendix \ref{appx:sec:full_quantitative_results}). }
\label{tbl:appx_LPIPS}
\begin{adjustbox}{max width=.99\textwidth}
\setlength\tabcolsep{5pt}
\begin{tabular}{lcccccc}
\toprule
 & \multicolumn{2}{|c|}{Single normal classes} & \multicolumn{4}{|c|}{Multiple normal classes} \\
Method & MNIST & CIFAR-10 & C-MNIST & MNIST & CIFAR-10 & GTSDB \\
\midrule
BCE OE & 0.16 $\pm$ 0.041 & 0.43 $\pm$ 0.021 & 0.26 $\pm$ 0.029 & 0.10 $\pm$ 0.015 & 0.37 $\pm$ 0.023 & 0.52 $\pm$ 0.065 \\
HSC OE & 0.15 $\pm$ 0.037 & 0.60 $\pm$ 0.048 & 0.26 $\pm$ 0.023 & 0.08 $\pm$ 0.012 & 0.57 $\pm$ 0.023 & 0.53 $\pm$ 0.057 \\
DSVDD & 0.07 $\pm$ 0.035 & 0.42 $\pm$ 0.021 & 0.18 $\pm$ 0.025 & 0.05 $\pm$ 0.021 & 0.30 $\pm$ 0.020 & 0.51 $\pm$ 0.057 \\
\bottomrule
\end{tabular}
\end{adjustbox}
\end{table}

\section{The Detection Performance of the Detectors} \label{sec:ad_performance}
Table \ref{tbl:main_AD_AuROC} shows the AuROC of anomaly scores for normal test data vs.~ground-truth test anomalies. We consider the BCE, HSC, and DSVDD detector from the main paper. The values match the reported scores from the literature. 
\begin{table}[!htb]
    \centering
    \caption{The detection performance of the considered detectors in terms of AuROC of anomaly scores for normal vs.~anomalous test data.}
    \label{tbl:main_AD_AuROC}
    \setlength\tabcolsep{4pt}
    \renewcommand{\arraystretch}{1}
    \begin{tabular}{@{}l|lccc@{}}
    \toprule
    &\multirow{2.5}{*}{Datasets} & \multicolumn{3}{c}{Methods}\\
    \cmidrule{3-5}
    & & BCE OE & HSC OE & DSVDD\\
    \midrule
    \multirow{3}{3.2em}{Single normal class} & MNIST & 97.7 $\pm$ 1.5 & 97.6 $\pm$ 1.6 & 78.8 $\pm$ 8.6 \\
                                             & CIFAR-10 & 96.0 $\pm$ 2.5 & 95.9 $\pm$ 2.5 & 55.4 $\pm$ 4.7 \\
                                             & INN & 93.6 $\pm$ 5.7 & 92.6 $\pm$ 6.7  & $\times$ \\\midrule
    \multirow{4}{3.2em}{Multiple normal classes}& C-MNIST & 97.1 $\pm$ 1.0 & 95.7 $\pm$ 2.3 & 76.9 $\pm$ 6.5 \\
                                               & MNIST & 93.5 $\pm$ 2.8 & 92.9 $\pm$ 3.3 & 75.4 $\pm$ 7.1 \\
                                               & CIFAR-10 & 93.8 $\pm$ 2.7 & 94.0 $\pm$ 2.7 & 52.6 $\pm$ 3.6 \\
                                               & GTDSB & 94.3 $\pm$ 4.7 & 93.0 $\pm$ 5.6 & 58.2 $\pm$ 6.7 \\
    \bottomrule
    \end{tabular}
\end{table}

\section{Concept Figures} \label{appx:concept_figures}
\begin{figure}[htbp]
    \centering
    %
    \begin{subfigure}[b]{0.70\textwidth}
        \centering
        \includegraphics[width=\textwidth]{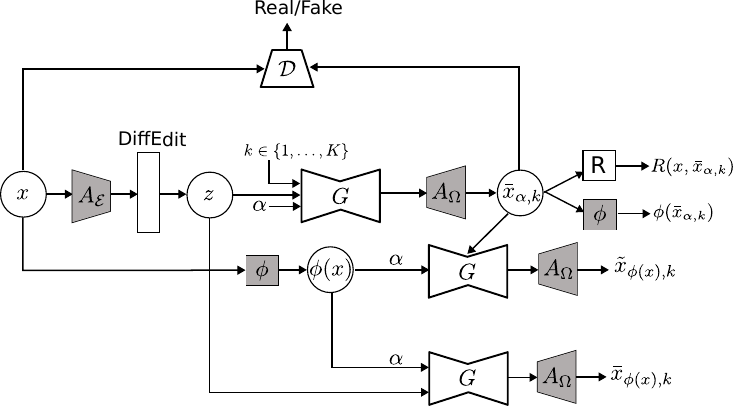}
        \vspace{1em}
        \caption{}
        \label{fig:concept_diffusion_disentangle_training}
    \end{subfigure}
    \hspace{3em}
    \begin{subfigure}[b]{0.18\textwidth}
        \centering
        \includegraphics[width=\textwidth]{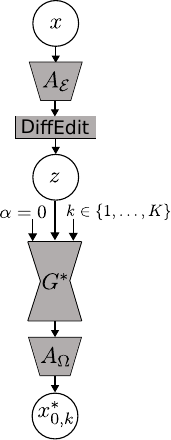}
        \caption{}
        \label{fig:concept_diffusion_disentangle_inference}
    \end{subfigure}
    \caption{Schematic overview of the proposed disentangled CE framework when using a pre-trained diffusion model. Gray nodes correspond to models that are not further optimized.}
    \label{fig:concept_diffusion_disentangle}
\end{figure}

In the main paper, we show a schematic overview of the proposed framework in Section~\ref{sec:method}.
Here, we provide a modification of this figure when using diffusion models as a backbone in Figure \ref{fig:concept_diffusion_disentangle}. Figure \ref{fig:concept_diffusion_disentangle_training} shows the training framework, which includes the encoder $A_\mathcal{\gE}$ and decoder $A_\Omega$ from stable diffusion \citep{rombach2022high}. We apply DiffEdit \citep{couairon2023diffedit} to the image in the latent space and feed the output $z$ to the generator, which is trained to generate disentangled explanations as before. The output of the generator is then passed to $A_\Omega$. The generator thus operates on the latent space of DiffEdit, while all other modules still operate on the full-resolution image space. The objectives to train the framework are described in Section~\ref{sec:ct_for_cf_on_ad}. Figure \ref{fig:concept_diffusion_disentangle_inference} shows the inference framework that utilizes the encoder and decoder from stable diffusion and our trained generator to produce CEs.

\section{Hyperparameter Sensitivity Analysis} \label{appx:sec:hp_sensivity_analysis}
\begin{figure}[h!] 
    \centering
    \begin{subfigure}[b]{0.32\textwidth}
        \includegraphics[width=\textwidth]{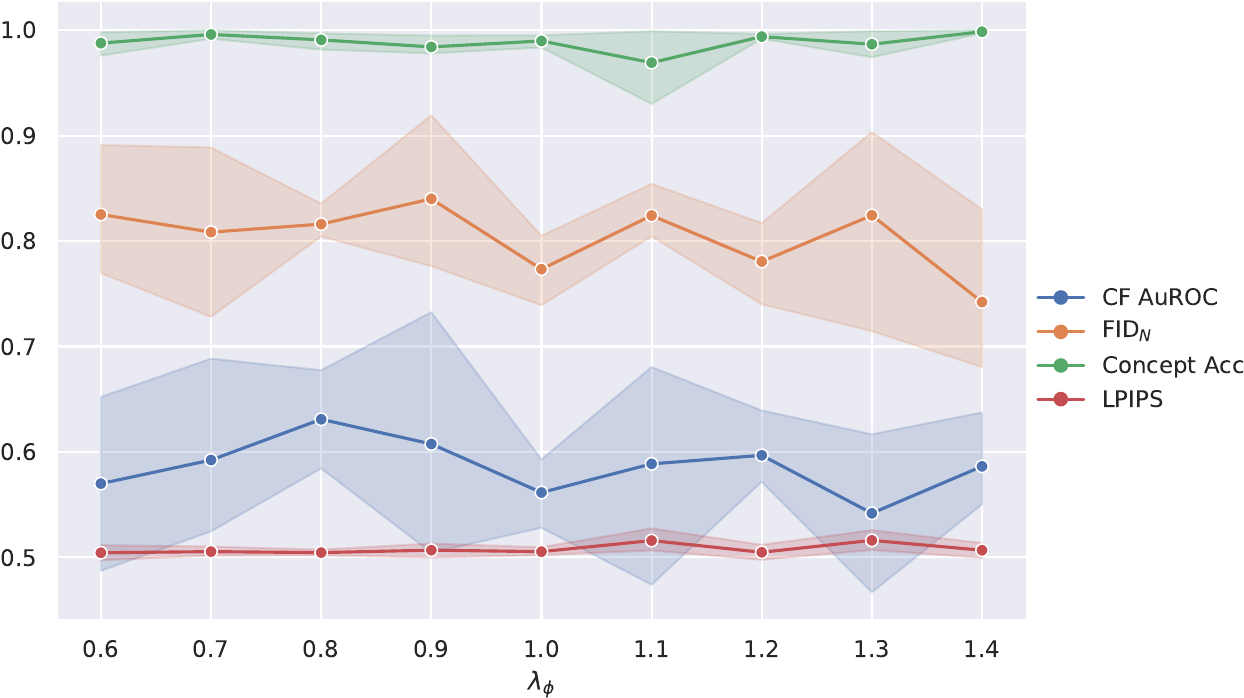}
        \caption{$\lambda_\phi$ for BCE}
    \end{subfigure}
    \hfill
    \begin{subfigure}[b]{0.32\textwidth}
        \includegraphics[width=\textwidth]{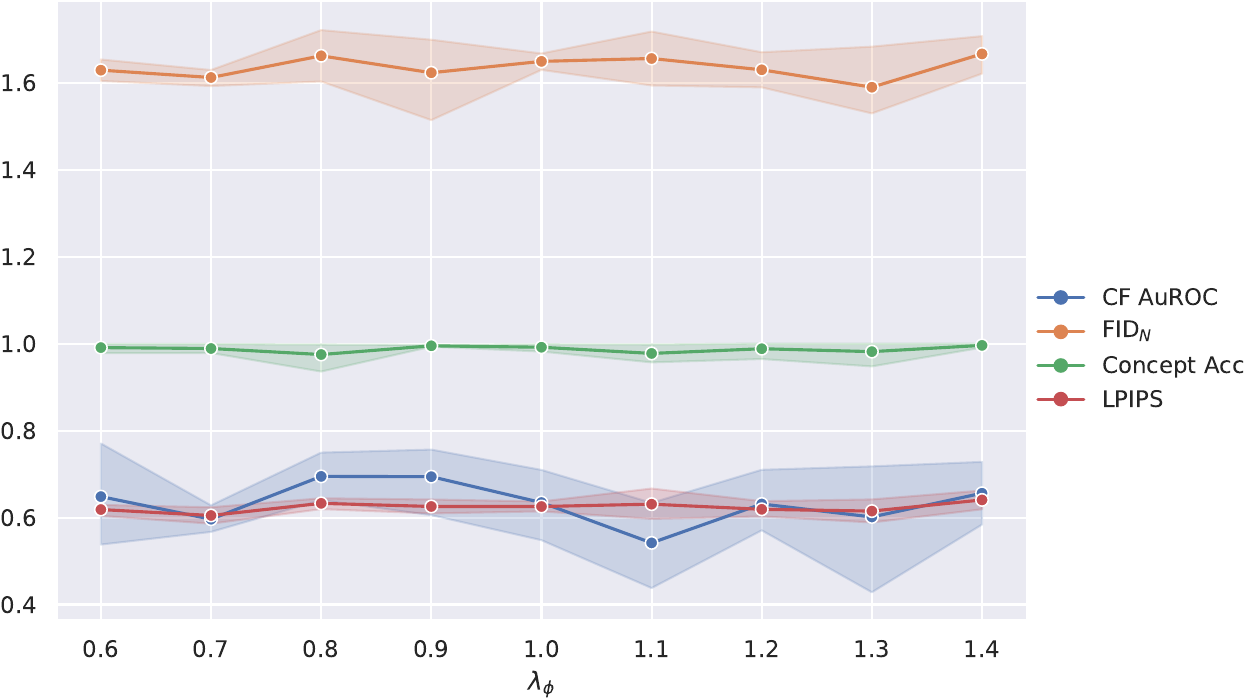}
        \caption{$\lambda_\phi$ for DSVDD}
    \end{subfigure}
    \hfill
    \begin{subfigure}[b]{0.32\textwidth}
        \includegraphics[width=\textwidth]{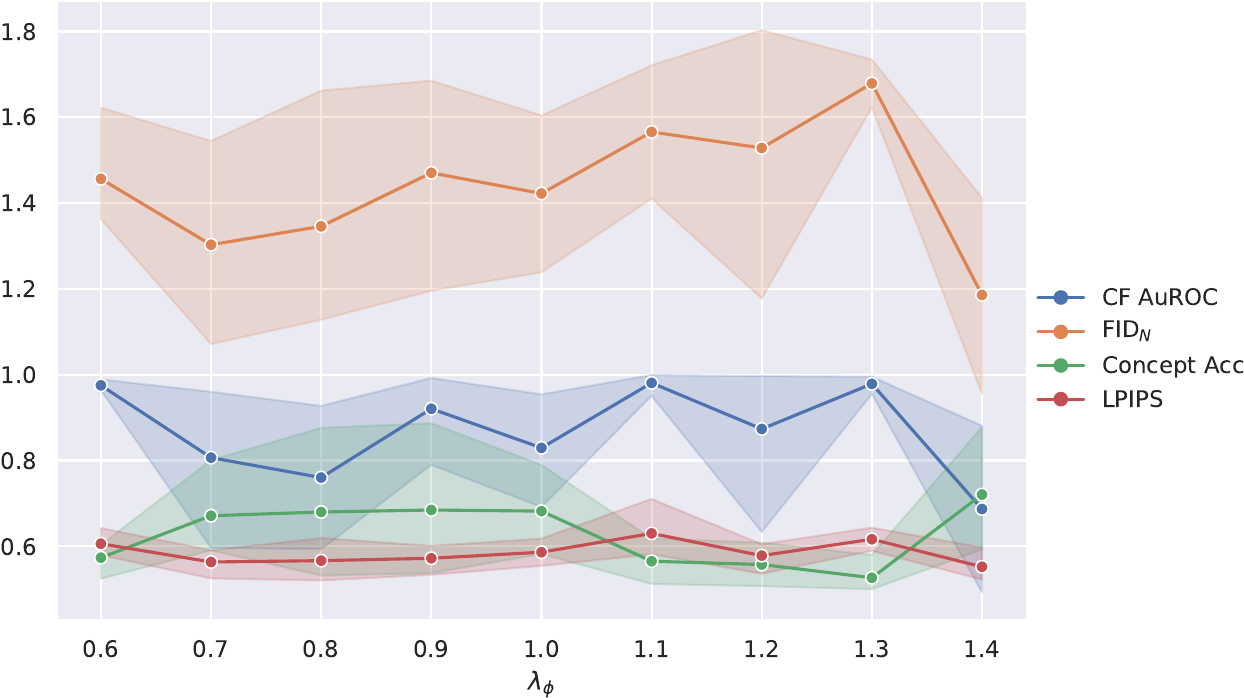}
        \caption{$\lambda_\phi$ for HSC}
    \end{subfigure}

    \vspace{0.3cm}

    \begin{subfigure}[b]{0.32\textwidth}
        \includegraphics[width=\textwidth]{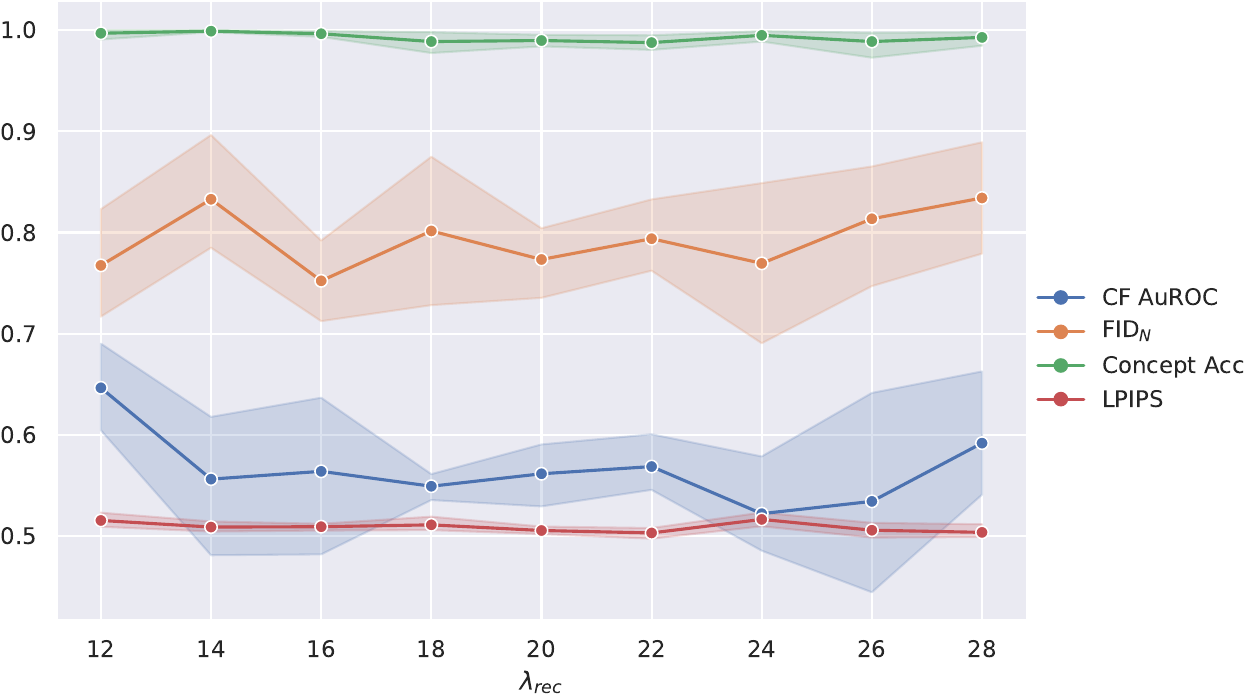}
        \caption{$\lambda_{rec}$ for BCE}
    \end{subfigure}
    \hfill
    \begin{subfigure}[b]{0.32\textwidth}
        \includegraphics[width=\textwidth]{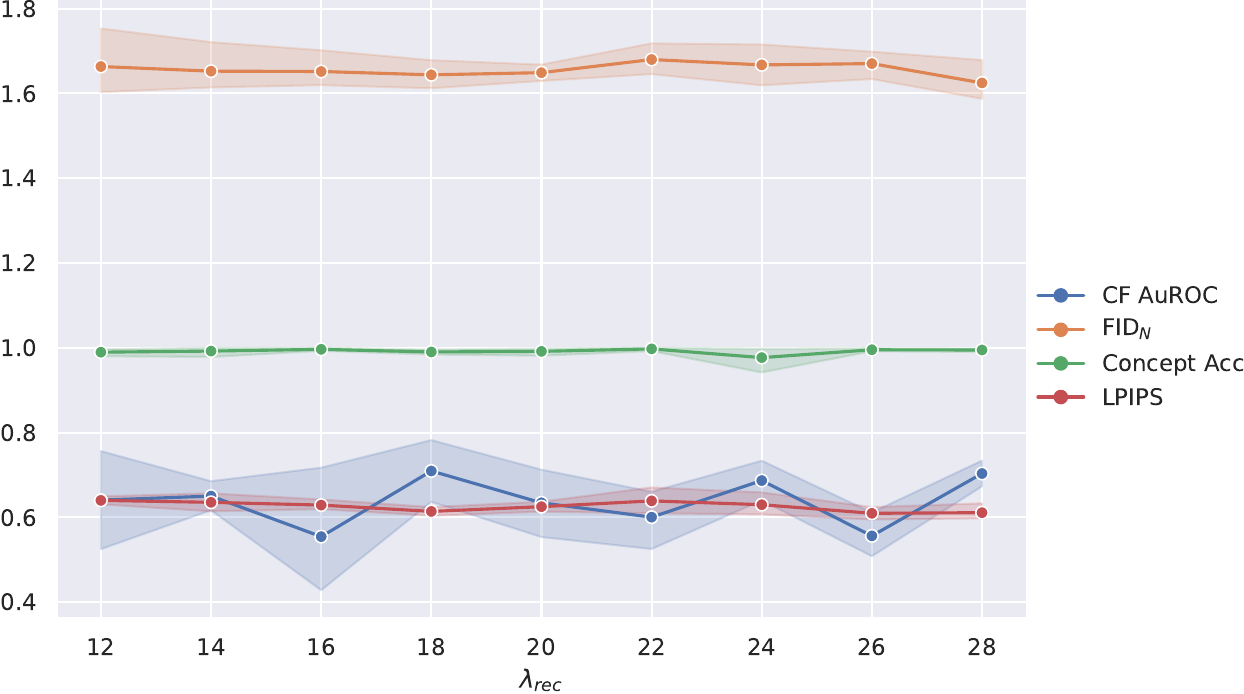}
        \caption{$\lambda_{rec}$ for DSVDD}
    \end{subfigure}
    \hfill
    \begin{subfigure}[b]{0.32\textwidth}
        \includegraphics[width=\textwidth]{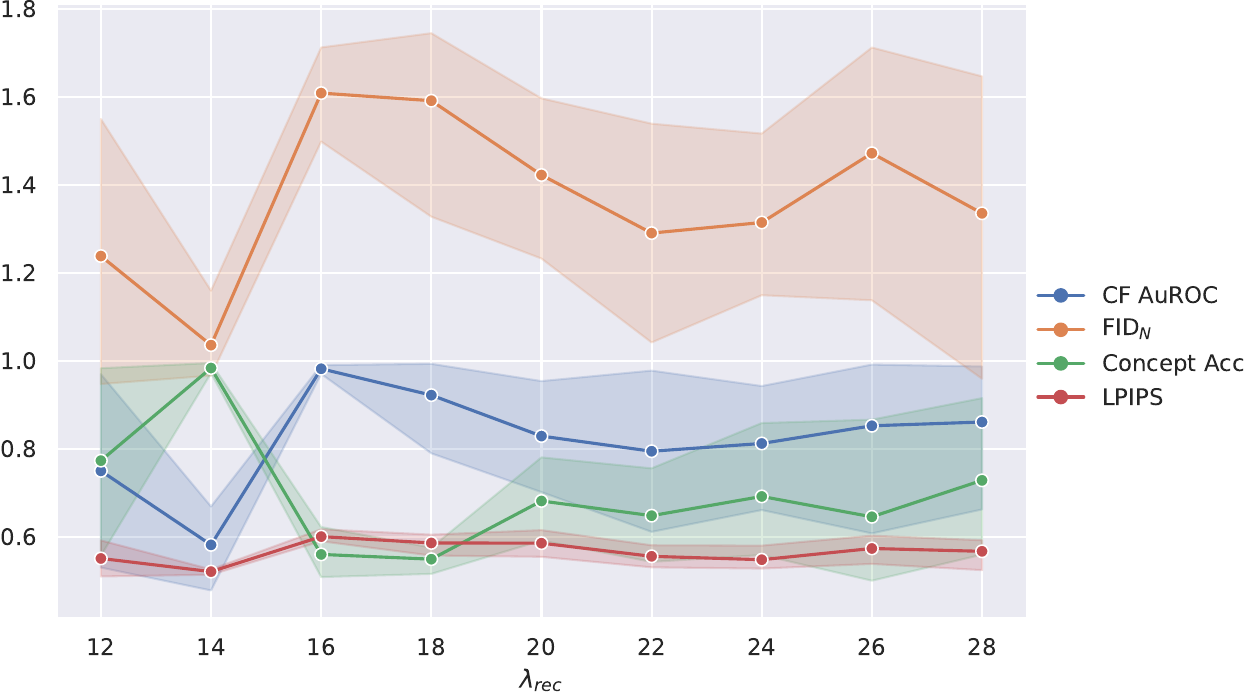}
        \caption{$\lambda_{rec}$ for HSC}
    \end{subfigure}

    \vspace{0.3cm}

    \begin{subfigure}[b]{0.32\textwidth}
        \includegraphics[width=\textwidth]{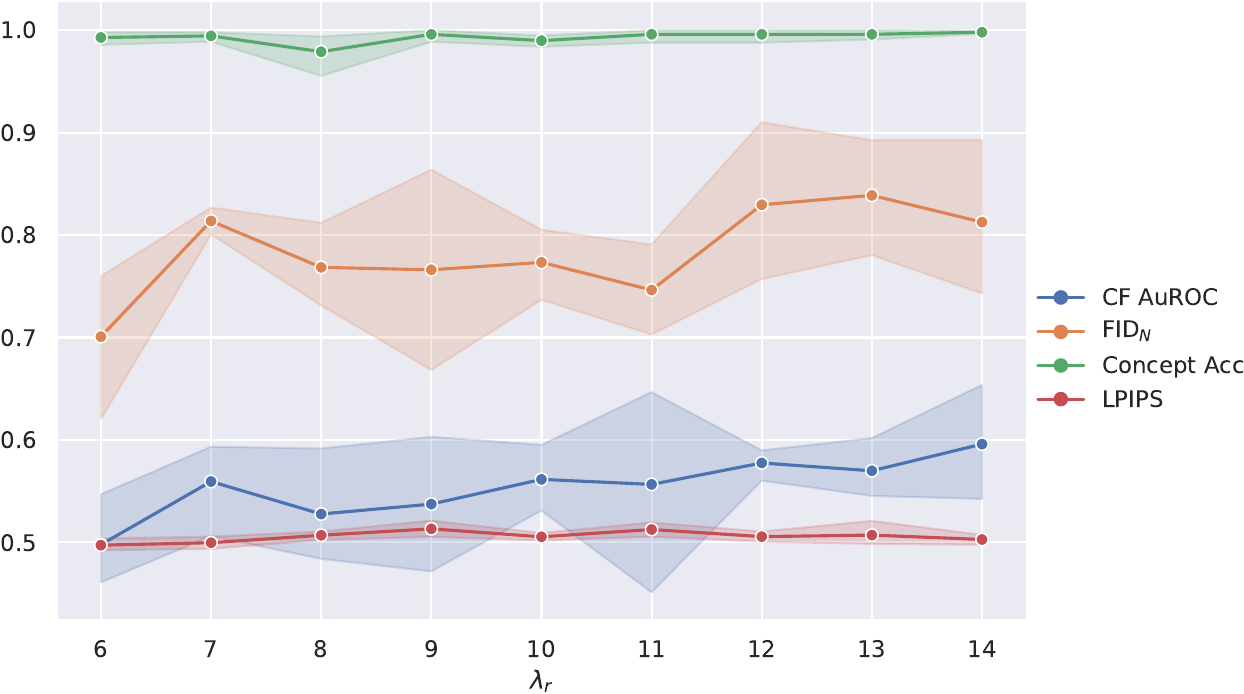}
        \caption{$\lambda_{r}$ for BCE}
    \end{subfigure}
    \hfill
    \begin{subfigure}[b]{0.32\textwidth}
        \includegraphics[width=\textwidth]{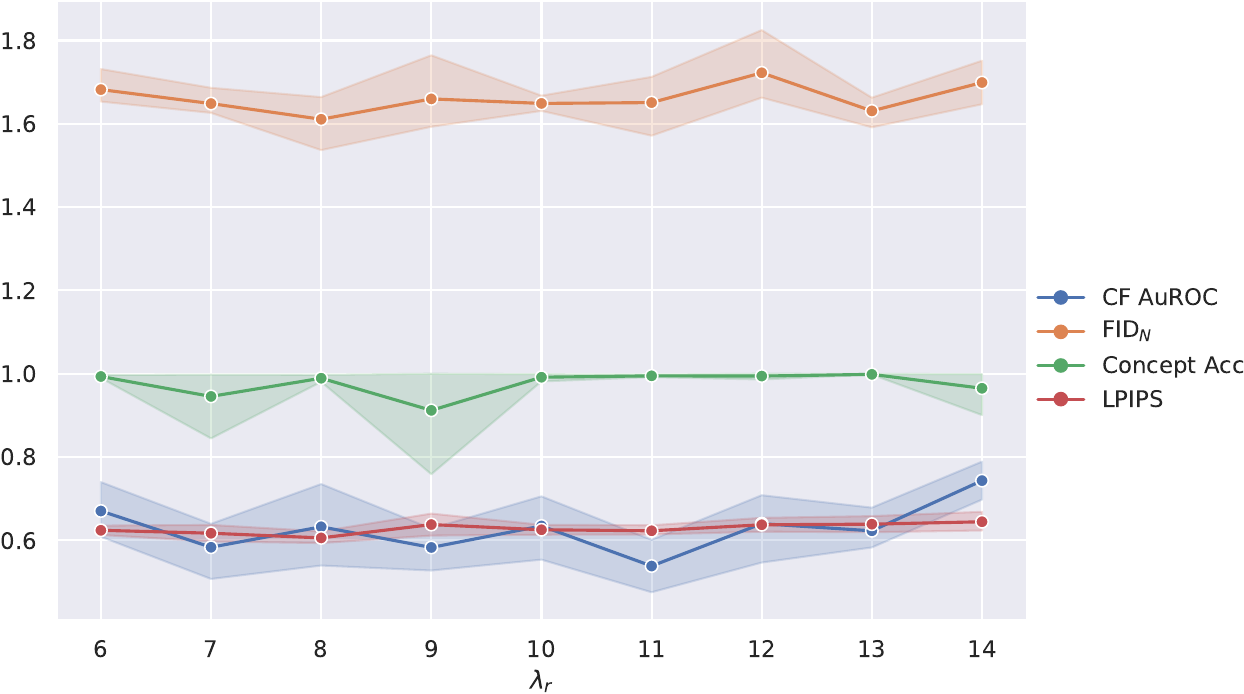}
        \caption{$\lambda_{r}$ for DSVDD}
    \end{subfigure}
    \hfill
    \begin{subfigure}[b]{0.32\textwidth}
        \includegraphics[width=\textwidth]{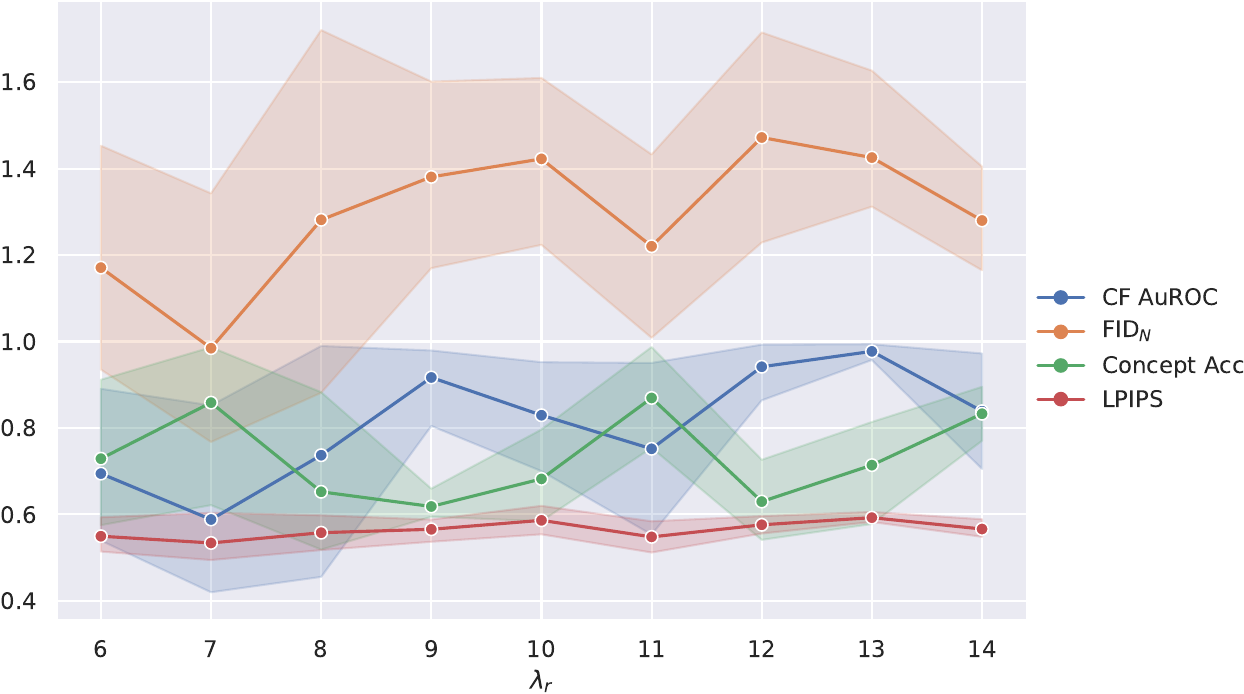}
        \caption{$\lambda_{r}$ for HSC}
    \end{subfigure}

    \vspace{0.3cm}

    \begin{subfigure}[b]{0.32\textwidth}
        \includegraphics[width=\textwidth]{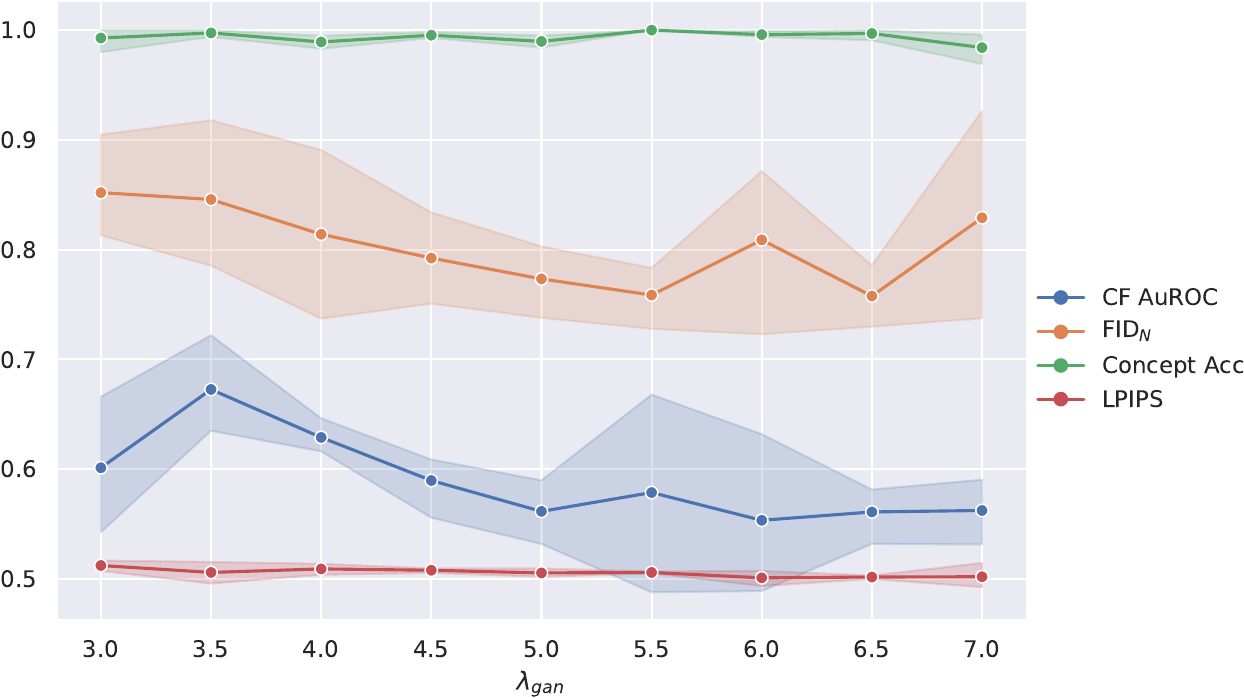}
        \caption{$\lambda_{gan}$ for BCE}
    \end{subfigure}
    \hfill
    \begin{subfigure}[b]{0.32\textwidth}
        \includegraphics[width=\textwidth]{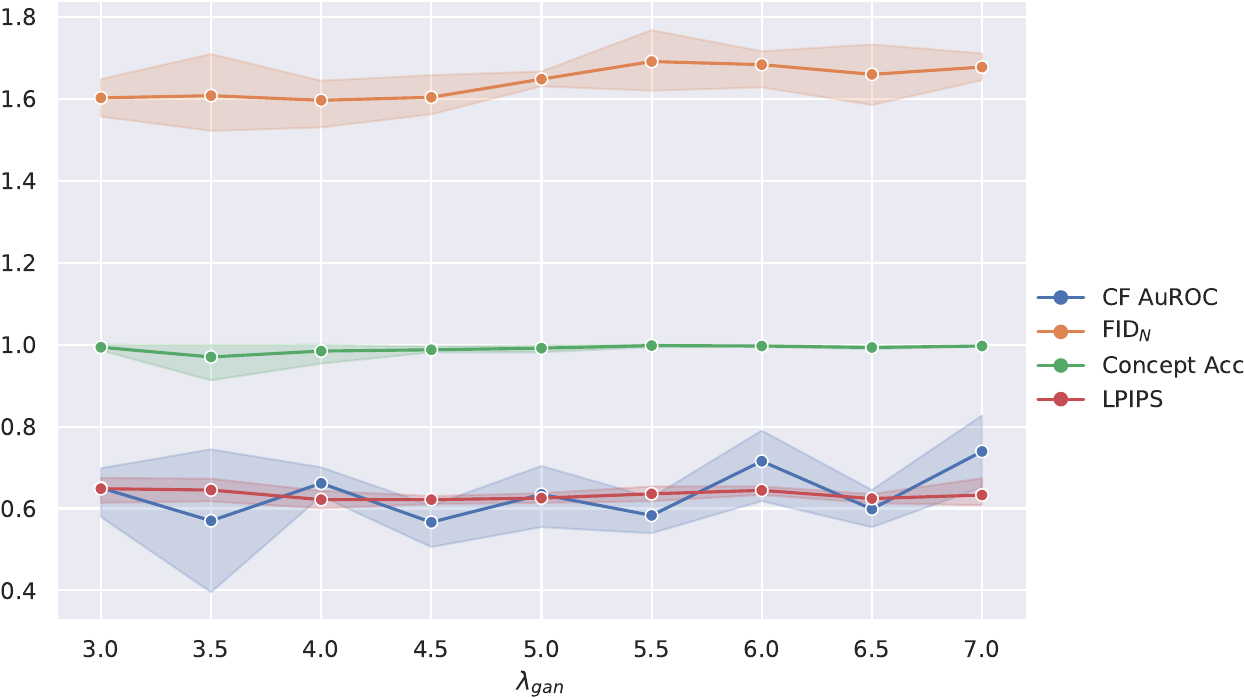}
        \caption{$\lambda_{gan}$ for DSVDD}
    \end{subfigure}
    \hfill
    \begin{subfigure}[b]{0.32\textwidth}
        \includegraphics[width=\textwidth]{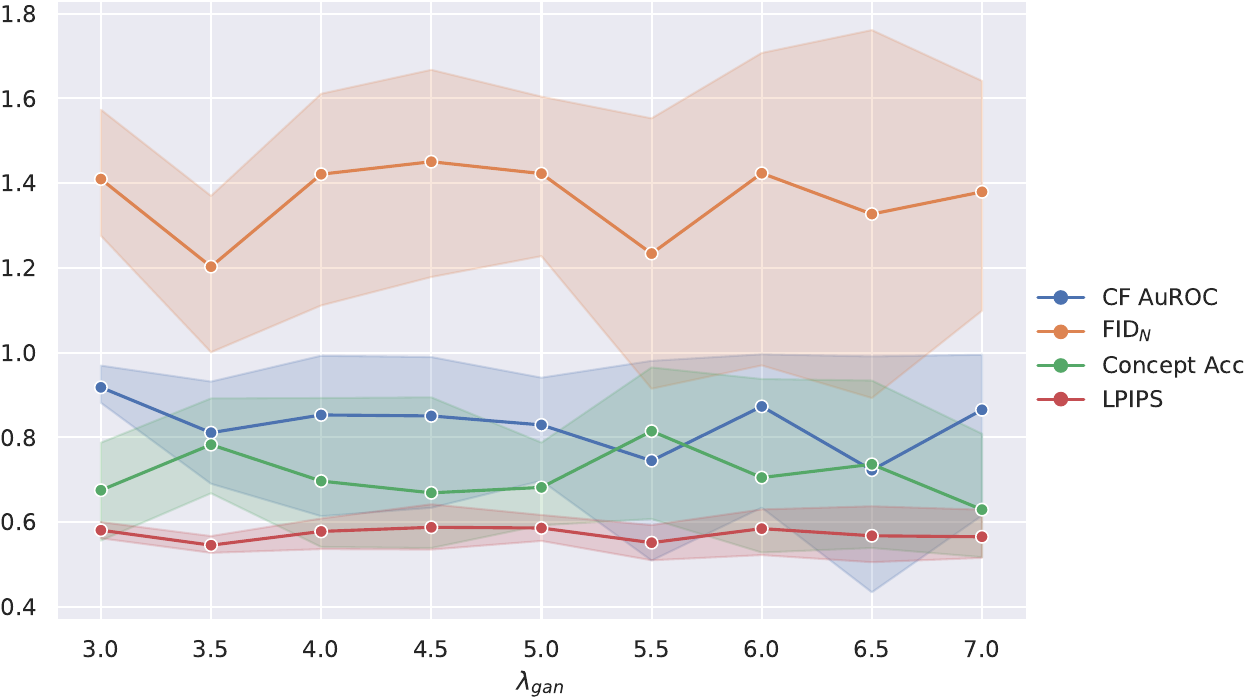}
        \caption{$\lambda_{gan}$ for HSC}
    \end{subfigure}

    \caption{The plots show the impact of the lambda hyperparameters that weigh the interacting loss functions of our counterfactual framework. In each plot, we vary a $\lambda$ between $60\%$ and $140\%$ of the value used in the experiments in the main paper (see Appendix \ref{appx:sec:hyperparameters}), while keeping the other $\lambda$ parameters fixed. The reported metrics are averaged over four seeds for the specific setting of speed signs being normal in GTSDB. The blue line reports the CF AuROC, the orange one the FID$_{N}$, the green the concept accuracy, and the red one the LPIPS metric.}
    \label{appx:fig:hyperparameters_gtsdb_speed_sign}
\end{figure}

The proposed counterfactual training objective consists of six interacting losses, which are weighted by factors $\lambda_{gan}$, $\lambda_{\phi}$, $\lambda_{rec}$, and $\lambda_{r}$ (see Problem~\ref{eq:minmax}). These factors are hyperparameters that must be set prior to training. For the experiments in the main paper, we selected reasonable values that worked across most settings (see Appendix \ref{appx:sec:hyperparameters}). Here, we perform a small hyperparameter sensitivity analysis, varying each $\lambda$ between $60\%$ and $140\%$ of the value used in the experiments while keeping the other $\lambda$ parameters fixed. The resulting evaluation metrics are reported in Figure~\ref{appx:fig:hyperparameters_gtsdb_speed_sign}.

\section{Deep Anomaly Detection Methods} \label{appx:deep_ad_methods_details}
This paper studies the application of the proposed method to three state-of-the-art anomaly detectors. Here, we provide a detailed description of those methods.

\paragraph{DSVDD} 
One of the first deep approaches to AD is Deep Support Vector Data Description (DSVDD) \citep{ruff2018deep}. 
Similar to many AD methods, DSVDD is unsupervised, employing an unlabeled corpus of (mostly normal) data for training \citep{ruff2018deep}. DSVDD trains a neural network $\phi_\theta: \R^D \rightarrow \R^d $ with parameters $\theta$ to map the training data $\vx_1,\dots,\vx_n \in \R^D$ into a semantic space $\R^d$, where it can be enclosed by a minimal volume hypersphere: $\min_\theta \sum_{i=1}^n ||\phi(\vx_i) - \vc||^2$. 
In contrast to shallow SVDD \citep{tax2004support}, the hypersphere center $\vc \in \R^d$ is first randomly initialized and then kept fixed while training. 
DSVDD trains the network to make normal data cluster tightly in the semantic space. Anomalies will have a larger distance from the center. 
The distance is used as the anomaly score. Since the CE generator requires bounded anomaly scores, we slightly adjust the DSVDD objective to: $
    \min_\theta \sum_{i=1}^n \frac{||\phi(\vx_i) - \vc||^2}{1 + ||\phi(\vx_i) - \vc||^2}.
$

\paragraph{Outlier Exposure}
AD has traditionally been approached as an unsupervised learning problem due to insufficient training data to represent the diverse anomaly class, which encompasses \emph{everything different} from the normal data. However, 
\citet{hendrycks2019deep} showed that \emph{Outlier Exposure} (OE)---using a large unstructured collection of natural images as example anomalies during training---consistently outperforms previous AD methods across various image-AD benchmarks, while still being unsupervised. The unlabeled auxiliary data are called OE samples. 
It has been found that training a Binary Cross Entropy (BCE) loss to differentiate normal data from OE samples is competitive for most image-AD tasks. 
To the best of our knowledge, OE with BCE still holds the state of the art in semantic  image anomaly detection benchmarks \citep{liznerski2022exposing}.
Since we use the same data for training $G$ as was used for training the detector $\phi$, we include the OE samples for training the generator $G$. The generator $G$ is thus trained on a more diverse training set, including additional presumably anomalous OE samples. 

\paragraph{Hypersphere Classification}
Although OE performs well in many benchmarks, there are still scenarios where OE samples do not adequately represent anomalies, especially when the normal data are not natural images \citep{liznerski2022exposing}. To address this problem, the community has developed \emph{semi-supervised} AD methods \citep{Grnitz2014TowardSA,ruff2020}. One of the most competitive semi-supervised AD techniques is \emph{HyperSphere Classification} (HSC) \citep{liznerski2022exposing}. The authors find that combining it with OE makes the AD more robust to the selection of OE data. The HSC loss is a semi-supervised modification of the DSVDD loss:
\begin{equation*}\label{eqn:radial}
\frac{1}{n} \sum_{i=1}^n y_i \cdot h\left(\phi(\vx_i)\right)- (1{-}y_i)\log{(1{-}\exp{({-}h\left(\phi(\vx_i)\right))})},
\end{equation*}
where $h$ is the Pseudo-Huber loss $h(\vz) = (\left\lVert\vz\right\rVert^2+1)^{1/2}-1$.
We employ HSC's original objective but modify the anomaly score from $h\left(\phi(\vx_i)\right)$ to 
$1 - \exp(-h\left(\phi(\vx_i)\right))$,
again obtaining bounded anomaly scores for training the proposed counterfactual generator.

\section{Datasets} \label{appx:sec:datasets}
In our experiments, we consider the following datasets:
\begin{itemize}[itemsep=0.5em,topsep=0.5em,leftmargin=1.5em]
    \item MNIST \citep{mnist} is a dataset of grayscale handwritten digits with a class for each digit. Following \citet{liznerski2021}, we use EMNIST \citep{emnist} as OE.
    \item Colored-MNIST, where each sample in MNIST is replicated in seven colors (red, yellow, green, cyan, blue, pink, and gray). We use a colored version of EMNIST as OE.
    \item CIFAR-10 \citep{krizhevsky2009learning} is a dataset of natural images with ten classes. Previous work used 80 Mio.~Tiny Images as OE \citep{hendrycks2019using}. Since this dataset has been withdrawn due to offensive data \citep{birhane2021large}, we instead use the disjunct CIFAR-100 dataset as OE, which yields approximately the same performance (reported in Table \ref{tbl:appx_cifar10_bce_single_normal_class}).
    \item GTSDB \citep{Houben2013gtsdb} is a dataset of German traffic signs. We use CIFAR-100 as OE.
    \item We introduce ImageNet-Neighbors (INN), a subset of ImageNet-1k~\citep{ILSVRC15} designed for AD tasks. INN comprises multiple AD setups; in each setup, one ImageNet-1k class is considered normal, and the ten most semantically similar classes, based on the Wu-Palmer similarity metric~\citep{wu1994verb}, are defined as ground-truth test anomalies. For outlier exposure (OE), we use the disjoint ImageNet-21k dataset.  
\end{itemize}

\section{Metrics} \label{appx:sec:metrics}
In this section, we provide details of the metrics used for the quantitative analysis in Section \ref{sec:quantitative_results}.

\paragraph{Normality of counterfactuals}
To assess the normality of the generated CEs, we computed the AuROC of normal test samples against CEs generated for all ground-truth anomalies from the test set. 
The Area Under the ROC curve (AuROC) is a widely recognized metric in the AD literature for comparing anomaly scores of normal and anomalous samples \citep{hanley1982meaning}.
An AuROC of $1$ indicates perfect separation between anomalies and normal samples, $0.5$ corresponds to random guessing, and a score below $0.5$ suggests that anomalies appear more normal than the actual normal samples. 
To assess the normality of our CEs, we computed the AuROC with the anomalies being CEs.
Then, an AuROC of significantly more than $0.5$ indicates that the CEs retain some degree of anomalousness according to the chosen detector.
An AuROC of $0.5$ indicates that CEs appear completely normal, and for below $0.5$ the CEs are even more normal than the normal test samples. 
This may happen when the anomaly detector does not generalize perfectly and hence perceives some normal test samples as somewhat anomalous. 


\paragraph{Realism of counterfactuals}
To assess the realism of generated samples, the standard approach involves computing the Fréchet inception distance (FID) introduced by \citet{heusel2017gans} for GANs.
The FID is the Wasserstein distance between the feature distributions of a generated dataset and a ground-truth dataset.
The larger the distance, the less the generated dataset resembles the ground truth.
The features are extracted using an InceptionNet v3 model \citep{szegedy2015going} trained on ImageNet. 
In this paper, we used the normal test set as ground truth and a collection of CEs for all test anomalies as the generated dataset.
For a more intuitive scoring, we also computed a second FID with the test anomalies as the generated dataset. 
Then, we normalize the FID for CEs by dividing through the FID for test anomalies.
The normalized FID is $100\%$ if the CEs are as realistic as the test anomalies, below $100\%$ if they are more realistic, and $0\%$ if they exactly match the normal test set.
It is important to note that, although anomalies are naturally anomalous, they are still \emph{realistic} in the sense that they come from the same classification dataset and thus follow the general distribution of, e.g., handwritten digits.
A normalized FID of $100\%$ is therefore sufficient for a counterfactual to be expressive.
A normalized FID of close to $0\%$ would actually be spurious, as the generator then seems to entirely reproduce normal samples that do not retain non-anomalous features from the anomaly.


\section{Hyperparameters} \label{appx:sec:hyperparameters}
In this section, we provide an exhaustive enumeration of all the hyperparameters that we used for training our AD and CE module. 
All hyperparameters were adopted from existing research \citep{ruff2018deep, ghandeharioun_dissect_2021, liznerski2022exposing}.
We start by describing the CE module, which is the same for all datasets and AD objectives.
Then we separately describe the AD module and other hyperparameters for MNIST, Colored-MNIST, CIFAR-10, and GTSDB.

\subsection{The CE Module}

\paragraph{Generator}
The generator is a wide ResNet \citep{zagoruyko2016wide} structured as an encoder-decoder network.
The encoder consists of a sequential arrangement of a batch normalization layer, a convolutional layer with $64$ kernels, and three residual blocks.
Each residual block comprises two sets, each containing a conditional batch normalization layer \citep{de2017modulating}, followed by an activation function (ReLU), and a convolutional layer. The convolutional layers in these sets have $256$, $512$, and $1024$ kernels, respectively, for the first, second, and third block.
The initial two residual blocks employ average pooling in each set to reduce the spatial dimension of the feature maps by one-half of the input, while the third residual block is implemented without average pooling to maintain the spatial dimension.
Conversely, the decoder follows a similar sequential arrangement, featuring three residual blocks, followed by a batch normalization layer, a final convolutional layer mapping to the image space, and an activation function (ReLU).
Again, each residual block comprises two sets, each containing a conditional batch normalization layer, followed by RelU activation, and a convolutional layer.
The convolutional layers in these sets have $1024$, $512$, and $256$ kernels, respectively, for the first, second, and third block.
The first residual block in the decoder retains the spatial dimension, while the subsequent two residual blocks employ an interpolation layer in each set to upsample the spatial dimension by a multiplicative factor of 2 using nearest-neighbor interpolation.
We apply spectral normalization to all layers of the decoder, following \citep{miyato2018spectral}.
The last layer of the decoder uses a tanh activation.
The conditional information, i.e., the discretized target anomaly score $\alpha$ and the target concept $k$ are transformed into a single categorical condition and processed through the categorical conditional batch normalization layers.

\paragraph{Discriminator}
The discriminator contains four residual blocks arranged sequentially, followed by a final linear layer mapping to a scalar. 
The first block is implemented with two convolutional layers with $64$ kernels, where the first layer is followed by a ReLU activation and the second layer is followed by an average pooling with a kernel size of $2$.
The next two residual blocks consist of two convolutional layers, where each one is preceded by a ReLU activation and followed by an average pooling layer in the end to halve the spatial dimension.
The fourth residual block also contains two convolutional layers preceded by a ReLU, but does not use any downsampling. 
The number of kernels in the convolutional layers from the second to fourth block is $128$, $256$, and $512$, respectively.
We apply spectral normalization to all layers.

\paragraph{Concept Classifier} The concept classifier is composed of two sequentially arranged residual blocks, succeeded by a linear layer with two outputs for the classification of two concepts.
In the first residual block, three convolutional layers are employed with $64$ kernels each. 
The initial convolutional layer is succeeded by a ReLU activation, and the last two convolutional layers are followed by average pooling layers, which reduce the spatial dimension by a factor of two.
The second residual block consists of two convolutional layers with $128$ kernels, each followed by a ReLU activation, followed by an average pooling with a kernel size of two.
We take the sum over the remaining spatial dimension to prepare the output for the final linear layer.
Again, we apply spectral normalization to all layers.

\paragraph{Training}
We train the GAN-based generator to generate CEs with two disentangled concepts and a discretized target anomaly score $\alpha \in {0, 0.5, 1}$.
The CE module is trained for $350$ ($2000$ for GTSDB) epochs with a batch size of $64$ normal and, if used, $64$ OE samples.
The initial learning rate is set to $2e^{-4}$, with reductions by a multiplicative factor of $0.1$ occurring after $300$ and $325$ epochs.
For GTSDB, we instead use an initial learning rate of $1e^{-4}$ and reduce it after $1750$ and $1900$ epochs.
We employ the Adam optimizer, with the generator and discriminator optimized every 1 and 5 batches, respectively.
The CE objective involves a combination of different losses which are weighted using $\lambda$ hyperparameters.
Specifically, we set $\lambda_{gan} = 1$, $\lambda_{rec} = 100$, $\lambda_{\phi} = 1$, and $\lambda_r = 10$.
For GTSDB, we instead set $\lambda_{gan} = 5$, $\lambda_{rec} = 20$, $\lambda_{\phi} = 1$, and $\lambda_r = 10$.

For INN, we use a different set of hyperparameters. We employ the diffusion model-based CE method as described in the main paper. We set $\lambda_{gan} = 10$, $\lambda_{rec} = 1$, $\lambda_{\phi} = 1$, and $\lambda_r = 0.5$.
Also, we set $\alpha=0$ for training, as we train the generator with only OE samples to reduce the training time, while the discriminator is trained with normal and generated samples. Due to the immense VRAM requirements of the diffusion model, we train with a batch size of $1$ and use the running statistics of all BatchNorm layers during training. The initial learning rate is set to $1e^{-4}$. It is reduced by a factor of $0.5$ at $100, 120, 130, 140$, and $145$ epochs. The model is trained for $150$ epochs in total.

\subsection{AD on MNIST}
For MNIST and all the following datasets, we trained anomaly detectors with a binary cross entropy (BCE) and hypersphere classification (HSC) loss, both with Outlier Exposure (OE) \citep{hendrycks2019deep}, as well as DSVDD \citep{ruff2018deep} without OE. 

We use a LeNet-style neural network comprising layers arranged sequentially without residual connections. The network contains four convolutional layers and two fully-connected layers. Each convolutional layer is followed by batch normalization, a leaky ReLU activation, and max-pooling. The first fully connected layer is followed by batch normalization and a leaky ReLU activation, while the last layer is only a linear transformation. The number of kernels in the convolutional layers is, from first to last, $4$, $8$, $16$, and $32$. The kernel size is increased from the default of $3$ to $5$ for all of these. The two fully connected layers have $64$ and $32$ units, respectively. For DSVDD we remove bias from the network, following \citep{ruff2018deep}, and for BCE we add another linear layer with sigmoid activation.

We used Adam for optimization and balanced every batch to contain 128 normal and 128 OE samples during training. We trained the AD model for $80$ epochs starting with a learning rate of $1e^{-4}$, which we reduced to $1e^{-5}$ after $60$ epochs.

\subsection{AD on Colored-MNIST}
Based on the MNIST dataset, we create Colored-MNIST where for each sample in MNIST six copies in different colors (red, yellow, green, cyan, blue, pink) are created. We use a colored version of EMNIST as OE. The network for Colored-MNIST is a slight variation of the AD network used on MNIST. We remove the last convolutional layer and change the number of kernels for the convolutional layers to $16$, $32$, and $64$, respectively. 

We use Adam for optimization, balance every batch to contain 128 normal and 128 OE samples during training, and train the AD model for $120$ epochs, starting with a learning rate of $5e^{-5}$, reduced to $5e^{-6}$ after $100$ epochs.

\subsection{AD on CIFAR-10 and GTSDB}
For CIFAR-10, previous work used 80 Mio.~Tiny Images as OE \citep{hendrycks2019using}. However, since 80 Mio.~Tiny Images has officially been withdrawn due to offensive data, we instead use the disjunct CIFAR-100 dataset as OE.
We found that this does not cause a significant drop of performance. 
Again, we use a slight variation of the AD network used on MNIST. 
We remove the last convolutional layer and change the number of kernels for the convolutional layers to $32$, $64$, and $128$, respectively.
The fully connected layers have $512$ and $256$ units instead. 

We use Adam for optimization and balance every batch to contain 128 normal and 128 OE samples during training. 
We train the AD model for $200$ epochs starting with a learning rate of $1e^{-3}$, which we reduce by a factor of $0.1$ after $100$ and $150$ epochs. 
We use the same setup for GTSDB.


\subsection{AD on ImageNet-Neighbors}
For ImageNet-Neighbors (INN), we use the disjoint ImageNet-21k as OE and the same WideResNet architecture as in \citep{hendrycks2019using, liznerski2022exposing}.
We use Adam for optimization and balance every batch to contain 64 normal and 64 OE samples during training. 
We train the AD model for $150$ epochs starting with a learning rate of $1e^{-3}$, which we reduce by a factor of $0.1$ after $100$ and $125$ epochs.

\section{Computing Resources} \label{appx:sec:compute_resources}
Most experiments with MNIST, Colored-MNIST, CIFAR-10, and GTSDB were carried out on a NVIDIA DGX-1 server equipped with eight V100 GPUs with 32 GB memory each.
We here define an experiment as the training and evaluation of an anomaly detector and corresponding counterfactual generator for a single random seed and normal class setting.
We used only one GPU per experiment and ran the experiments in parallel.

Execution times varied by dataset: Experiments on Colored-MNIST took approximately 1.5 days each; Each MNIST and CIFAR-10 experiment took around 8 hours; and GTSDB experiments required roughly 3 hours.
Experiments involving the INN-based method were primarily executed on a NVIDIA DGX A100 server with eight A100 GPUs with 40 GB memory each. In this case, a single INN experiment typically required around ten days to complete.

Inference time for the GAN component is negligible---for example, generating counterfactuals for the entire CIFAR-10 test set takes approximately 40 seconds. For the INN dataset, the diffusion-based method generates a set of counterfactuals for a single test image in about ten seconds.

All the durations are approximate and may vary depending on the specific configuration.
We did not attempt to run the code without GPU acceleration, as all experiments were designed with GPU support in mind.
Each experiment required less than 32 GB of main memory and used up to 8 CPU cores. 
For the full research project, many experiments were repeated multiple times during the method development and prototyping phases. 
Some datasets, such as ImageNet-1k, require substantial storage space (e.g., 300 GB for ImageNet-1k). 
Additionally, experiment logs can become quite large, often reaching several gigabytes per experiment, especially when the training involves more epochs. 
This is due to the logging of downscaled intermediate results at regular intervals throughout training.

\section{Full Quantitative Results per Normal Class} \label{appx:sec:full_quantitative_results}
In the main paper, we presented several objective evaluation techniques to validate the proposed CEs' performance on MNIST, Colored-MNIST (C-MNIST), CIFAR-10, GTSDB, and ImageNet-Neighbors (INN) across different definitions of normality.
Following previous work on semantic image-AD \citep{ruff2018deep, golan2018deep, hendrycks2019deep, hendrycks2019using, ruff2020, tack2020csi, ruff2021, liznerski2021, liznerski2022exposing}, we turned classification datasets into AD benchmarks by defining a subset of the classes to be normal and using the remainder as ground-truth anomalies for testing. 
If only one class is normal, this approach is termed \emph{one vs.~rest} AD.
Apart from investigating one vs.~rest, we also explored a variation with multiple classes being normal. 
For our experiments, we considered all classes of MNIST and CIFAR-10 as single normal classes and, to keep the computational load at a reasonable level, a subset of 20 normal class combinations.
The class combinations were chosen from $\left\lbrace \left(i, (i+1) \mod 10\right)| i \in \lbrace0,\dots,9\rbrace \right\rbrace \cup \left\lbrace \left(i, (i+2) \mod 10\right)| i \in \lbrace0,\dots,9\rbrace \right\rbrace$.
For Colored-MNIST, we considered all combinations of color and the digit one as normal.
For GTSDB, we considered the following combination of street signs as normal: all four combinations of speed limit signs, the ``give way'' and stop sign, and the ``danger'' and ``construction'' warning sign. Additionally, we considered four larger sets of normal classes: all ``restriction ends'' signs, all speed limit signs, all blue signs, and all warning signs.
In total, we consider ten different scenarios of normal definitions for GTSDB.

We introduced ImageNet-Neighbors (INN), which is a subset of ImageNet-1K. As before, we define an AD setup by considering one of the classes normal. However, instead of using the entire remainder as ground-truth test anomalies, we choose only the ten most similar classes, based on the Wu-Palmer similarity metric \citep{wu1994verb}, as test anomalies.
This AD setup becomes harder as compared to the usual one vs.~rest AD setup \citep{hendrycks2019deep}, as the anomalies are more similar to the normal class and thus harder to detect, especially in an unsupervised manner. 
In this paper, we consider five different AD setups for INN. (1) An airliner is normal with airship, wreck, warplane, balloon, monocycle, fireboat, schooner, space shuttle, pirate ship, and gondola as test anomalies. (2) An ambulance is normal with limousine, taxi, waggon, racing car, minivan, jeep, sports car, golf cart, Model T, and convertible as test anomalies. (3) A black widow (spider) is normal with centipede, trilobite, wolf spider, garden spider, barn spider, harvestman, scorpion, black and gold garden spider, tarantula, and tick as test anomalies. (4) A lion is normal with cougar, cheetah, jaguar, tiger cat, leopard, snow leopard, lynx, tiger, tabby cat, and Siamese cat as test anomalies. (5) A zebra is normal with sorrel, llama, warthog, boar, hamster, armadillo, hog, beaver, Arabian camel, and hippo as test anomalies.

For each scenario on each dataset, a new AD model and counterfactual generator was trained for four random seeds. 
We reported quantitative results averaged over all normal definitions in the main paper.
Here, we report results averaged over four random seeds separately for each normal definition.
We consider the following metrics:
\begin{itemize}[itemsep=0.5em,topsep=0.5em,leftmargin=1.5em]
    \item The AD AuROC (Appendix \ref{sec:ad_performance}) is the AuROC of normal vs.~anomalous test samples, thereby measuring the AD performance of the AD model. $50\%$ is random, $100\%$ indicates optimal separation.
    \item The CF AuROC (Section \ref{sec:cf_are_normal}) is the AuROC of normal test samples vs.~counterfactuals. The counterfactuals appear entirely normal for an AuROC $\leq 50\%$. 
    \item The Sub.~AuROC is the AuROC of normal vs.~anomalous test samples when the AD is trained with counterfactuals in place of the normal training set. 
    \item The $FID_N$ (Section \ref{sec:cf_are_realistic}) denotes the normalized FID scores. $0\%$ indicates that the counterfactuals follow the same feature distribution as normal samples, $100\%$ as anomalies, which are also realistic, and above $100\%$ indicates less realistic counterfactuals. 
    \item The Concept Acc (Appendix \ref{sec:cf_disentanglement}) is the accuracy of the concept classifier. A $100\%$ accuracy indicates optimal disentanglement of the concepts. 
    \item The ``Score distance'', which is the L1 distance between the average anomaly score of normal and anomalous test samples.
\end{itemize}

Tables \ref{tbl:appx_mnist_bce_single_normal_class}, \ref{tbl:appx_mnist_hsc_single_normal_class}, and \ref{tbl:appx_mnist_dsvdd_single_normal_class} show results for MNIST and single normal classes for BCE, HSC, and DSVDD, respectively. 
In Tables \ref{tbl:appx_cifar10_bce_single_normal_class}, \ref{tbl:appx_cifar10_hsc_single_normal_class}, and \ref{tbl:appx_cifar10_dsvdd_single_normal_class}, we instead report results for CIFAR-10 and single normal classes for BCE, HSC, and DSVDD, respectively. 
Tables \ref{tbl:appx_coloredmnist_bce_normal_class_sets}, \ref{tbl:appx_coloredmnist_hsc_normal_class_sets}, and \ref{tbl:appx_coloredmnist_dsvdd_normal_class_sets} show results for Colored-MNIST (here abbreviated as C-MNIST) for BCE, HSC, and DSVDD, respectively. 
Tables \ref{tbl:appx_gtsdb_bce_normal_class_sets}, \ref{tbl:appx_gtsdb_hsc_normal_class_sets}, and \ref{tbl:appx_gtsdb_dsvdd_normal_class_sets} show results for GTSDB and combined normal classes for BCE, HSC, and DSVDD, respectively. 
Tables \ref{tbl:appx_mnist_bce_normal_class_sets}, \ref{tbl:appx_mnist_hsc_normal_class_sets}, and \ref{tbl:appx_mnist_dsvdd_normal_class_sets} show results for MNIST and combined normal classes for BCE, HSC, and DSVDD, respectively. 
Tables \ref{tbl:appx_cifar10_bce_normal_class_sets}, \ref{tbl:appx_cifar10_hsc_normal_class_sets}, and \ref{tbl:appx_cifar10_dsvdd_normal_class_sets} show results for CIFAR-10 and combined normal classes for BCE, HSC, and DSVDD, respectively. 
Tables \ref{tbl:appx_inn_bce_single_normal_class} and \ref{tbl:appx_inn_hsc_single_normal_class} show results for ImageNet-Neighbors and single normal classes for BCE and HSC, respectively.

\begin{table}[h]
\centering
\vspace{40pt}
\caption{AD and explanation performance averaged over 4 random seeds on MNIST for BCE (OE). Each row shows results for a different normal definition.}
\label{tbl:appx_mnist_bce_single_normal_class}
\begin{adjustbox}{max width=\linewidth}
\begin{tabular}{lcccccc}
\toprule
 & \multicolumn{2}{|c|}{AD} & \multicolumn{4}{|c|}{Explanation} \\
Normal & AuROC & Score distance & CF AuROC & Sub.~AuROC & FID$_{N}$ & Concept Acc \\
\midrule
zero & 0.99 $\pm$ 0.0010 & 0.78 $\pm$ 0.0079 & 0.76 $\pm$ 0.0684 & 0.93 $\pm$ 0.0104 & 0.42 $\pm$ 0.0366 & 0.97 $\pm$ 0.0360 \\
one & 1.00 $\pm$ 0.0005 & 0.87 $\pm$ 0.0155 & 0.66 $\pm$ 0.0977 & 0.97 $\pm$ 0.0107 & 0.47 $\pm$ 0.4474 & 0.99 $\pm$ 0.0082 \\
two & 0.97 $\pm$ 0.0083 & 0.69 $\pm$ 0.0379 & 0.75 $\pm$ 0.0253 & 0.85 $\pm$ 0.0183 & 0.56 $\pm$ 0.0431 & 0.87 $\pm$ 0.0505 \\
three & 0.99 $\pm$ 0.0018 & 0.67 $\pm$ 0.0286 & 0.77 $\pm$ 0.0242 & 0.94 $\pm$ 0.0073 & 0.33 $\pm$ 0.0392 & 0.89 $\pm$ 0.0834 \\
four & 0.97 $\pm$ 0.0090 & 0.75 $\pm$ 0.0359 & 0.70 $\pm$ 0.0787 & 0.88 $\pm$ 0.0457 & 0.48 $\pm$ 0.0954 & 0.91 $\pm$ 0.0563 \\
five & 0.97 $\pm$ 0.0058 & 0.65 $\pm$ 0.0398 & 0.66 $\pm$ 0.0076 & 0.84 $\pm$ 0.0184 & 0.44 $\pm$ 0.0405 & 0.98 $\pm$ 0.0252 \\
six & 1.00 $\pm$ 0.0010 & 0.90 $\pm$ 0.0106 & 0.71 $\pm$ 0.0527 & 0.98 $\pm$ 0.0066 & 0.33 $\pm$ 0.0348 & 0.96 $\pm$ 0.0359 \\
seven & 0.96 $\pm$ 0.0107 & 0.71 $\pm$ 0.0275 & 0.70 $\pm$ 0.0519 & 0.92 $\pm$ 0.0133 & 0.50 $\pm$ 0.0464 & 0.96 $\pm$ 0.0281 \\
eight & 0.95 $\pm$ 0.0102 & 0.54 $\pm$ 0.0337 & 0.72 $\pm$ 0.0817 & 0.87 $\pm$ 0.0054 & 0.31 $\pm$ 0.0271 & 0.94 $\pm$ 0.0794 \\
nine & 0.96 $\pm$ 0.0092 & 0.60 $\pm$ 0.0329 & 0.77 $\pm$ 0.0147 & 0.94 $\pm$ 0.0080 & 0.47 $\pm$ 0.0593 & 0.97 $\pm$ 0.0189 \\
\midrule
mean & 0.98 $\pm$ 0.0154 & 0.72 $\pm$ 0.1067 & 0.72 $\pm$ 0.0400 & 0.91 $\pm$ 0.0456 & 0.43 $\pm$ 0.0808 & 0.94 $\pm$ 0.0385 \\
\bottomrule
\end{tabular}
\end{adjustbox}
\end{table}

\begin{table}[h!]
\centering
\vspace{40pt}
\caption{AD and explanation performance averaged over 4 random seeds on MNIST for HSC (OE). Each row shows results for a different normal definition.}
\label{tbl:appx_mnist_hsc_single_normal_class}
\begin{adjustbox}{max width=\linewidth}
\begin{tabular}{lcccccc}
\toprule
 & \multicolumn{2}{|c|}{AD} & \multicolumn{4}{|c|}{Explanation} \\
Normal & AuROC & Score distance & CF AuROC & Sub.~AuROC & FID$_{N}$ & Concept Acc \\
\midrule
zero & 0.99 $\pm$ 0.0011 & 0.81 $\pm$ 0.0306 & 0.84 $\pm$ 0.0772 & 0.91 $\pm$ 0.0101 & 0.58 $\pm$ 0.1412 & 0.98 $\pm$ 0.0106 \\
one & 1.00 $\pm$ 0.0011 & 0.89 $\pm$ 0.0231 & 0.88 $\pm$ 0.0783 & 0.95 $\pm$ 0.0089 & 0.60 $\pm$ 0.3820 & 0.90 $\pm$ 0.0868 \\
two & 0.98 $\pm$ 0.0013 & 0.72 $\pm$ 0.0338 & 0.77 $\pm$ 0.0332 & 0.77 $\pm$ 0.0438 & 0.80 $\pm$ 0.3295 & 0.92 $\pm$ 0.0575 \\
three & 0.98 $\pm$ 0.0056 & 0.67 $\pm$ 0.0166 & 0.82 $\pm$ 0.0717 & 0.85 $\pm$ 0.0209 & 0.48 $\pm$ 0.2057 & 0.83 $\pm$ 0.1941 \\
four & 0.96 $\pm$ 0.0038 & 0.73 $\pm$ 0.0269 & 0.80 $\pm$ 0.0658 & 0.84 $\pm$ 0.0394 & 0.83 $\pm$ 0.2911 & 0.81 $\pm$ 0.1526 \\
five & 0.96 $\pm$ 0.0054 & 0.62 $\pm$ 0.0334 & 0.83 $\pm$ 0.0603 & 0.70 $\pm$ 0.1316 & 0.77 $\pm$ 0.1088 & 0.92 $\pm$ 0.1010 \\
six & 1.00 $\pm$ 0.0010 & 0.88 $\pm$ 0.0211 & 0.77 $\pm$ 0.0607 & 0.98 $\pm$ 0.0076 & 0.84 $\pm$ 0.3493 & 0.95 $\pm$ 0.0547 \\
seven & 0.97 $\pm$ 0.0052 & 0.71 $\pm$ 0.0066 & 0.70 $\pm$ 0.0319 & 0.92 $\pm$ 0.0112 & 0.52 $\pm$ 0.0301 & 0.91 $\pm$ 0.0675 \\
eight & 0.95 $\pm$ 0.0069 & 0.52 $\pm$ 0.0334 & 0.89 $\pm$ 0.0278 & 0.73 $\pm$ 0.0590 & 0.88 $\pm$ 0.3052 & 0.94 $\pm$ 0.0739 \\
nine & 0.97 $\pm$ 0.0043 & 0.59 $\pm$ 0.0192 & 0.80 $\pm$ 0.0227 & 0.92 $\pm$ 0.0031 & 0.53 $\pm$ 0.0739 & 0.91 $\pm$ 0.0512 \\
\midrule
mean & 0.98 $\pm$ 0.0157 & 0.72 $\pm$ 0.1156 & 0.81 $\pm$ 0.0526 & 0.86 $\pm$ 0.0919 & 0.68 $\pm$ 0.1464 & 0.91 $\pm$ 0.0478 \\
\bottomrule
\end{tabular}
\end{adjustbox}
\end{table}

\begin{table}[ht]
\centering
\caption{AD and explanation performance averaged over 4 random seeds on MNIST for DSVDD. Each row shows results for a different normal definition.}
\label{tbl:appx_mnist_dsvdd_single_normal_class}
\begin{adjustbox}{max width=\linewidth}
\begin{tabular}{lcccccc}
\toprule
 & \multicolumn{2}{|c|}{AD} & \multicolumn{4}{|c|}{Explanation} \\
Normal & AuROC & Score distance & CF AuROC & Sub.~AuROC & FID$_{N}$ & Concept Acc \\
\midrule
zero & 0.82 $\pm$ 0.0685 & 0.01 $\pm$ 0.0038 & 0.76 $\pm$ 0.0870 & 0.41 $\pm$ 0.0680 & 1.16 $\pm$ 0.5100 & 0.96 $\pm$ 0.0467 \\
one & 1.00 $\pm$ 0.0020 & 0.05 $\pm$ 0.0086 & 0.99 $\pm$ 0.0054 & 0.76 $\pm$ 0.1219 & 1.02 $\pm$ 0.0600 & 0.84 $\pm$ 0.1254 \\
two & 0.72 $\pm$ 0.1254 & 0.01 $\pm$ 0.0057 & 0.69 $\pm$ 0.1664 & 0.34 $\pm$ 0.0203 & 0.89 $\pm$ 0.0117 & 0.49 $\pm$ 0.1150 \\
three & 0.72 $\pm$ 0.0274 & 0.00 $\pm$ 0.0036 & 0.70 $\pm$ 0.0545 & 0.42 $\pm$ 0.0527 & 0.90 $\pm$ 0.0234 & 0.59 $\pm$ 0.1276 \\
four & 0.72 $\pm$ 0.0517 & 0.01 $\pm$ 0.0040 & 0.65 $\pm$ 0.0669 & 0.46 $\pm$ 0.0180 & 0.88 $\pm$ 0.1156 & 0.80 $\pm$ 0.1840 \\
five & 0.73 $\pm$ 0.0316 & 0.01 $\pm$ 0.0050 & 0.71 $\pm$ 0.0562 & 0.44 $\pm$ 0.0632 & 0.97 $\pm$ 0.0869 & 0.87 $\pm$ 0.1221 \\
six & 0.83 $\pm$ 0.0964 & 0.01 $\pm$ 0.0126 & 0.80 $\pm$ 0.1238 & 0.44 $\pm$ 0.0466 & 1.08 $\pm$ 0.0339 & 0.84 $\pm$ 0.1877 \\
seven & 0.84 $\pm$ 0.0450 & 0.01 $\pm$ 0.0135 & 0.80 $\pm$ 0.0533 & 0.46 $\pm$ 0.0858 & 1.04 $\pm$ 0.0408 & 0.88 $\pm$ 0.0291 \\
eight & 0.70 $\pm$ 0.0359 & 0.00 $\pm$ 0.0007 & 0.69 $\pm$ 0.0440 & 0.46 $\pm$ 0.0792 & 0.99 $\pm$ 0.0775 & 0.82 $\pm$ 0.0962 \\
nine & 0.81 $\pm$ 0.0331 & 0.01 $\pm$ 0.0056 & 0.74 $\pm$ 0.0568 & 0.44 $\pm$ 0.0599 & 1.09 $\pm$ 0.0822 & 0.65 $\pm$ 0.3127 \\
\midrule
mean & 0.79 $\pm$ 0.0865 & 0.01 $\pm$ 0.0119 & 0.75 $\pm$ 0.0916 & 0.46 $\pm$ 0.1050 & 1.00 $\pm$ 0.0876 & 0.78 $\pm$ 0.1410 \\
\bottomrule
\end{tabular}
\end{adjustbox}
\end{table}

\begin{table}[h]
\centering
\caption{AD and explanation performance averaged over 2 random seeds on ImageNet-Neighbors for BCE (OE). Each row shows results for a different normal definition.}
\label{tbl:appx_inn_bce_single_normal_class}
\begin{adjustbox}{max width=\linewidth}
\begin{tabular}{lcccccc}
\toprule
 & \multicolumn{1}{|c|}{AD} & \multicolumn{4}{|c|}{Explanation} \\
Normal & AuROC & CF AuROC & Sub.~AuROC & FID$_{N}$ & Concept Acc \\
\midrule
airliner & $96.63 \pm 0.22$ & $76.32 \pm 0.82$ & $65.01 \pm 4.57$ & $95.75 \pm 9.65$ & $99.70 \pm 0.20$ \\ 
ambulance & $98.23 \pm 0.03$ & $83.91 \pm 2.48$ & $63.52 \pm 4.41$ & $105.45 \pm 4.33$ & $99.85 \pm 0.15$ \\ 
black widow & $90.31 \pm 0.41$ & $68.64 \pm 4.25$ & $56.22 \pm 5.19$ & $100.86 \pm 20.66$ & $86.20 \pm 11.40$ \\ 
lion & $84.00 \pm 0.07$ & $34.38 \pm 1.10$ & $61.97 \pm 0.11$ & $94.49 \pm 7.87$ & $100.00 \pm 0.00$ \\ 
zebra & $98.97 \pm 0.02$ & $82.16 \pm 0.65$ & $49.16 \pm 8.66$ & $28.29 \pm 0.43$ & $99.00 \pm 0.70$ \\ 

\midrule
mean & $93.63 \pm 5.70$ & $69.08 \pm 18.15$ & $59.18 \pm 5.83$ & $84.97 \pm 28.61$ & $96.95 \pm 5.39$ \\ 
\bottomrule

\end{tabular}
\end{adjustbox}
\end{table}
\begin{table}[h]
\centering
\caption{AD and explanation performance averaged over 2 random seeds on ImageNet-Neighbors for HSC (OE). Each row shows results for a different normal definition.}
\label{tbl:appx_inn_hsc_single_normal_class}
\begin{adjustbox}{max width=\linewidth}
\begin{tabular}{lcccccc}
\toprule
 & \multicolumn{1}{|c|}{AD} & \multicolumn{4}{|c|}{Explanation} \\
Normal & AuROC & CF AuROC & Sub.~AuROC & FID$_{N}$ & Concept Acc \\
\midrule
airliner & $96.70 \pm 0.04$ & $83.04 \pm 0.32$ & $37.43 \pm 0.32$ & $80.26 \pm 2.12$ & $97.30 \pm 2.10$ \\ 
ambulance & $97.82 \pm 0.01$ & $83.42 \pm 0.67$ & $51.84 \pm 17.77$ & $104.30 \pm 2.86$ & $99.95 \pm 0.05$ \\ 
black widow & $88.20 \pm 0.20$ & $59.68 \pm 0.52$ & $55.09 \pm 1.12$ & $120.69 \pm 10.51$ & $99.60 \pm 0.40$ \\ 
lion & $81.35 \pm 0.74$ & $49.83 \pm 7.35$ & $49.20 \pm 5.02$ & $70.58 \pm 11.86$ & $97.85 \pm 1.85$ \\ 
zebra & $98.78 \pm 0.02$ & $63.84 \pm 3.86$ & $71.63 \pm 1.02$ & $51.17 \pm 6.16$ & $99.70 \pm 0.31$ \\ 

\midrule
mean & $92.57 \pm 6.76$ & $67.96 \pm 13.27$ & $53.04 \pm 11.04$ & $85.40 \pm 24.58$ & $98.88 \pm 1.09$ \\ 
\bottomrule

\end{tabular}
\end{adjustbox}
\end{table}

\begin{table}[h]
\centering
\caption{AD and explanation performance averaged over 4 random seeds on CIFAR-10 for BCE OE. Each row shows results for a different normal definition.}
\label{tbl:appx_cifar10_bce_single_normal_class}
\begin{adjustbox}{max width=\linewidth}
\begin{tabular}{lcccccc}
\toprule
 & \multicolumn{2}{|c|}{AD} & \multicolumn{4}{|c|}{Explanation} \\
Normal & AuROC & Score distance & CF AuROC & Sub.~AuROC & FID$_{N}$ & Concept Acc \\
\midrule
airplane & 0.96 $\pm$ 0.0009 & 0.78 $\pm$ 0.0083 & 0.47 $\pm$ 0.0372 & 0.65 $\pm$ 0.0322 & 1.48 $\pm$ 0.1439 & 0.93 $\pm$ 0.0659 \\
automobile & 0.99 $\pm$ 0.0005 & 0.87 $\pm$ 0.0026 & 0.62 $\pm$ 0.0540 & 0.62 $\pm$ 0.0347 & 1.08 $\pm$ 0.0582 & 0.92 $\pm$ 0.0757 \\
bird & 0.93 $\pm$ 0.0030 & 0.65 $\pm$ 0.0020 & 0.42 $\pm$ 0.0378 & 0.53 $\pm$ 0.0138 & 1.42 $\pm$ 0.0777 & 0.99 $\pm$ 0.0069 \\
cat & 0.91 $\pm$ 0.0035 & 0.55 $\pm$ 0.0127 & 0.30 $\pm$ 0.0054 & 0.53 $\pm$ 0.0159 & 1.37 $\pm$ 0.0773 & 0.91 $\pm$ 0.1449 \\
deer & 0.96 $\pm$ 0.0020 & 0.74 $\pm$ 0.0043 & 0.40 $\pm$ 0.0209 & 0.53 $\pm$ 0.0103 & 1.09 $\pm$ 0.1095 & 0.99 $\pm$ 0.0151 \\
dog & 0.94 $\pm$ 0.0013 & 0.64 $\pm$ 0.0051 & 0.36 $\pm$ 0.0061 & 0.57 $\pm$ 0.0134 & 1.23 $\pm$ 0.0777 & 0.93 $\pm$ 0.1008 \\
frog & 0.98 $\pm$ 0.0011 & 0.79 $\pm$ 0.0067 & 0.50 $\pm$ 0.0247 & 0.54 $\pm$ 0.0127 & 0.80 $\pm$ 0.0652 & 0.88 $\pm$ 0.1341 \\
horse & 0.98 $\pm$ 0.0006 & 0.82 $\pm$ 0.0060 & 0.59 $\pm$ 0.0303 & 0.64 $\pm$ 0.0213 & 1.21 $\pm$ 0.1013 & 0.99 $\pm$ 0.0107 \\
ship & 0.98 $\pm$ 0.0002 & 0.85 $\pm$ 0.0032 & 0.55 $\pm$ 0.0098 & 0.72 $\pm$ 0.0300 & 0.93 $\pm$ 0.0810 & 0.89 $\pm$ 0.0760 \\
truck & 0.97 $\pm$ 0.0018 & 0.78 $\pm$ 0.0080 & 0.54 $\pm$ 0.0602 & 0.56 $\pm$ 0.0242 & 1.03 $\pm$ 0.1231 & 0.88 $\pm$ 0.2031 \\
\midrule
mean & 0.96 $\pm$ 0.0252 & 0.75 $\pm$ 0.0964 & 0.47 $\pm$ 0.1000 & 0.59 $\pm$ 0.0610 & 1.16 $\pm$ 0.2078 & 0.93 $\pm$ 0.0429 \\
\bottomrule
\end{tabular}
\end{adjustbox}
\end{table}

\begin{table}[h]
\centering
\caption{AD and explanation performance averaged over 4 random seeds on CIFAR-10 for HSC OE. Each row shows results for a different normal definition.}
\label{tbl:appx_cifar10_hsc_single_normal_class}
\begin{adjustbox}{max width=\linewidth}
\begin{tabular}{lcccccc}
\toprule
 & \multicolumn{2}{|c|}{AD} & \multicolumn{4}{|c|}{Explanation} \\
Normal & AuROC & Score distance & CF AuROC & Sub.~AuROC & FID$_{N}$ & Concept Acc \\
\midrule
airplane & 0.96 $\pm$ 0.0012 & 0.75 $\pm$ 0.0056 & 0.51 $\pm$ 0.0754 & 0.52 $\pm$ 0.0111 & 2.95 $\pm$ 0.1509 & 0.89 $\pm$ 0.0873 \\
automobile & 0.99 $\pm$ 0.0005 & 0.85 $\pm$ 0.0030 & 0.58 $\pm$ 0.0152 & 0.59 $\pm$ 0.0129 & 1.71 $\pm$ 0.1914 & 0.99 $\pm$ 0.0054 \\
bird & 0.93 $\pm$ 0.0015 & 0.62 $\pm$ 0.0018 & 0.46 $\pm$ 0.0293 & 0.52 $\pm$ 0.0149 & 4.81 $\pm$ 0.2365 & 1.00 $\pm$ 0.0007 \\
cat & 0.90 $\pm$ 0.0020 & 0.53 $\pm$ 0.0072 & 0.43 $\pm$ 0.0255 & 0.52 $\pm$ 0.0088 & 3.98 $\pm$ 0.4753 & 1.00 $\pm$ 0.0009 \\
deer & 0.96 $\pm$ 0.0007 & 0.71 $\pm$ 0.0040 & 0.51 $\pm$ 0.0121 & 0.57 $\pm$ 0.0230 & 3.45 $\pm$ 0.3143 & 1.00 $\pm$ 0.0000 \\
dog & 0.95 $\pm$ 0.0012 & 0.65 $\pm$ 0.0047 & 0.46 $\pm$ 0.0317 & 0.53 $\pm$ 0.0257 & 3.09 $\pm$ 0.2897 & 1.00 $\pm$ 0.0023 \\
frog & 0.98 $\pm$ 0.0004 & 0.77 $\pm$ 0.0043 & 0.52 $\pm$ 0.0062 & 0.57 $\pm$ 0.0569 & 2.92 $\pm$ 0.4138 & 1.00 $\pm$ 0.0009 \\
horse & 0.98 $\pm$ 0.0008 & 0.79 $\pm$ 0.0040 & 0.54 $\pm$ 0.0466 & 0.54 $\pm$ 0.0281 & 3.13 $\pm$ 0.0463 & 1.00 $\pm$ 0.0001 \\
ship & 0.98 $\pm$ 0.0003 & 0.83 $\pm$ 0.0027 & 0.48 $\pm$ 0.0257 & 0.56 $\pm$ 0.0316 & 1.86 $\pm$ 0.5187 & 1.00 $\pm$ 0.0032 \\
truck & 0.97 $\pm$ 0.0011 & 0.77 $\pm$ 0.0055 & 0.51 $\pm$ 0.0257 & 0.57 $\pm$ 0.0623 & 2.19 $\pm$ 0.1318 & 1.00 $\pm$ 0.0010 \\
\midrule
mean & 0.96 $\pm$ 0.0254 & 0.73 $\pm$ 0.0939 & 0.50 $\pm$ 0.0438 & 0.55 $\pm$ 0.0259 & 3.01 $\pm$ 0.8998 & 0.99 $\pm$ 0.0325 \\
\bottomrule
\end{tabular}
\end{adjustbox}
\end{table}

\begin{table}[h]
\centering
\caption{AD and explanation performance averaged over 4 random seeds on CIFAR-10 for DSVDD. Each row shows results for a different normal definition.}
\label{tbl:appx_cifar10_dsvdd_single_normal_class}
\begin{adjustbox}{max width=\linewidth}
\begin{tabular}{lcccccc}
\toprule
 & \multicolumn{2}{|c|}{AD} & \multicolumn{4}{|c|}{Explanation} \\
Normal & AuROC & Score distance & CF AuROC & Sub.~AuROC & FID$_{N}$ & Concept Acc \\
\midrule
airplane & 0.48 $\pm$ 0.0952 & -0.00 $\pm$ 0.0022 & 0.54 $\pm$ 0.0733 & 0.45 $\pm$ 0.0265 & 1.28 $\pm$ 0.0382 & 0.98 $\pm$ 0.0114 \\
automobile & 0.51 $\pm$ 0.0339 & 0.00 $\pm$ 0.0003 & 0.52 $\pm$ 0.0606 & 0.49 $\pm$ 0.0198 & 1.15 $\pm$ 0.0266 & 0.99 $\pm$ 0.0076 \\
bird & 0.54 $\pm$ 0.0375 & 0.00 $\pm$ 0.0005 & 0.52 $\pm$ 0.0601 & 0.51 $\pm$ 0.0133 & 1.23 $\pm$ 0.0548 & 0.91 $\pm$ 0.1548 \\
cat & 0.52 $\pm$ 0.0216 & 0.00 $\pm$ 0.0008 & 0.51 $\pm$ 0.0513 & 0.50 $\pm$ 0.0260 & 1.38 $\pm$ 0.1380 & 0.98 $\pm$ 0.0221 \\
deer & 0.65 $\pm$ 0.0312 & 0.01 $\pm$ 0.0030 & 0.62 $\pm$ 0.0996 & 0.53 $\pm$ 0.0611 & 1.12 $\pm$ 0.0467 & 1.00 $\pm$ 0.0028 \\
dog & 0.53 $\pm$ 0.0259 & 0.00 $\pm$ 0.0030 & 0.51 $\pm$ 0.0296 & 0.50 $\pm$ 0.0195 & 1.21 $\pm$ 0.0830 & 0.96 $\pm$ 0.0523 \\
frog & 0.60 $\pm$ 0.0692 & 0.01 $\pm$ 0.0027 & 0.54 $\pm$ 0.0371 & 0.57 $\pm$ 0.0747 & 0.99 $\pm$ 0.0550 & 0.99 $\pm$ 0.0074 \\
horse & 0.56 $\pm$ 0.0253 & 0.00 $\pm$ 0.0025 & 0.53 $\pm$ 0.0281 & 0.51 $\pm$ 0.0143 & 1.21 $\pm$ 0.0094 & 1.00 $\pm$ 0.0037 \\
ship & 0.57 $\pm$ 0.0543 & 0.00 $\pm$ 0.0010 & 0.58 $\pm$ 0.0350 & 0.53 $\pm$ 0.0561 & 0.97 $\pm$ 0.0611 & 0.93 $\pm$ 0.0758 \\
truck & 0.58 $\pm$ 0.0673 & 0.00 $\pm$ 0.0008 & 0.58 $\pm$ 0.0470 & 0.48 $\pm$ 0.0224 & 1.10 $\pm$ 0.0258 & 0.97 $\pm$ 0.0417 \\
\midrule
mean & 0.55 $\pm$ 0.0473 & 0.00 $\pm$ 0.0022 & 0.55 $\pm$ 0.0336 & 0.51 $\pm$ 0.0315 & 1.16 $\pm$ 0.1195 & 0.97 $\pm$ 0.0287 \\
\bottomrule
\end{tabular}
\end{adjustbox}
\end{table}

\begin{table}[h]
\centering
\caption{AD and explanation performance averaged over 4 random seeds on C-MNIST for BCE (OE). Each row shows results for a different normal definition.}
\label{tbl:appx_coloredmnist_bce_normal_class_sets}
\begin{adjustbox}{max width=\linewidth}
\begin{tabular}{lcccccc}
\toprule
 & \multicolumn{2}{|c|}{AD} & \multicolumn{4}{|c|}{Explanation} \\
Normal & AuROC & Score distance & CF AuROC & Sub.~AuROC & FID$_{N}$ & Concept Acc \\
\midrule
gray+one & 0.96 $\pm$ 0.0037 & 0.17 $\pm$ 0.0127 & 0.55 $\pm$ 0.1105 & 0.75 $\pm$ 0.0429 & 0.75 $\pm$ 0.3352 & 0.96 $\pm$ 0.0327 \\
yellow+one & 0.97 $\pm$ 0.0027 & 0.24 $\pm$ 0.0129 & 0.56 $\pm$ 0.0252 & 0.74 $\pm$ 0.0082 & 0.60 $\pm$ 0.1572 & 1.00 $\pm$ 0.0001 \\
cyan+one & 0.96 $\pm$ 0.0138 & 0.19 $\pm$ 0.0373 & 0.54 $\pm$ 0.0410 & 0.83 $\pm$ 0.0180 & 0.38 $\pm$ 0.0340 & 1.00 $\pm$ 0.0007 \\
green+one & 0.99 $\pm$ 0.0044 & 0.49 $\pm$ 0.0546 & 0.58 $\pm$ 0.0457 & 0.80 $\pm$ 0.0676 & 0.60 $\pm$ 0.2606 & 1.00 $\pm$ 0.0001 \\
blue+one & 0.98 $\pm$ 0.0034 & 0.48 $\pm$ 0.0110 & 0.55 $\pm$ 0.0075 & 0.81 $\pm$ 0.0640 & 0.52 $\pm$ 0.1925 & 1.00 $\pm$ 0.0002 \\
pink+one & 0.97 $\pm$ 0.0021 & 0.25 $\pm$ 0.0193 & 0.57 $\pm$ 0.0279 & 0.88 $\pm$ 0.0127 & 0.43 $\pm$ 0.0647 & 1.00 $\pm$ 0.0003 \\
red+one & 0.98 $\pm$ 0.0031 & 0.42 $\pm$ 0.0364 & 0.54 $\pm$ 0.1100 & 0.83 $\pm$ 0.0938 & 0.69 $\pm$ 0.4817 & 1.00 $\pm$ 0.0015 \\
\midrule
mean & 0.97 $\pm$ 0.0101 & 0.32 $\pm$ 0.1265 & 0.56 $\pm$ 0.0154 & 0.81 $\pm$ 0.0451 & 0.57 $\pm$ 0.1240 & 0.99 $\pm$ 0.0132 \\
\bottomrule
\end{tabular}
\end{adjustbox}
\end{table}

\begin{table}[h]
\centering
\caption{AD and explanation performance averaged over 4 random seeds on C-MNIST for HSC (OE). Each row shows results for a different normal definition.}
\label{tbl:appx_coloredmnist_hsc_normal_class_sets}
\begin{adjustbox}{max width=\linewidth}
\begin{tabular}{lcccccc}
\toprule
 & \multicolumn{2}{|c|}{AD} & \multicolumn{4}{|c|}{Explanation} \\
Normal & AuROC & Score distance & CF AuROC & Sub.~AuROC & FID$_{N}$ & Concept Acc \\
\midrule
gray+one & 0.92 $\pm$ 0.0075 & 0.27 $\pm$ 0.0410 & 0.51 $\pm$ 0.0486 & 0.76 $\pm$ 0.0457 & 0.86 $\pm$ 0.1567 & 0.99 $\pm$ 0.0136 \\
yellow+one & 0.94 $\pm$ 0.0251 & 0.43 $\pm$ 0.0509 & 0.54 $\pm$ 0.0615 & 0.82 $\pm$ 0.0081 & 0.82 $\pm$ 0.2713 & 1.00 $\pm$ 0.0020 \\
cyan+one & 0.97 $\pm$ 0.0196 & 0.39 $\pm$ 0.0630 & 0.56 $\pm$ 0.0296 & 0.88 $\pm$ 0.0462 & 0.63 $\pm$ 0.2201 & 1.00 $\pm$ 0.0000 \\
green+one & 0.98 $\pm$ 0.0139 & 0.52 $\pm$ 0.0258 & 0.56 $\pm$ 0.0323 & 0.89 $\pm$ 0.0102 & 0.94 $\pm$ 0.2280 & 1.00 $\pm$ 0.0005 \\
blue+one & 0.99 $\pm$ 0.0028 & 0.65 $\pm$ 0.0159 & 0.66 $\pm$ 0.0896 & 0.75 $\pm$ 0.1384 & 1.66 $\pm$ 1.1219 & 0.94 $\pm$ 0.0834 \\
pink+one & 0.94 $\pm$ 0.0139 & 0.38 $\pm$ 0.0323 & 0.52 $\pm$ 0.0751 & 0.83 $\pm$ 0.0339 & 0.83 $\pm$ 0.0292 & 1.00 $\pm$ 0.0015 \\
red+one & 0.98 $\pm$ 0.0031 & 0.60 $\pm$ 0.0127 & 0.57 $\pm$ 0.0244 & 0.78 $\pm$ 0.0674 & 0.93 $\pm$ 0.3331 & 1.00 $\pm$ 0.0055 \\
\midrule
mean & 0.96 $\pm$ 0.0231 & 0.46 $\pm$ 0.1226 & 0.56 $\pm$ 0.0472 & 0.82 $\pm$ 0.0482 & 0.95 $\pm$ 0.3047 & 0.99 $\pm$ 0.0198 \\
\bottomrule
\end{tabular}
\end{adjustbox}
\end{table}

\begin{table}[h]
\centering
\caption{AD and explanation performance averaged over 4 random seeds on C-MNIST for DSVDD. Each row shows results for a different normal definition.}
\label{tbl:appx_coloredmnist_dsvdd_normal_class_sets}
\begin{adjustbox}{max width=\linewidth}
\begin{tabular}{lcccccc}
\toprule
 & \multicolumn{2}{|c|}{AD} & \multicolumn{4}{|c|}{Explanation} \\
Normal & AuROC & Score distance & CF AuROC & Sub.~AuROC & FID$_{N}$ & Concept Acc \\
\midrule
gray+one & 0.73 $\pm$ 0.0350 & 0.00 $\pm$ 0.0001 & 0.56 $\pm$ 0.0449 & 0.71 $\pm$ 0.0755 & 0.85 $\pm$ 0.2079 & 0.91 $\pm$ 0.0834 \\
yellow+one & 0.86 $\pm$ 0.0262 & 0.00 $\pm$ 0.0010 & 0.60 $\pm$ 0.0595 & 0.65 $\pm$ 0.0639 & 0.82 $\pm$ 0.2240 & 1.00 $\pm$ 0.0044 \\
cyan+one & 0.83 $\pm$ 0.0866 & 0.00 $\pm$ 0.0005 & 0.61 $\pm$ 0.0781 & 0.63 $\pm$ 0.0589 & 0.79 $\pm$ 0.0524 & 0.99 $\pm$ 0.0057 \\
green+one & 0.64 $\pm$ 0.1336 & 0.00 $\pm$ 0.0003 & 0.57 $\pm$ 0.0250 & 0.60 $\pm$ 0.0755 & 0.69 $\pm$ 0.0350 & 1.00 $\pm$ 0.0019 \\
blue+one & 0.78 $\pm$ 0.1502 & 0.00 $\pm$ 0.0001 & 0.68 $\pm$ 0.2173 & 0.42 $\pm$ 0.1223 & 1.01 $\pm$ 0.1866 & 1.00 $\pm$ 0.0016 \\
pink+one & 0.75 $\pm$ 0.1343 & 0.00 $\pm$ 0.0001 & 0.67 $\pm$ 0.1040 & 0.61 $\pm$ 0.0999 & 0.85 $\pm$ 0.0998 & 0.97 $\pm$ 0.0214 \\
red+one & 0.79 $\pm$ 0.0424 & 0.00 $\pm$ 0.0004 & 0.62 $\pm$ 0.0917 & 0.57 $\pm$ 0.1607 & 0.81 $\pm$ 0.1763 & 0.99 $\pm$ 0.0149 \\
\midrule
mean & 0.77 $\pm$ 0.0650 & 0.00 $\pm$ 0.0003 & 0.61 $\pm$ 0.0430 & 0.60 $\pm$ 0.0841 & 0.83 $\pm$ 0.0875 & 0.98 $\pm$ 0.0297 \\
\bottomrule
\end{tabular}
\end{adjustbox}
\end{table}

\begin{table}[h]
\centering
\caption{AD and explanation performance averaged over 4 random seeds on GTSDB for BCE OE. Each row shows results for a different normal definition.}
\label{tbl:appx_gtsdb_bce_normal_class_sets}
\begin{adjustbox}{max width=\linewidth}
\setlength\tabcolsep{5pt}
\begin{tabular}{lcccccc}
\toprule
 & \multicolumn{2}{|c|}{AD} & \multicolumn{4}{|c|}{Explanation} \\
Normal & AuROC & Score distance & CF AuROC & Sub.~AuROC & FID$_{N}$ & Concept Acc \\
\midrule
speed limit 30 + 50 & 0.92 $\pm$ 0.0037 & 0.65 $\pm$ 0.0103 & 0.51 $\pm$ 0.0563 & 0.88 $\pm$ 0.0158 & 0.77 $\pm$ 0.3590 & 1.00 $\pm$ 0.0018 \\
speed limit 50 + 70 & 0.88 $\pm$ 0.0151 & 0.59 $\pm$ 0.0188 & 0.49 $\pm$ 0.0576 & 0.86 $\pm$ 0.0066 & 0.69 $\pm$ 0.3249 & 0.99 $\pm$ 0.0080 \\
speed limit 70 + 100 & 0.88 $\pm$ 0.0053 & 0.57 $\pm$ 0.0048 & 0.55 $\pm$ 0.0708 & 0.89 $\pm$ 0.0136 & 0.42 $\pm$ 0.1348 & 0.99 $\pm$ 0.0130 \\
speed limit 100 + 120 & 0.89 $\pm$ 0.0200 & 0.55 $\pm$ 0.0409 & 0.49 $\pm$ 0.1331 & 0.87 $\pm$ 0.0297 & 0.51 $\pm$ 0.0854 & 0.99 $\pm$ 0.0115 \\
give way + stop & 0.99 $\pm$ 0.0021 & 0.89 $\pm$ 0.0131 & 0.66 $\pm$ 0.0758 & 0.81 $\pm$ 0.1369 & 2.29 $\pm$ 0.4255 & 0.99 $\pm$ 0.0184 \\
danger + construction warning & 0.93 $\pm$ 0.0078 & 0.73 $\pm$ 0.0072 & 0.43 $\pm$ 0.0799 & 0.91 $\pm$ 0.0155 & 3.60 $\pm$ 0.5202 & 1.00 $\pm$ 0.0040 \\
all restriction ends signs & 1.00 $\pm$ 0.0029 & 0.90 $\pm$ 0.0167 & 0.56 $\pm$ 0.1341 & 1.00 $\pm$ 0.0033 & 0.24 $\pm$ 0.1129 & 0.97 $\pm$ 0.0183 \\
all speed limit signs & 0.99 $\pm$ 0.0016 & 0.79 $\pm$ 0.0226 & 0.54 $\pm$ 0.0172 & 0.96 $\pm$ 0.0085 & 0.41 $\pm$ 0.0870 & 0.99 $\pm$ 0.0134 \\
all blue signs & 1.00 $\pm$ 0.0023 & 0.93 $\pm$ 0.0131 & 0.40 $\pm$ 0.0381 & 0.90 $\pm$ 0.0258 & 0.64 $\pm$ 0.1553 & 0.98 $\pm$ 0.0109 \\
all warning signs & 0.96 $\pm$ 0.0089 & 0.89 $\pm$ 0.0132 & 0.38 $\pm$ 0.0343 & 0.95 $\pm$ 0.0035 & 1.51 $\pm$ 0.5426 & 0.99 $\pm$ 0.0076 \\
\midrule
mean & 0.94 $\pm$ 0.0474 & 0.75 $\pm$ 0.1437 & 0.50 $\pm$ 0.0803 & 0.90 $\pm$ 0.0526 & 1.11 $\pm$ 1.0182 & 0.99 $\pm$ 0.0085 \\
\bottomrule
\end{tabular}
\end{adjustbox}
\end{table}

\begin{table}[h]
\centering
\caption{AD and explanation performance averaged over 4 random seeds on GTSDB for HSC OE. Each row shows results for a different normal definition.}
\label{tbl:appx_gtsdb_hsc_normal_class_sets}
\begin{adjustbox}{max width=.99\textwidth}
\setlength\tabcolsep{5pt}
\begin{tabular}{lcccccc}
\toprule
 & \multicolumn{2}{|c|}{AD} & \multicolumn{4}{|c|}{Explanation} \\
Normal & AuROC & Score distance & CF AuROC & Sub.~AuROC & FID$_{N}$ & Concept Acc \\
\midrule
speed limit 30 + 50 & 0.88 $\pm$ 0.0014 & 0.63 $\pm$ 0.0126 & 0.31 $\pm$ 0.1032 & 0.88 $\pm$ 0.0113 & 0.79 $\pm$ 0.2196 & 0.96 $\pm$ 0.0420 \\
speed limit 50 + 70 & 0.89 $\pm$ 0.0111 & 0.57 $\pm$ 0.0170 & 0.49 $\pm$ 0.1537 & 0.85 $\pm$ 0.0135 & 1.45 $\pm$ 0.6565 & 1.00 $\pm$ 0.0000 \\
speed limit 70 + 100 & 0.86 $\pm$ 0.0164 & 0.56 $\pm$ 0.0146 & 0.60 $\pm$ 0.1389 & 0.85 $\pm$ 0.0379 & 0.69 $\pm$ 0.4033 & 0.91 $\pm$ 0.0807 \\
speed limit 100 + 120 & 0.85 $\pm$ 0.0112 & 0.50 $\pm$ 0.0132 & 0.66 $\pm$ 0.0952 & 0.86 $\pm$ 0.0172 & 0.59 $\pm$ 0.2818 & 0.95 $\pm$ 0.0613 \\
give way + stop & 0.98 $\pm$ 0.0056 & 0.81 $\pm$ 0.0415 & 0.70 $\pm$ 0.1508 & 0.83 $\pm$ 0.0929 & 1.00 $\pm$ 0.1991 & 0.70 $\pm$ 0.0922 \\
danger + construction warning & 0.91 $\pm$ 0.0099 & 0.68 $\pm$ 0.0121 & 0.32 $\pm$ 0.0889 & 0.90 $\pm$ 0.0137 & 2.82 $\pm$ 0.2851 & 0.97 $\pm$ 0.0210 \\
all restriction ends signs & 1.00 $\pm$ 0.0000 & 0.93 $\pm$ 0.0127 & 0.60 $\pm$ 0.0791 & 1.00 $\pm$ 0.0039 & 0.21 $\pm$ 0.0519 & 0.94 $\pm$ 0.0221 \\
all speed limit signs & 0.96 $\pm$ 0.0174 & 0.79 $\pm$ 0.0075 & 0.51 $\pm$ 0.0419 & 0.95 $\pm$ 0.0175 & 0.29 $\pm$ 0.0730 & 0.97 $\pm$ 0.0469 \\
all blue signs & 1.00 $\pm$ 0.0011 & 0.94 $\pm$ 0.0165 & 0.34 $\pm$ 0.0640 & 0.91 $\pm$ 0.0224 & 0.38 $\pm$ 0.0667 & 1.00 $\pm$ 0.0023 \\
all warning signs & 0.97 $\pm$ 0.0042 & 0.86 $\pm$ 0.0182 & 0.33 $\pm$ 0.0692 & 0.96 $\pm$ 0.0061 & 1.31 $\pm$ 0.2118 & 1.00 $\pm$ 0.0036 \\
\midrule
mean & 0.93 $\pm$ 0.0563 & 0.73 $\pm$ 0.1517 & 0.49 $\pm$ 0.1439 & 0.90 $\pm$ 0.0508 & 0.95 $\pm$ 0.7345 & 0.94 $\pm$ 0.0840 \\
\bottomrule
\end{tabular}
\end{adjustbox}
\end{table}

\begin{table}[h]
\centering
\caption{AD and explanation performance averaged over 4 random seeds on GTSDB for DSVDD. Each row shows results for a different normal definition.}
\label{tbl:appx_gtsdb_dsvdd_normal_class_sets}
\begin{adjustbox}{max width=.99\textwidth}
\setlength\tabcolsep{5pt}
\begin{tabular}{lcccccc}
\toprule
 & \multicolumn{2}{|c|}{AD} & \multicolumn{4}{|c|}{Explanation} \\
Normal & AuROC & Score distance & CF AuROC & Sub.~AuROC & FID$_{N}$ & Concept Acc \\
\midrule
speed limit 30 + 50 & 0.53 $\pm$ 0.0718 & 0.06 $\pm$ 0.0214 & 0.56 $\pm$ 0.0583 & 0.57 $\pm$ 0.0240 & 1.07 $\pm$ 0.4804 & 0.95 $\pm$ 0.0439 \\
speed limit 50 + 70 & 0.55 $\pm$ 0.0487 & 0.07 $\pm$ 0.0640 & 0.60 $\pm$ 0.1042 & 0.57 $\pm$ 0.0485 & 3.59 $\pm$ 3.8551 & 0.87 $\pm$ 0.1167 \\
speed limit 70 + 100 & 0.56 $\pm$ 0.0433 & 0.02 $\pm$ 0.0108 & 0.53 $\pm$ 0.1288 & 0.63 $\pm$ 0.0291 & 0.34 $\pm$ 0.0187 & 0.92 $\pm$ 0.0376 \\
speed limit 100 + 120 & 0.61 $\pm$ 0.0497 & 0.04 $\pm$ 0.0171 & 0.53 $\pm$ 0.0625 & 0.64 $\pm$ 0.0488 & 0.28 $\pm$ 0.0315 & 0.95 $\pm$ 0.0302 \\
give way + stop & 0.49 $\pm$ 0.0673 & 0.00 $\pm$ 0.0150 & 0.46 $\pm$ 0.0981 & 0.49 $\pm$ 0.0725 & 1.88 $\pm$ 0.5662 & 0.98 $\pm$ 0.0138 \\
danger + construction warning & 0.61 $\pm$ 0.0429 & 0.02 $\pm$ 0.0049 & 0.59 $\pm$ 0.0402 & 0.47 $\pm$ 0.0348 & 3.04 $\pm$ 0.3589 & 0.90 $\pm$ 0.1063 \\
all restriction ends signs & 0.70 $\pm$ 0.0860 & 0.06 $\pm$ 0.0450 & 0.53 $\pm$ 0.1242 & 0.69 $\pm$ 0.0862 & 0.26 $\pm$ 0.1251 & 0.94 $\pm$ 0.0273 \\
all speed limit signs & 0.69 $\pm$ 0.0473 & 0.05 $\pm$ 0.0095 & 0.57 $\pm$ 0.0533 & 0.64 $\pm$ 0.0145 & 0.51 $\pm$ 0.1984 & 0.98 $\pm$ 0.0182 \\
all blue signs & 0.51 $\pm$ 0.1008 & 0.02 $\pm$ 0.0161 & 0.49 $\pm$ 0.0985 & 0.64 $\pm$ 0.0117 & 0.20 $\pm$ 0.0484 & 0.86 $\pm$ 0.0565 \\
all warning signs & 0.56 $\pm$ 0.0242 & 0.01 $\pm$ 0.0087 & 0.46 $\pm$ 0.0616 & 0.51 $\pm$ 0.0484 & 1.93 $\pm$ 0.5590 & 1.00 $\pm$ 0.0034 \\
\midrule
mean & 0.58 $\pm$ 0.0668 & 0.04 $\pm$ 0.0233 & 0.53 $\pm$ 0.0478 & 0.58 $\pm$ 0.0699 & 1.31 $\pm$ 1.1807 & 0.93 $\pm$ 0.0453 \\
\bottomrule
\end{tabular}
\end{adjustbox}
\end{table}

\begin{table}[h]
\centering
\caption{AD and explanation performance averaged over 4 random seeds on MNIST for BCE (OE). Each row shows results for a different normal definition.}
\label{tbl:appx_mnist_bce_normal_class_sets}
\begin{adjustbox}{max width=0.999\linewidth}
\begin{tabular}{lcccccc}
\toprule
 & \multicolumn{2}{|c|}{AD} & \multicolumn{4}{|c|}{Explanation} \\
Normal & AuROC & Score distance & CF AuROC & Sub.~AuROC & FID$_{N}$ & Concept Acc \\
\midrule
zero+one & 0.97 $\pm$ 0.0062 & 0.51 $\pm$ 0.0596 & 0.79 $\pm$ 0.0864 & 0.45 $\pm$ 0.0944 & 1.00 $\pm$ 0.0674 & 0.98 $\pm$ 0.0154 \\
zero+two & 0.95 $\pm$ 0.0129 & 0.44 $\pm$ 0.0694 & 0.82 $\pm$ 0.0696 & 0.59 $\pm$ 0.0292 & 0.77 $\pm$ 0.0372 & 0.95 $\pm$ 0.0520 \\
one+two & 0.94 $\pm$ 0.0188 & 0.46 $\pm$ 0.0688 & 0.74 $\pm$ 0.0251 & 0.40 $\pm$ 0.0411 & 1.25 $\pm$ 0.0237 & 0.99 $\pm$ 0.0101 \\
one+three & 0.95 $\pm$ 0.0097 & 0.45 $\pm$ 0.0222 & 0.70 $\pm$ 0.0433 & 0.56 $\pm$ 0.0241 & 1.18 $\pm$ 0.0250 & 0.97 $\pm$ 0.0192 \\
two+three & 0.97 $\pm$ 0.0095 & 0.56 $\pm$ 0.0667 & 0.76 $\pm$ 0.0720 & 0.79 $\pm$ 0.0188 & 0.51 $\pm$ 0.0498 & 0.99 $\pm$ 0.0131 \\
two+four & 0.89 $\pm$ 0.0196 & 0.35 $\pm$ 0.0551 & 0.75 $\pm$ 0.0415 & 0.42 $\pm$ 0.0421 & 0.83 $\pm$ 0.0824 & 1.00 $\pm$ 0.0017 \\
three+four & 0.91 $\pm$ 0.0070 & 0.33 $\pm$ 0.0250 & 0.81 $\pm$ 0.0290 & 0.58 $\pm$ 0.0415 & 0.85 $\pm$ 0.0359 & 0.93 $\pm$ 0.0687 \\
three+five & 0.95 $\pm$ 0.0058 & 0.48 $\pm$ 0.0487 & 0.74 $\pm$ 0.0213 & 0.67 $\pm$ 0.0515 & 0.43 $\pm$ 0.0501 & 0.95 $\pm$ 0.0360 \\
four+five & 0.90 $\pm$ 0.0259 & 0.30 $\pm$ 0.0148 & 0.83 $\pm$ 0.0474 & 0.40 $\pm$ 0.0485 & 0.92 $\pm$ 0.0715 & 0.82 $\pm$ 0.1926 \\
four+six & 0.95 $\pm$ 0.0052 & 0.57 $\pm$ 0.0364 & 0.77 $\pm$ 0.0333 & 0.63 $\pm$ 0.0650 & 0.67 $\pm$ 0.1253 & 0.98 $\pm$ 0.0277 \\
five+six & 0.97 $\pm$ 0.0063 & 0.60 $\pm$ 0.0319 & 0.82 $\pm$ 0.0672 & 0.63 $\pm$ 0.0514 & 0.55 $\pm$ 0.0666 & 0.91 $\pm$ 0.0797 \\
five+seven & 0.88 $\pm$ 0.0228 & 0.40 $\pm$ 0.0453 & 0.76 $\pm$ 0.0546 & 0.59 $\pm$ 0.0416 & 1.02 $\pm$ 0.0697 & 0.94 $\pm$ 0.0361 \\
six+seven & 0.94 $\pm$ 0.0143 & 0.44 $\pm$ 0.0618 & 0.85 $\pm$ 0.0437 & 0.66 $\pm$ 0.0622 & 0.92 $\pm$ 0.1281 & 0.82 $\pm$ 0.1436 \\
six+eight & 0.95 $\pm$ 0.0145 & 0.45 $\pm$ 0.0398 & 0.81 $\pm$ 0.0474 & 0.63 $\pm$ 0.0608 & 0.38 $\pm$ 0.0205 & 0.96 $\pm$ 0.0539 \\
seven+eight & 0.87 $\pm$ 0.0208 & 0.33 $\pm$ 0.0300 & 0.73 $\pm$ 0.0562 & 0.70 $\pm$ 0.0264 & 0.90 $\pm$ 0.0669 & 0.91 $\pm$ 0.0795 \\
seven+nine & 0.95 $\pm$ 0.0209 & 0.58 $\pm$ 0.0374 & 0.77 $\pm$ 0.0628 & 0.88 $\pm$ 0.0201 & 0.94 $\pm$ 0.1804 & 0.86 $\pm$ 0.1010 \\
eight+nine & 0.93 $\pm$ 0.0189 & 0.42 $\pm$ 0.0492 & 0.80 $\pm$ 0.0483 & 0.83 $\pm$ 0.0144 & 0.48 $\pm$ 0.0423 & 0.93 $\pm$ 0.1050 \\
eight+zero & 0.93 $\pm$ 0.0100 & 0.39 $\pm$ 0.0219 & 0.77 $\pm$ 0.0908 & 0.69 $\pm$ 0.0240 & 0.46 $\pm$ 0.0200 & 0.98 $\pm$ 0.0177 \\
nine+zero & 0.95 $\pm$ 0.0047 & 0.49 $\pm$ 0.0184 & 0.85 $\pm$ 0.0398 & 0.77 $\pm$ 0.0424 & 0.54 $\pm$ 0.0610 & 0.92 $\pm$ 0.0678 \\
nine+one & 0.93 $\pm$ 0.0157 & 0.39 $\pm$ 0.0365 & 0.73 $\pm$ 0.0944 & 0.57 $\pm$ 0.0461 & 1.09 $\pm$ 0.0559 & 0.97 $\pm$ 0.0191 \\
\midrule
mean & 0.93 $\pm$ 0.0283 & 0.45 $\pm$ 0.0868 & 0.78 $\pm$ 0.0412 & 0.62 $\pm$ 0.1325 & 0.78 $\pm$ 0.2596 & 0.94 $\pm$ 0.0512 \\
\bottomrule
\end{tabular}
\end{adjustbox}
\end{table}

\begin{table}[h]
\centering
\caption{AD and explanation performance averaged over 4 random seeds on MNIST for HSC (OE). Each row shows results for a different normal definition.}
\label{tbl:appx_mnist_hsc_normal_class_sets}
\begin{adjustbox}{max width=0.95\linewidth}
\begin{tabular}{lcccccc}
\toprule
 & \multicolumn{2}{|c|}{AD} & \multicolumn{4}{|c|}{Explanation} \\
Normal & AuROC & Score distance & CF AuROC & Sub.~AuROC & FID$_{N}$ & Concept Acc \\
\midrule
zero+one & 0.98 $\pm$ 0.0056 & 0.53 $\pm$ 0.0871 & 0.88 $\pm$ 0.0450 & 0.46 $\pm$ 0.0714 & 1.13 $\pm$ 0.0433 & 0.92 $\pm$ 0.1256 \\
zero+two & 0.95 $\pm$ 0.0120 & 0.52 $\pm$ 0.0508 & 0.87 $\pm$ 0.0267 & 0.39 $\pm$ 0.0644 & 0.96 $\pm$ 0.0884 & 0.94 $\pm$ 0.0697 \\
one+two & 0.96 $\pm$ 0.0061 & 0.48 $\pm$ 0.0493 & 0.83 $\pm$ 0.0163 & 0.46 $\pm$ 0.1134 & 1.23 $\pm$ 0.0469 & 0.95 $\pm$ 0.0382 \\
one+three & 0.95 $\pm$ 0.0081 & 0.51 $\pm$ 0.0142 & 0.84 $\pm$ 0.0519 & 0.55 $\pm$ 0.0545 & 1.24 $\pm$ 0.0717 & 0.85 $\pm$ 0.2038 \\
two+three & 0.95 $\pm$ 0.0116 & 0.58 $\pm$ 0.0371 & 0.74 $\pm$ 0.0500 & 0.59 $\pm$ 0.0706 & 0.73 $\pm$ 0.1404 & 0.87 $\pm$ 0.1477 \\
two+four & 0.86 $\pm$ 0.0132 & 0.33 $\pm$ 0.0276 & 0.77 $\pm$ 0.0338 & 0.39 $\pm$ 0.0131 & 0.92 $\pm$ 0.0227 & 0.98 $\pm$ 0.0168 \\
three+four & 0.87 $\pm$ 0.0190 & 0.34 $\pm$ 0.0472 & 0.73 $\pm$ 0.0515 & 0.55 $\pm$ 0.0355 & 0.87 $\pm$ 0.0564 & 0.87 $\pm$ 0.1123 \\
three+five & 0.93 $\pm$ 0.0294 & 0.50 $\pm$ 0.0450 & 0.80 $\pm$ 0.0902 & 0.54 $\pm$ 0.0523 & 0.54 $\pm$ 0.0908 & 0.85 $\pm$ 0.1274 \\
four+five & 0.87 $\pm$ 0.0160 & 0.33 $\pm$ 0.0228 & 0.86 $\pm$ 0.0449 & 0.42 $\pm$ 0.0571 & 1.35 $\pm$ 0.4027 & 0.58 $\pm$ 0.0420 \\
four+six & 0.95 $\pm$ 0.0128 & 0.55 $\pm$ 0.0598 & 0.82 $\pm$ 0.0360 & 0.50 $\pm$ 0.1191 & 0.82 $\pm$ 0.0307 & 0.97 $\pm$ 0.0223 \\
five+six & 0.95 $\pm$ 0.0058 & 0.57 $\pm$ 0.0471 & 0.83 $\pm$ 0.0505 & 0.54 $\pm$ 0.0711 & 1.03 $\pm$ 0.3435 & 0.83 $\pm$ 0.0677 \\
five+seven & 0.89 $\pm$ 0.0022 & 0.40 $\pm$ 0.0223 & 0.83 $\pm$ 0.0281 & 0.58 $\pm$ 0.0241 & 1.33 $\pm$ 0.2102 & 0.80 $\pm$ 0.1326 \\
six+seven & 0.92 $\pm$ 0.0166 & 0.43 $\pm$ 0.0602 & 0.81 $\pm$ 0.0535 & 0.54 $\pm$ 0.0695 & 1.02 $\pm$ 0.3005 & 0.87 $\pm$ 0.0852 \\
six+eight & 0.94 $\pm$ 0.0031 & 0.44 $\pm$ 0.0373 & 0.81 $\pm$ 0.0184 & 0.51 $\pm$ 0.0417 & 0.51 $\pm$ 0.1461 & 0.88 $\pm$ 0.0918 \\
seven+eight & 0.90 $\pm$ 0.0090 & 0.42 $\pm$ 0.0328 & 0.78 $\pm$ 0.0331 & 0.66 $\pm$ 0.0287 & 1.14 $\pm$ 0.0710 & 0.91 $\pm$ 0.0864 \\
seven+nine & 0.96 $\pm$ 0.0034 & 0.63 $\pm$ 0.0163 & 0.85 $\pm$ 0.0637 & 0.81 $\pm$ 0.0430 & 1.17 $\pm$ 0.2448 & 0.65 $\pm$ 0.2011 \\
eight+nine & 0.93 $\pm$ 0.0049 & 0.44 $\pm$ 0.0268 & 0.83 $\pm$ 0.0483 & 0.69 $\pm$ 0.0317 & 0.67 $\pm$ 0.1301 & 0.87 $\pm$ 0.1908 \\
eight+zero & 0.93 $\pm$ 0.0075 & 0.44 $\pm$ 0.0215 & 0.83 $\pm$ 0.0602 & 0.55 $\pm$ 0.0547 & 0.80 $\pm$ 0.4024 & 0.85 $\pm$ 0.1161 \\
nine+zero & 0.94 $\pm$ 0.0052 & 0.48 $\pm$ 0.0601 & 0.85 $\pm$ 0.0379 & 0.61 $\pm$ 0.0466 & 0.65 $\pm$ 0.0405 & 0.77 $\pm$ 0.1480 \\
nine+one & 0.95 $\pm$ 0.0119 & 0.44 $\pm$ 0.0212 & 0.83 $\pm$ 0.0464 & 0.60 $\pm$ 0.0340 & 1.13 $\pm$ 0.0206 & 0.92 $\pm$ 0.0678 \\
\midrule
mean & 0.93 $\pm$ 0.0332 & 0.47 $\pm$ 0.0809 & 0.82 $\pm$ 0.0378 & 0.55 $\pm$ 0.0987 & 0.96 $\pm$ 0.2502 & 0.86 $\pm$ 0.0963 \\
\bottomrule
\end{tabular}
\end{adjustbox}
\end{table}

\begin{table}[h]
\centering
\caption{AD and explanation performance averaged over 4 random seeds on MNIST for DSVDD. Each row shows results for a different normal definition.}
\label{tbl:appx_mnist_dsvdd_normal_class_sets}
\begin{adjustbox}{max width=0.95\linewidth}
\begin{tabular}{lcccccc}
\toprule
 & \multicolumn{2}{|c|}{AD} & \multicolumn{4}{|c|}{Explanation} \\
Normal & AuROC & Score distance & CF AuROC & Sub.~AuROC & FID$_{N}$ & Concept Acc \\
\midrule
zero+one & 0.93 $\pm$ 0.0323 & 0.00 $\pm$ 0.0018 & 0.90 $\pm$ 0.0393 & 0.57 $\pm$ 0.0150 & 1.05 $\pm$ 0.1323 & 0.97 $\pm$ 0.0254 \\
zero+two & 0.71 $\pm$ 0.1290 & 0.00 $\pm$ 0.0015 & 0.70 $\pm$ 0.1319 & 0.36 $\pm$ 0.0439 & 0.99 $\pm$ 0.0301 & 0.54 $\pm$ 0.2298 \\
one+two & 0.73 $\pm$ 0.0542 & 0.00 $\pm$ 0.0003 & 0.73 $\pm$ 0.0648 & 0.38 $\pm$ 0.0584 & 1.16 $\pm$ 0.0277 & 0.92 $\pm$ 0.0666 \\
one+three & 0.77 $\pm$ 0.0422 & 0.00 $\pm$ 0.0002 & 0.78 $\pm$ 0.0470 & 0.43 $\pm$ 0.1285 & 1.13 $\pm$ 0.0103 & 0.87 $\pm$ 0.1073 \\
two+three & 0.69 $\pm$ 0.0508 & 0.00 $\pm$ 0.0015 & 0.67 $\pm$ 0.0495 & 0.38 $\pm$ 0.1011 & 0.86 $\pm$ 0.0373 & 0.81 $\pm$ 0.2033 \\
two+four & 0.85 $\pm$ 0.0253 & 0.00 $\pm$ 0.0009 & 0.80 $\pm$ 0.0380 & 0.39 $\pm$ 0.0484 & 0.75 $\pm$ 0.1440 & 0.85 $\pm$ 0.2204 \\
three+four & 0.77 $\pm$ 0.0716 & 0.00 $\pm$ 0.0015 & 0.73 $\pm$ 0.0736 & 0.46 $\pm$ 0.0377 & 0.92 $\pm$ 0.0610 & 0.72 $\pm$ 0.2467 \\
three+five & 0.66 $\pm$ 0.0275 & 0.00 $\pm$ 0.0003 & 0.66 $\pm$ 0.0346 & 0.43 $\pm$ 0.0459 & 0.86 $\pm$ 0.0218 & 0.76 $\pm$ 0.1619 \\
four+five & 0.71 $\pm$ 0.1077 & 0.00 $\pm$ 0.0026 & 0.70 $\pm$ 0.0907 & 0.41 $\pm$ 0.0192 & 0.98 $\pm$ 0.0285 & 0.71 $\pm$ 0.0798 \\
four+six & 0.81 $\pm$ 0.0719 & 0.01 $\pm$ 0.0037 & 0.80 $\pm$ 0.0915 & 0.37 $\pm$ 0.0288 & 1.03 $\pm$ 0.0127 & 0.86 $\pm$ 0.1675 \\
five+six & 0.72 $\pm$ 0.0814 & 0.00 $\pm$ 0.0028 & 0.70 $\pm$ 0.0749 & 0.41 $\pm$ 0.0568 & 0.93 $\pm$ 0.0151 & 0.73 $\pm$ 0.1704 \\
five+seven & 0.72 $\pm$ 0.0564 & 0.00 $\pm$ 0.0009 & 0.69 $\pm$ 0.0281 & 0.44 $\pm$ 0.0658 & 0.96 $\pm$ 0.0983 & 0.85 $\pm$ 0.1442 \\
six+seven & 0.84 $\pm$ 0.0609 & 0.00 $\pm$ 0.0015 & 0.79 $\pm$ 0.0271 & 0.41 $\pm$ 0.0469 & 1.13 $\pm$ 0.0494 & 0.94 $\pm$ 0.0260 \\
six+eight & 0.78 $\pm$ 0.0681 & 0.00 $\pm$ 0.0013 & 0.75 $\pm$ 0.0787 & 0.44 $\pm$ 0.0241 & 0.93 $\pm$ 0.1650 & 0.79 $\pm$ 0.1834 \\
seven+eight & 0.70 $\pm$ 0.0095 & 0.00 $\pm$ 0.0002 & 0.70 $\pm$ 0.0046 & 0.39 $\pm$ 0.0721 & 1.12 $\pm$ 0.0105 & 0.95 $\pm$ 0.0364 \\
seven+nine & 0.74 $\pm$ 0.0744 & 0.00 $\pm$ 0.0020 & 0.75 $\pm$ 0.0758 & 0.38 $\pm$ 0.0345 & 1.10 $\pm$ 0.0419 & 0.72 $\pm$ 0.1768 \\
eight+nine & 0.69 $\pm$ 0.0688 & 0.00 $\pm$ 0.0006 & 0.68 $\pm$ 0.0712 & 0.42 $\pm$ 0.0329 & 0.95 $\pm$ 0.1594 & 0.97 $\pm$ 0.0480 \\
eight+zero & 0.66 $\pm$ 0.0560 & 0.00 $\pm$ 0.0009 & 0.65 $\pm$ 0.0630 & 0.37 $\pm$ 0.0299 & 1.05 $\pm$ 0.0253 & 0.82 $\pm$ 0.1814 \\
nine+zero & 0.72 $\pm$ 0.0834 & 0.00 $\pm$ 0.0016 & 0.67 $\pm$ 0.1228 & 0.46 $\pm$ 0.0408 & 0.99 $\pm$ 0.1008 & 0.65 $\pm$ 0.3174 \\
nine+one & 0.84 $\pm$ 0.0555 & 0.00 $\pm$ 0.0010 & 0.85 $\pm$ 0.0489 & 0.42 $\pm$ 0.1575 & 1.13 $\pm$ 0.0173 & 0.91 $\pm$ 0.0509 \\
\midrule
mean & 0.75 $\pm$ 0.0712 & 0.00 $\pm$ 0.0013 & 0.73 $\pm$ 0.0649 & 0.42 $\pm$ 0.0450 & 1.00 $\pm$ 0.1074 & 0.82 $\pm$ 0.1132 \\
\bottomrule
\end{tabular}
\end{adjustbox}
\end{table}

\begin{table}
\centering
\caption{AD and explanation performance averaged over 4 random seeds on CIFAR-10 for BCE OE. Each row shows results for a different normal definition.}
\label{tbl:appx_cifar10_bce_normal_class_sets}
\begin{adjustbox}{max width=.99\textwidth}
\setlength\tabcolsep{5pt}
\begin{tabular}{lcccccc}
\toprule
 & \multicolumn{2}{|c|}{AD} & \multicolumn{4}{|c|}{Explanation} \\
Normal & AuROC & Score distance & CF AuROC & Sub.~AuROC & FID$_{N}$ & Concept Acc \\
\midrule
airplane+automobile & 0.96 $\pm$ 0.0024 & 0.79 $\pm$ 0.0066 & 0.59 $\pm$ 0.0300 & 0.66 $\pm$ 0.0187 & 1.04 $\pm$ 0.0824 & 0.75 $\pm$ 0.1067 \\
airplane+bird & 0.92 $\pm$ 0.0017 & 0.68 $\pm$ 0.0043 & 0.45 $\pm$ 0.0226 & 0.61 $\pm$ 0.0087 & 1.34 $\pm$ 0.2551 & 0.88 $\pm$ 0.1167 \\
automobile+bird & 0.93 $\pm$ 0.0023 & 0.70 $\pm$ 0.0029 & 0.57 $\pm$ 0.0340 & 0.59 $\pm$ 0.0264 & 1.79 $\pm$ 0.0164 & 0.73 $\pm$ 0.2012 \\
automobile+cat & 0.90 $\pm$ 0.0038 & 0.61 $\pm$ 0.0005 & 0.46 $\pm$ 0.0113 & 0.54 $\pm$ 0.0060 & 1.73 $\pm$ 0.0686 & 0.87 $\pm$ 0.0738 \\
bird+cat & 0.87 $\pm$ 0.0022 & 0.53 $\pm$ 0.0019 & 0.35 $\pm$ 0.0207 & 0.54 $\pm$ 0.0140 & 1.19 $\pm$ 0.1377 & 0.81 $\pm$ 0.1128 \\
bird+deer & 0.92 $\pm$ 0.0004 & 0.64 $\pm$ 0.0046 & 0.39 $\pm$ 0.0233 & 0.53 $\pm$ 0.0069 & 0.92 $\pm$ 0.0889 & 0.97 $\pm$ 0.0038 \\
cat+deer & 0.90 $\pm$ 0.0025 & 0.58 $\pm$ 0.0077 & 0.39 $\pm$ 0.0301 & 0.53 $\pm$ 0.0148 & 0.94 $\pm$ 0.0475 & 0.89 $\pm$ 0.1547 \\
cat+dog & 0.91 $\pm$ 0.0023 & 0.59 $\pm$ 0.0108 & 0.30 $\pm$ 0.0103 & 0.58 $\pm$ 0.0099 & 0.91 $\pm$ 0.0472 & 0.81 $\pm$ 0.1551 \\
deer+dog & 0.92 $\pm$ 0.0006 & 0.64 $\pm$ 0.0040 & 0.42 $\pm$ 0.0333 & 0.55 $\pm$ 0.0137 & 0.88 $\pm$ 0.0511 & 0.93 $\pm$ 0.0495 \\
deer+frog & 0.94 $\pm$ 0.0014 & 0.70 $\pm$ 0.0042 & 0.49 $\pm$ 0.0381 & 0.52 $\pm$ 0.0124 & 0.76 $\pm$ 0.0422 & 0.82 $\pm$ 0.1905 \\
dog+frog & 0.93 $\pm$ 0.0010 & 0.67 $\pm$ 0.0053 & 0.46 $\pm$ 0.0181 & 0.56 $\pm$ 0.0121 & 0.93 $\pm$ 0.0769 & 0.94 $\pm$ 0.0597 \\
dog+horse & 0.95 $\pm$ 0.0022 & 0.71 $\pm$ 0.0056 & 0.50 $\pm$ 0.0085 & 0.58 $\pm$ 0.0106 & 1.01 $\pm$ 0.0391 & 0.89 $\pm$ 0.1399 \\
frog+horse & 0.96 $\pm$ 0.0007 & 0.76 $\pm$ 0.0080 & 0.55 $\pm$ 0.0314 & 0.56 $\pm$ 0.0170 & 1.03 $\pm$ 0.0501 & 0.81 $\pm$ 0.1722 \\
frog+ship & 0.95 $\pm$ 0.0010 & 0.76 $\pm$ 0.0046 & 0.53 $\pm$ 0.0225 & 0.62 $\pm$ 0.0188 & 1.06 $\pm$ 0.2823 & 0.88 $\pm$ 0.0802 \\
horse+ship & 0.97 $\pm$ 0.0010 & 0.80 $\pm$ 0.0047 & 0.58 $\pm$ 0.0259 & 0.61 $\pm$ 0.0420 & 0.95 $\pm$ 0.1126 & 0.97 $\pm$ 0.0323 \\
horse+truck & 0.96 $\pm$ 0.0008 & 0.77 $\pm$ 0.0046 & 0.56 $\pm$ 0.0293 & 0.60 $\pm$ 0.0195 & 1.08 $\pm$ 0.0864 & 0.87 $\pm$ 0.1812 \\
ship+truck & 0.96 $\pm$ 0.0011 & 0.77 $\pm$ 0.0059 & 0.54 $\pm$ 0.0200 & 0.62 $\pm$ 0.0171 & 0.78 $\pm$ 0.0594 & 0.93 $\pm$ 0.1109 \\
ship+airplane & 0.97 $\pm$ 0.0008 & 0.80 $\pm$ 0.0044 & 0.52 $\pm$ 0.0392 & 0.71 $\pm$ 0.0113 & 0.77 $\pm$ 0.1048 & 0.97 $\pm$ 0.0441 \\
truck+airplane & 0.95 $\pm$ 0.0008 & 0.75 $\pm$ 0.0027 & 0.55 $\pm$ 0.0137 & 0.61 $\pm$ 0.0370 & 0.93 $\pm$ 0.0557 & 0.73 $\pm$ 0.1478 \\
truck+automobile & 0.98 $\pm$ 0.0010 & 0.85 $\pm$ 0.0041 & 0.62 $\pm$ 0.0429 & 0.60 $\pm$ 0.0240 & 0.75 $\pm$ 0.0793 & 0.80 $\pm$ 0.1978 \\
\midrule
mean & 0.94 $\pm$ 0.0266 & 0.71 $\pm$ 0.0839 & 0.49 $\pm$ 0.0847 & 0.59 $\pm$ 0.0460 & 1.04 $\pm$ 0.2794 & 0.86 $\pm$ 0.0745 \\
\bottomrule
\end{tabular}
\end{adjustbox}
\end{table}

\begin{table}
\centering
\caption{AD and explanation performance averaged over 4 random seeds on CIFAR-10 for HSC OE. Each row shows results for a different normal definition.}
\label{tbl:appx_cifar10_hsc_normal_class_sets}
\begin{adjustbox}{max width=.99\textwidth}
\setlength\tabcolsep{5pt}
\begin{tabular}{lcccccc}
\toprule
 & \multicolumn{2}{|c|}{AD} & \multicolumn{4}{|c|}{Explanation} \\
Normal & AuROC & Score distance & CF AuROC & Sub.~AuROC & FID$_{N}$ & Concept Acc \\
\midrule
airplane+automobile & 0.96 $\pm$ 0.0005 & 0.75 $\pm$ 0.0017 & 0.51 $\pm$ 0.0900 & 0.54 $\pm$ 0.0163 & 2.14 $\pm$ 0.0882 & 0.99 $\pm$ 0.0164 \\
airplane+bird & 0.93 $\pm$ 0.0012 & 0.67 $\pm$ 0.0024 & 0.44 $\pm$ 0.0439 & 0.52 $\pm$ 0.0059 & 2.21 $\pm$ 0.1630 & 1.00 $\pm$ 0.0002 \\
automobile+bird & 0.92 $\pm$ 0.0029 & 0.66 $\pm$ 0.0065 & 0.45 $\pm$ 0.0424 & 0.51 $\pm$ 0.0065 & 4.12 $\pm$ 1.1471 & 1.00 $\pm$ 0.0001 \\
automobile+cat & 0.91 $\pm$ 0.0011 & 0.62 $\pm$ 0.0054 & 0.53 $\pm$ 0.0285 & 0.50 $\pm$ 0.0023 & 3.10 $\pm$ 0.3450 & 1.00 $\pm$ 0.0011 \\
bird+cat & 0.87 $\pm$ 0.0019 & 0.47 $\pm$ 0.0046 & 0.32 $\pm$ 0.0328 & 0.53 $\pm$ 0.0401 & 3.34 $\pm$ 1.0615 & 1.00 $\pm$ 0.0002 \\
bird+deer & 0.92 $\pm$ 0.0026 & 0.63 $\pm$ 0.0097 & 0.38 $\pm$ 0.0144 & 0.54 $\pm$ 0.0248 & 3.49 $\pm$ 0.1061 & 1.00 $\pm$ 0.0012 \\
cat+deer & 0.90 $\pm$ 0.0017 & 0.54 $\pm$ 0.0053 & 0.35 $\pm$ 0.0228 & 0.52 $\pm$ 0.0166 & 2.58 $\pm$ 0.1145 & 1.00 $\pm$ 0.0000 \\
cat+dog & 0.93 $\pm$ 0.0018 & 0.59 $\pm$ 0.0085 & 0.39 $\pm$ 0.0252 & 0.52 $\pm$ 0.0042 & 1.97 $\pm$ 0.0935 & 1.00 $\pm$ 0.0003 \\
deer+dog & 0.92 $\pm$ 0.0017 & 0.60 $\pm$ 0.0095 & 0.38 $\pm$ 0.0401 & 0.52 $\pm$ 0.0107 & 2.44 $\pm$ 0.5742 & 0.96 $\pm$ 0.0734 \\
deer+frog & 0.95 $\pm$ 0.0011 & 0.68 $\pm$ 0.0010 & 0.42 $\pm$ 0.0065 & 0.56 $\pm$ 0.0535 & 2.27 $\pm$ 0.0879 & 1.00 $\pm$ 0.0002 \\
dog+frog & 0.93 $\pm$ 0.0014 & 0.63 $\pm$ 0.0045 & 0.43 $\pm$ 0.0110 & 0.51 $\pm$ 0.0036 & 2.53 $\pm$ 0.1879 & 1.00 $\pm$ 0.0001 \\
dog+horse & 0.96 $\pm$ 0.0003 & 0.70 $\pm$ 0.0064 & 0.44 $\pm$ 0.0062 & 0.52 $\pm$ 0.0190 & 3.22 $\pm$ 0.1861 & 1.00 $\pm$ 0.0001 \\
frog+horse & 0.96 $\pm$ 0.0015 & 0.73 $\pm$ 0.0027 & 0.48 $\pm$ 0.0143 & 0.52 $\pm$ 0.0176 & 2.75 $\pm$ 0.3541 & 1.00 $\pm$ 0.0001 \\
frog+ship & 0.96 $\pm$ 0.0009 & 0.75 $\pm$ 0.0084 & 0.48 $\pm$ 0.0313 & 0.56 $\pm$ 0.0346 & 3.29 $\pm$ 0.6680 & 1.00 $\pm$ 0.0001 \\
horse+ship & 0.96 $\pm$ 0.0007 & 0.77 $\pm$ 0.0036 & 0.40 $\pm$ 0.0675 & 0.53 $\pm$ 0.0124 & 1.87 $\pm$ 0.0485 & 1.00 $\pm$ 0.0005 \\
horse+truck & 0.95 $\pm$ 0.0016 & 0.73 $\pm$ 0.0074 & 0.50 $\pm$ 0.0339 & 0.53 $\pm$ 0.0520 & 2.93 $\pm$ 0.8821 & 1.00 $\pm$ 0.0011 \\
ship+truck & 0.96 $\pm$ 0.0005 & 0.76 $\pm$ 0.0051 & 0.41 $\pm$ 0.0426 & 0.57 $\pm$ 0.0625 & 1.73 $\pm$ 0.0526 & 0.99 $\pm$ 0.0075 \\
ship+airplane & 0.97 $\pm$ 0.0013 & 0.80 $\pm$ 0.0037 & 0.53 $\pm$ 0.0811 & 0.55 $\pm$ 0.0359 & 1.65 $\pm$ 0.2366 & 0.98 $\pm$ 0.0247 \\
truck+airplane & 0.95 $\pm$ 0.0020 & 0.72 $\pm$ 0.0041 & 0.46 $\pm$ 0.0542 & 0.53 $\pm$ 0.0176 & 1.85 $\pm$ 0.1448 & 0.97 $\pm$ 0.0579 \\
truck+automobile & 0.99 $\pm$ 0.0004 & 0.85 $\pm$ 0.0067 & 0.60 $\pm$ 0.0790 & 0.53 $\pm$ 0.0340 & 1.49 $\pm$ 0.1063 & 0.90 $\pm$ 0.1301 \\
\midrule
mean & 0.94 $\pm$ 0.0270 & 0.68 $\pm$ 0.0883 & 0.44 $\pm$ 0.0666 & 0.53 $\pm$ 0.0175 & 2.55 $\pm$ 0.6970 & 0.99 $\pm$ 0.0244 \\
\bottomrule
\end{tabular}
\end{adjustbox}
\end{table}

\begin{table}
\centering
\caption{AD and explanation performance averaged over 4 random seeds on CIFAR-10 for DSVDD. Each row shows results for a different normal definition.}
\label{tbl:appx_cifar10_dsvdd_normal_class_sets}
\begin{adjustbox}{max width=.99\textwidth}
\setlength\tabcolsep{5pt}
\begin{tabular}{lcccccc}
\toprule
 & \multicolumn{2}{|c|}{AD} & \multicolumn{4}{|c|}{Explanation} \\
Normal & AuROC & Score distance & CF AuROC & Sub.~AuROC & FID$_{N}$ & Concept Acc \\
\midrule
airplane+automobile & 0.50 $\pm$ 0.0357 & 0.00 $\pm$ 0.0002 & 0.48 $\pm$ 0.0517 & 0.46 $\pm$ 0.0260 & 1.20 $\pm$ 0.0111 & 0.84 $\pm$ 0.1424 \\
airplane+bird & 0.49 $\pm$ 0.0111 & 0.00 $\pm$ 0.0005 & 0.46 $\pm$ 0.0219 & 0.49 $\pm$ 0.0448 & 1.27 $\pm$ 0.0950 & 0.93 $\pm$ 0.0503 \\
automobile+bird & 0.49 $\pm$ 0.0145 & 0.00 $\pm$ 0.0002 & 0.49 $\pm$ 0.0081 & 0.49 $\pm$ 0.0184 & 1.23 $\pm$ 0.0524 & 0.93 $\pm$ 0.0859 \\
automobile+cat & 0.50 $\pm$ 0.0148 & 0.00 $\pm$ 0.0007 & 0.48 $\pm$ 0.0153 & 0.47 $\pm$ 0.0251 & 1.22 $\pm$ 0.0567 & 0.90 $\pm$ 0.0745 \\
bird+cat & 0.53 $\pm$ 0.0162 & 0.00 $\pm$ 0.0003 & 0.51 $\pm$ 0.0344 & 0.50 $\pm$ 0.0033 & 1.08 $\pm$ 0.0223 & 0.98 $\pm$ 0.0223 \\
bird+deer & 0.56 $\pm$ 0.0278 & 0.00 $\pm$ 0.0003 & 0.54 $\pm$ 0.0345 & 0.51 $\pm$ 0.0122 & 0.97 $\pm$ 0.0304 & 0.97 $\pm$ 0.0183 \\
cat+deer & 0.56 $\pm$ 0.0418 & 0.00 $\pm$ 0.0008 & 0.54 $\pm$ 0.0486 & 0.53 $\pm$ 0.0228 & 1.02 $\pm$ 0.0201 & 0.95 $\pm$ 0.0201 \\
cat+dog & 0.52 $\pm$ 0.0105 & 0.00 $\pm$ 0.0011 & 0.49 $\pm$ 0.0332 & 0.49 $\pm$ 0.0148 & 1.06 $\pm$ 0.0168 & 0.91 $\pm$ 0.0690 \\
deer+dog & 0.55 $\pm$ 0.0213 & 0.00 $\pm$ 0.0030 & 0.51 $\pm$ 0.0377 & 0.53 $\pm$ 0.0211 & 1.10 $\pm$ 0.0348 & 0.89 $\pm$ 0.1620 \\
deer+frog & 0.57 $\pm$ 0.1151 & 0.01 $\pm$ 0.0046 & 0.53 $\pm$ 0.1167 & 0.59 $\pm$ 0.0516 & 0.87 $\pm$ 0.0342 & 0.93 $\pm$ 0.0919 \\
dog+frog & 0.60 $\pm$ 0.0431 & 0.00 $\pm$ 0.0034 & 0.60 $\pm$ 0.0514 & 0.53 $\pm$ 0.0323 & 0.95 $\pm$ 0.0188 & 0.87 $\pm$ 0.0848 \\
dog+horse & 0.53 $\pm$ 0.0102 & 0.00 $\pm$ 0.0006 & 0.49 $\pm$ 0.0408 & 0.49 $\pm$ 0.0178 & 1.17 $\pm$ 0.0254 & 0.92 $\pm$ 0.0427 \\
frog+horse & 0.60 $\pm$ 0.0398 & 0.01 $\pm$ 0.0048 & 0.56 $\pm$ 0.0160 & 0.57 $\pm$ 0.0228 & 1.07 $\pm$ 0.0079 & 0.99 $\pm$ 0.0030 \\
frog+ship & 0.52 $\pm$ 0.0144 & 0.00 $\pm$ 0.0004 & 0.50 $\pm$ 0.0326 & 0.53 $\pm$ 0.0188 & 1.08 $\pm$ 0.0331 & 0.97 $\pm$ 0.0261 \\
horse+ship & 0.49 $\pm$ 0.0374 & 0.00 $\pm$ 0.0002 & 0.48 $\pm$ 0.0409 & 0.48 $\pm$ 0.0077 & 1.17 $\pm$ 0.0563 & 0.96 $\pm$ 0.0209 \\
horse+truck & 0.50 $\pm$ 0.0346 & 0.00 $\pm$ 0.0006 & 0.51 $\pm$ 0.0287 & 0.46 $\pm$ 0.0147 & 1.21 $\pm$ 0.0579 & 0.88 $\pm$ 0.1041 \\
ship+truck & 0.47 $\pm$ 0.0265 & 0.00 $\pm$ 0.0003 & 0.49 $\pm$ 0.0195 & 0.46 $\pm$ 0.0201 & 1.05 $\pm$ 0.0330 & 0.96 $\pm$ 0.0365 \\
ship+airplane & 0.50 $\pm$ 0.0246 & 0.00 $\pm$ 0.0002 & 0.48 $\pm$ 0.0400 & 0.42 $\pm$ 0.0326 & 1.10 $\pm$ 0.0722 & 0.87 $\pm$ 0.1070 \\
truck+airplane & 0.48 $\pm$ 0.0545 & 0.00 $\pm$ 0.0004 & 0.48 $\pm$ 0.0460 & 0.46 $\pm$ 0.0205 & 1.15 $\pm$ 0.0309 & 0.94 $\pm$ 0.0497 \\
truck+automobile & 0.51 $\pm$ 0.0279 & 0.00 $\pm$ 0.0009 & 0.52 $\pm$ 0.0356 & 0.45 $\pm$ 0.0143 & 1.06 $\pm$ 0.0331 & 0.86 $\pm$ 0.1105 \\
\midrule
mean & 0.53 $\pm$ 0.0356 & 0.00 $\pm$ 0.0023 & 0.51 $\pm$ 0.0332 & 0.50 $\pm$ 0.0414 & 1.10 $\pm$ 0.0998 & 0.92 $\pm$ 0.0424 \\
\bottomrule
\end{tabular}
\end{adjustbox}
\end{table}

\clearpage
\section{Random Collection of CEs} \label{appx:sec:full_qualitative_results}
\begin{figure}[h]
                    \centering
                    
                    \begin{subfigure}{0.99\textwidth}
                        \includegraphics[width=0.99\textwidth]{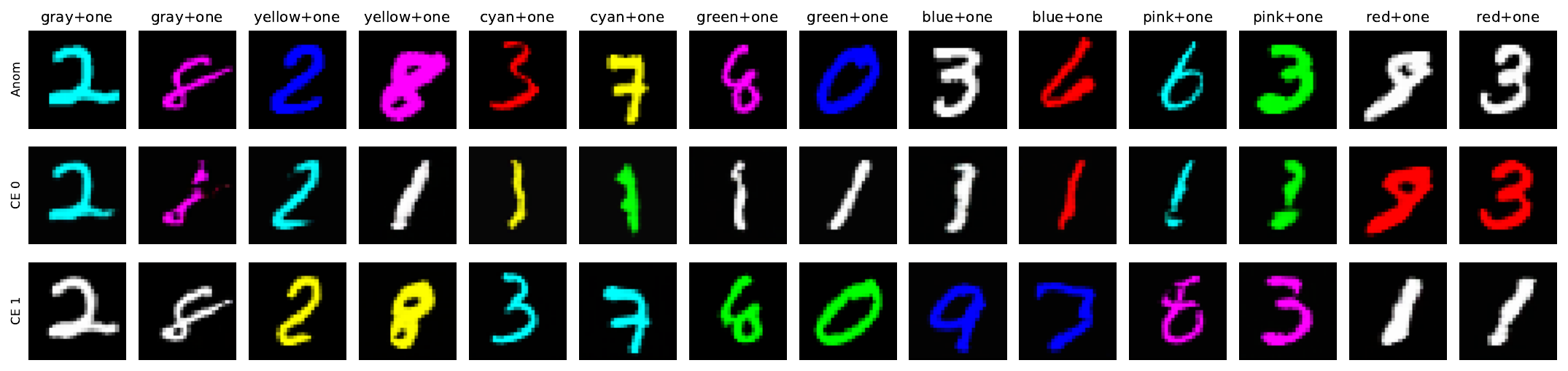}
                    \end{subfigure}
                    
                    \caption{CEs for Col-MNIST and an anomaly detector trained with BCE (OE). For each normal definition, a different detector and CE generator was trained. In each subfigure, the first row shows anomalies, the other two corresponding counterfactuals for two different concepts. Each column is labeled with the corresponding combined normal class at the top.}
                    \label{fig:appx_coloredmnist_bce_combined}
                    \end{figure}

\begin{figure}[h]
                    \centering
                    
                    \begin{subfigure}{0.99\textwidth}
                        \includegraphics[width=0.99\textwidth]{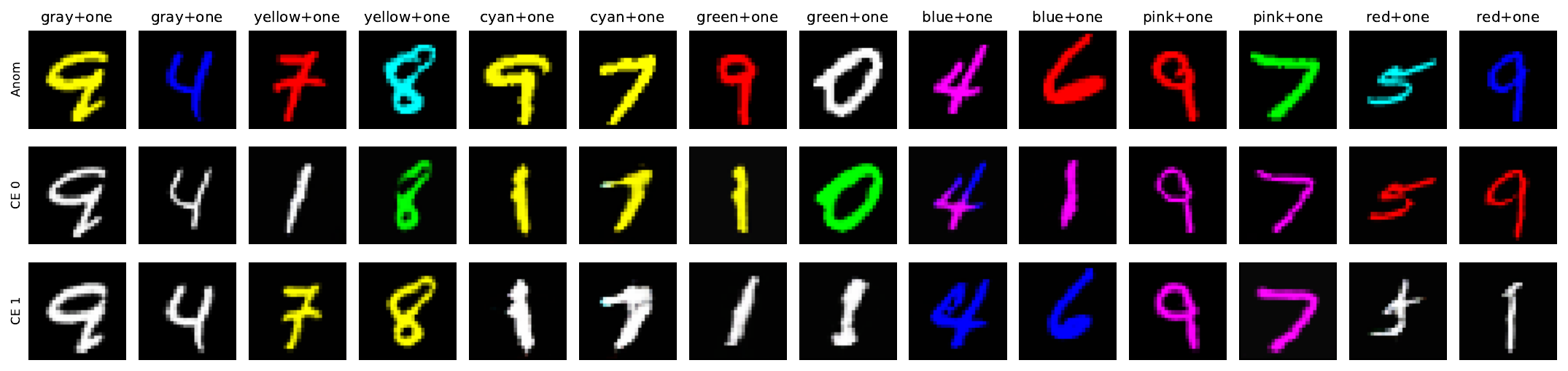}
                    \end{subfigure}
                    
                    \caption{CEs for Col-MNIST and an anomaly detector trained with HSC (OE). For each normal definition, a different detector and CE generator was trained. In each subfigure, the first row shows anomalies, the other two corresponding counterfactuals for two different concepts. Each column is labeled with the corresponding combined normal class at the top.}
                    \label{fig:appx_coloredmnist_hsc_combined}
                    \end{figure}

\begin{figure}[h!]
\centering

\begin{subfigure}{0.99\textwidth}
    \includegraphics[width=0.99\textwidth]{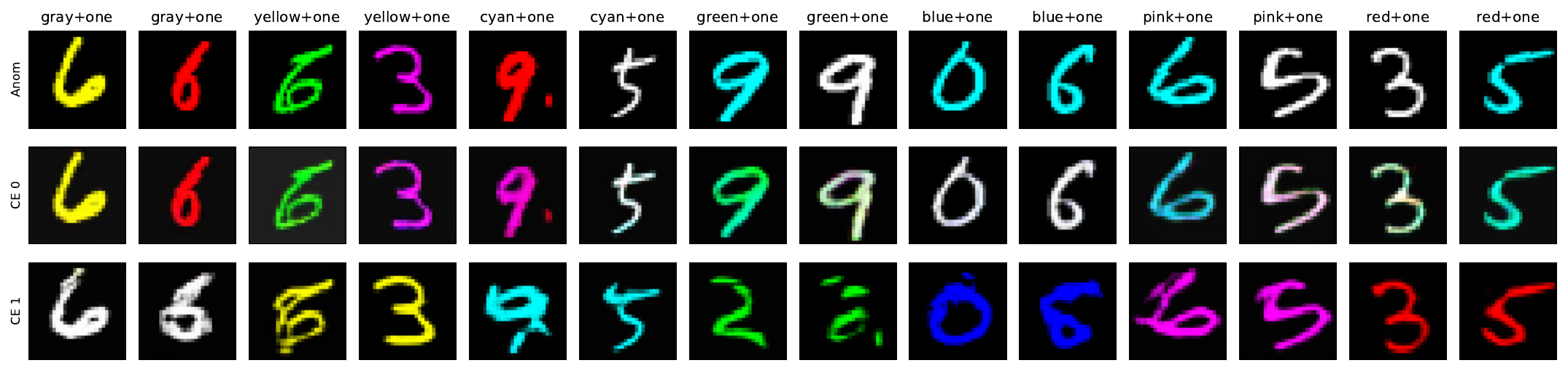}
\end{subfigure}

\caption{CEs for Col-MNIST and an anomaly detector trained with DSVDD. For each normal definition, a different detector and counterfactual generator was trained. In each subfigure, the first row shows anomalies, the other two corresponding counterfactuals for two different concepts. Each column is labeled with the corresponding combined normal class at the top.}
\label{fig:appx_coloredmnist_dsvdd_combined}
\end{figure}

We demonstrated the effectiveness of the proposed CEs by showing a small fraction of the generated CEs.
Here, we show a larger collection of CEs for all normal definitions.
For each normal definition, we randomly selected two samples to serve as examples. 
Figures \ref{fig:appx_coloredmnist_bce_combined}, \ref{fig:appx_coloredmnist_hsc_combined}, and \ref{fig:appx_coloredmnist_dsvdd_combined} show CEs for Colored-MNIST (Col-MNIST) and an AD trained with BCE, HSC, and DSVDD, respectively.  

\vspace{-0.5em}

\vspace{-2em}

\vspace{-2em}

\begin{figure}[h]
                    \centering
                    
                    \begin{subfigure}{0.99\textwidth}
                        \includegraphics[width=0.99\textwidth]{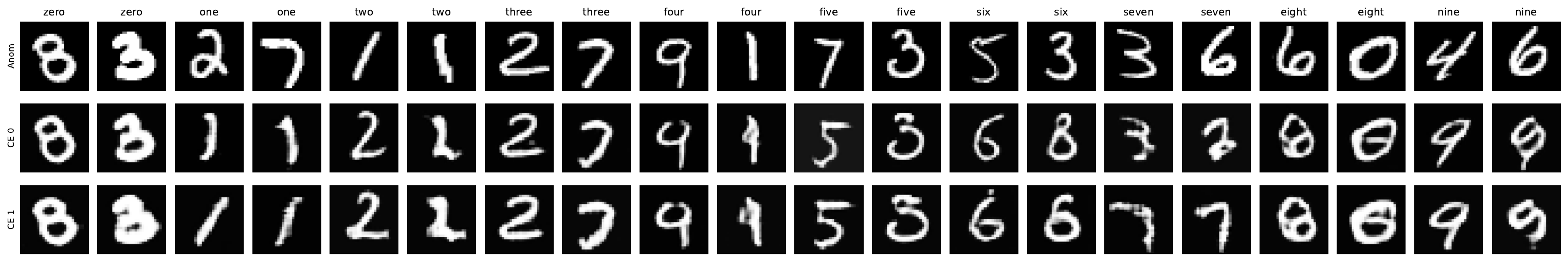}
                    \end{subfigure}
                    
                    \caption{
                    CEs for MNIST, diverse single normal classes, and an anomaly detector trained with BCE (OE). 
                    For each normal definition, a different detector and counterfactual generator was trained. 
                    In each subfigure, the first row shows anomalies, the other two corresponding counterfactuals for two different concepts. 
                    Each column is labeled with the corresponding single normal class at the top.
                    }
                    \label{fig:appx_mnist_bce_single}
                    \end{figure}

\begin{figure}[h]
                    \centering
                    
                    \begin{subfigure}{0.99\textwidth}
                        \includegraphics[width=0.99\textwidth]{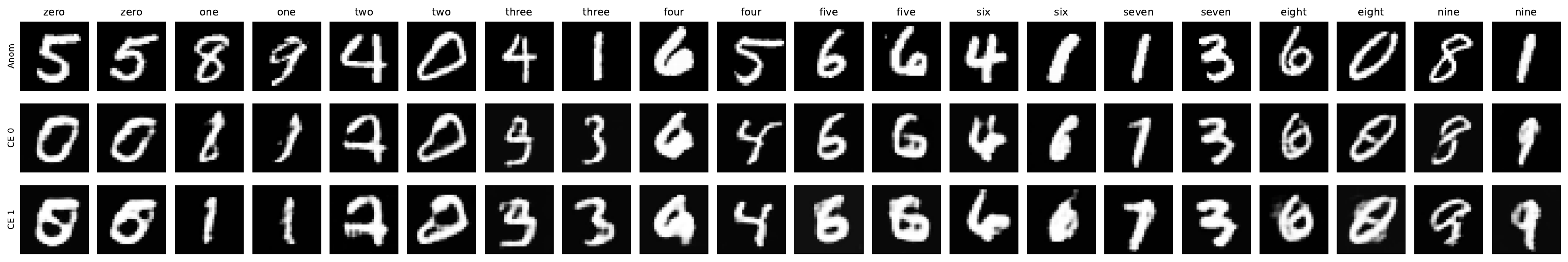}
                    \end{subfigure}
                    
                    \caption{CEs for MNIST, diverse single normal classes, and an anomaly detector trained with HSC (OE). For each normal definition, a different detector and counterfactual generator was trained. In each subfigure, the first row shows anomalies, the other two corresponding counterfactuals for two different concepts. Each column is labeled with the corresponding single normal class at the top.}
                    \label{fig:appx_mnist_hsc_single}
                    \end{figure}

\begin{figure}[h]
                    \centering
                    
                    \begin{subfigure}{0.99\textwidth}
                        \includegraphics[width=0.99\textwidth]{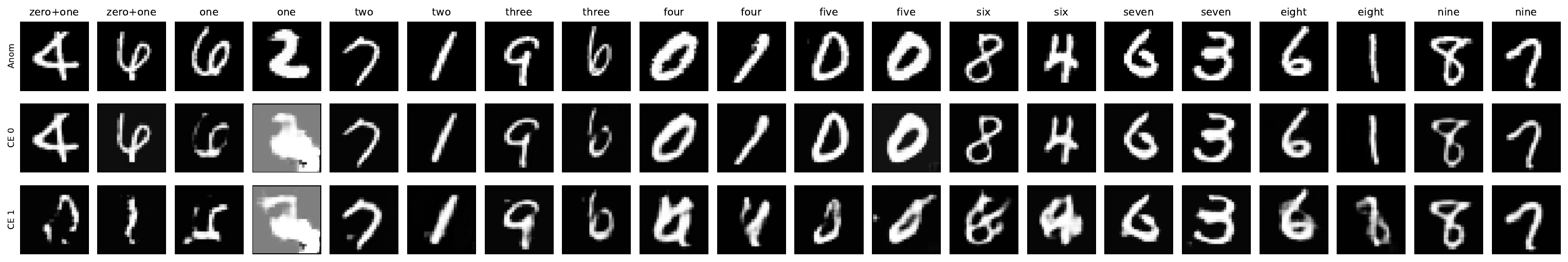}
                    \end{subfigure}
                    
                    \caption{CEs for MNIST, diverse single normal classes, and an anomaly detector trained with DSVDD. For each normal definition, a different detector and counterfactual generator was trained. In each subfigure, the first row shows anomalies, the other two corresponding counterfactuals for two different concepts. Each column is labeled with the corresponding single normal class at the top.}
                    \label{fig:appx_mnist_dsvdd_single}
                    \end{figure}

\clearpage
Figures \ref{fig:appx_mnist_bce_single}, \ref{fig:appx_mnist_hsc_single}, and \ref{fig:appx_mnist_dsvdd_single} show CEs for MNIST, single classes being normal, and an AD trained with BCE, HSC, and DSVDD, respectively.

\begin{figure}[h]
                    \centering
                    
                    \begin{subfigure}{0.99\textwidth}
                        \includegraphics[width=0.99\textwidth]{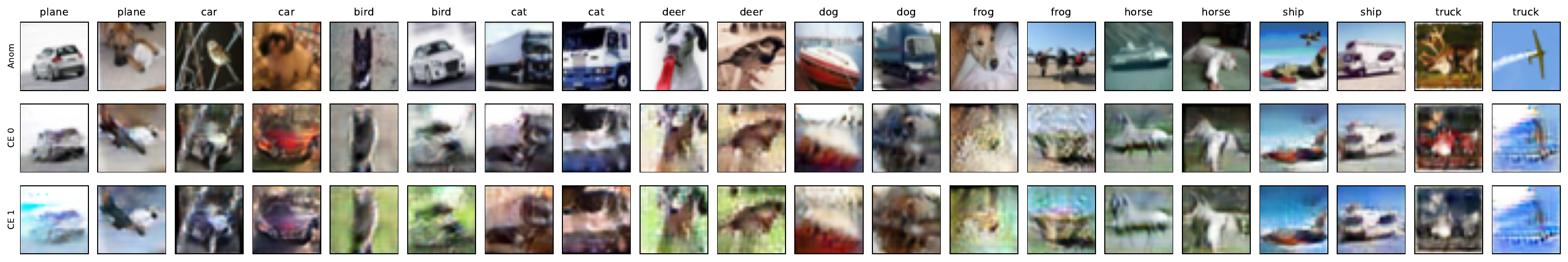}
                    \end{subfigure}
                    
                    \caption{CEs for CIFAR-10, diverse single normal classes, and an anomaly detector trained with BCE (OE). For each normal definition, a different detector and counterfactual generator was trained. In each subfigure, the first row shows anomalies, the other two corresponding counterfactuals for two different concepts. Each column is labeled with the corresponding single normal class at the top.}
                    \label{fig:appx_cifar10_bce_single}
                    \end{figure}

\begin{figure}[h]
                    \centering
                    
                    \begin{subfigure}{0.99\textwidth}
                        \includegraphics[width=0.99\textwidth]{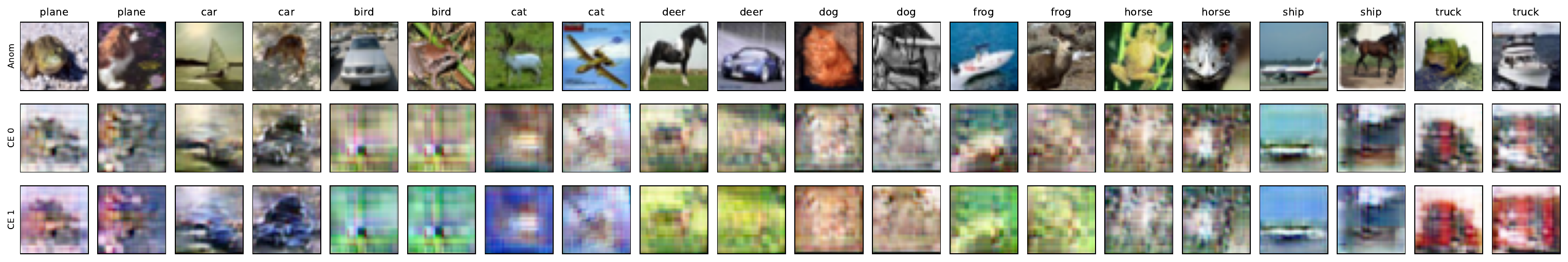}
                    \end{subfigure}
                    
                    \caption{CEs for CIFAR-10, diverse single normal classes, and an anomaly detector trained with HSC (OE). For each normal definition, a different detector and counterfactual generator was trained. In each subfigure, the first row shows anomalies, the other two corresponding counterfactuals for two different concepts. Each column is labeled with the corresponding single normal class at the top.}
                    \label{fig:appx_cifar10_hsc_single}
                    \end{figure}

\begin{figure}[h]
                    \centering
                    
                    \begin{subfigure}{0.99\textwidth}
                        \includegraphics[width=0.99\textwidth]{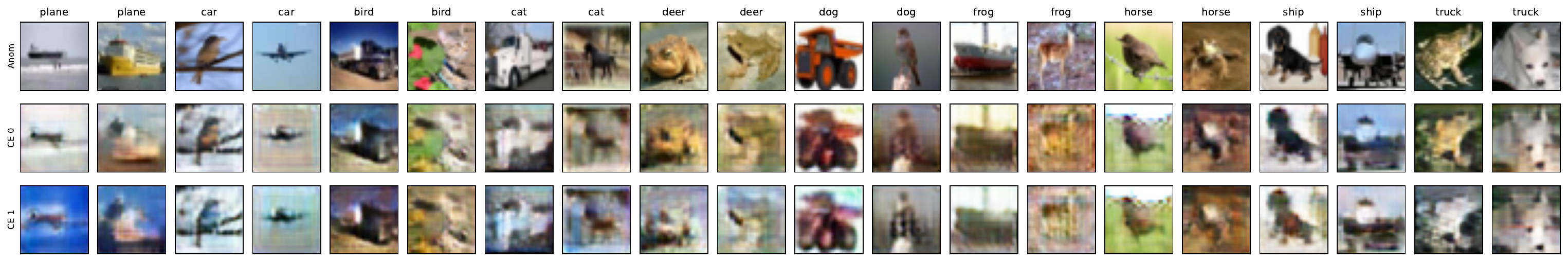}
                    \end{subfigure}
                    
                    \caption{CEs for CIFAR-10, diverse single normal classes, and an anomaly detector trained with DSVDD. For each normal definition, a different detector and counterfactual generator was trained. In each subfigure, the first row shows anomalies, the other two corresponding counterfactuals for two different concepts. Each column is labeled with the corresponding single normal class at the top.}
                    \label{fig:appx_cifar10_dsvdd_single}
                    \end{figure}

\clearpage
Figures \ref{fig:appx_cifar10_bce_single}, \ref{fig:appx_cifar10_hsc_single}, and \ref{fig:appx_cifar10_dsvdd_single} show CEs for CIFAR-10, single classes being normal, and an AD trained with BCE, HSC, and DSVDD, respectively.

\begin{figure}[h]
                    \centering
                    
                    \begin{subfigure}{0.99\textwidth}
                        \includegraphics[width=0.99\textwidth]{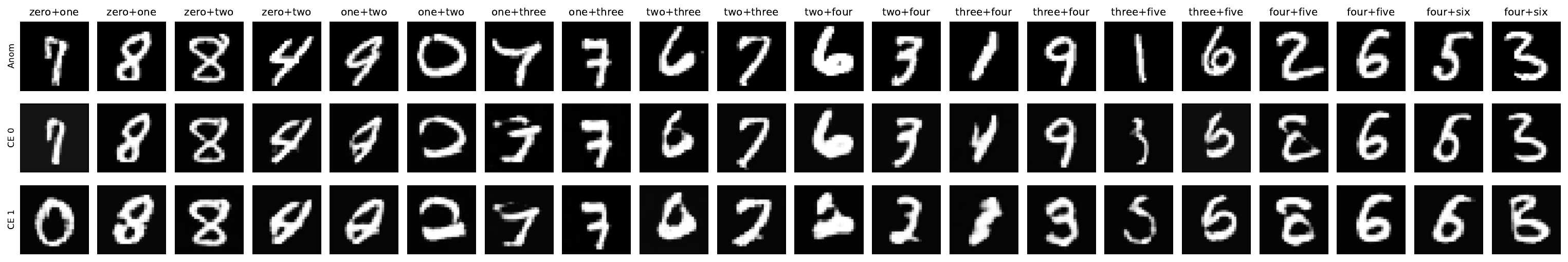}
                    \end{subfigure}

                    \begin{subfigure}{0.99\textwidth}
                        \includegraphics[width=0.99\textwidth]{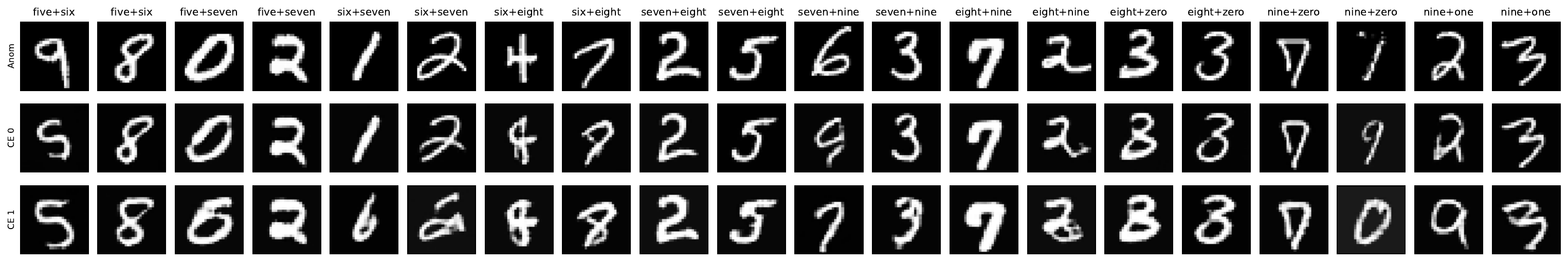}
                    \end{subfigure}
                    
                    \caption{
                    CEs for MNIST, diverse combined normal classes, and an anomaly detector trained with BCE (OE). For each normal definition, a different detector and counterfactual generator was trained. 
                    In each subfigure, the first row shows anomalies, the other two corresponding counterfactuals for two different concepts. 
                    Each column is labeled with the corresponding combined normal class at the top.
                    }
                    \label{fig:appx_mnist_bce_combined}
                    \end{figure}

\begin{figure}[h]
                    \centering
                    
                    \begin{subfigure}{0.99\textwidth}
                        \includegraphics[width=0.99\textwidth]{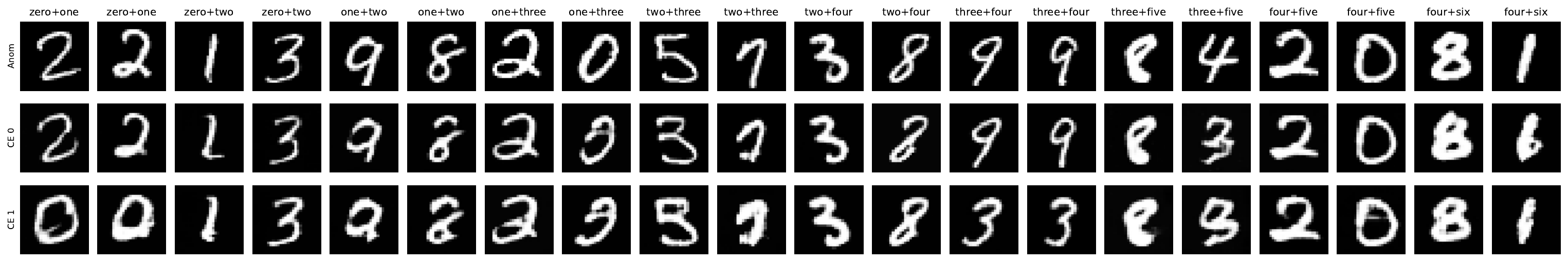}
                    \end{subfigure}

                    \begin{subfigure}{0.99\textwidth}
                        \includegraphics[width=0.99\textwidth]{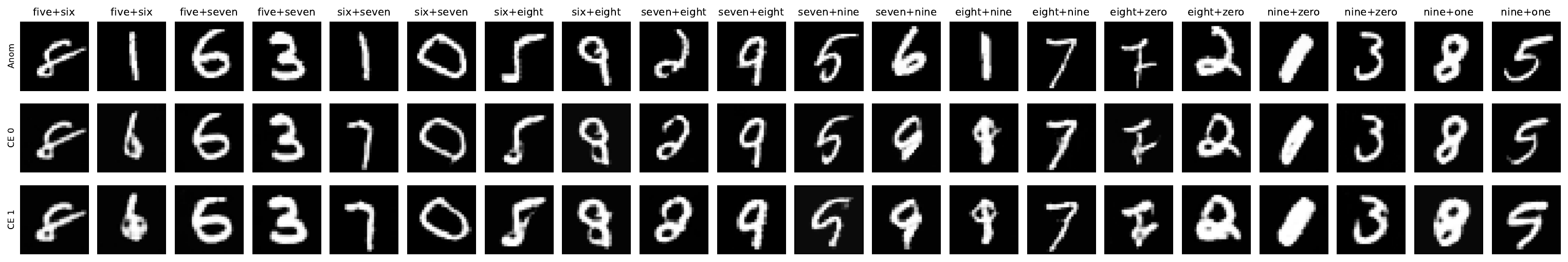}
                    \end{subfigure}
                    
                    \caption{
                    CEs for MNIST, diverse combined normal classes, and an anomaly detector trained with HSC (OE). 
                    For each normal definition, a different detector and counterfactual generator was trained. 
                    In each subfigure, the first row shows anomalies, the other two corresponding counterfactuals for two different concepts. 
                    Each column is labeled with the corresponding combined normal class at the top.
                    }
                    \label{fig:appx_mnist_hsc_combined}
                    \end{figure}

\begin{figure}[h!]
                    \centering
                    
                    \begin{subfigure}{0.99\textwidth}
                        \includegraphics[width=0.99\textwidth]{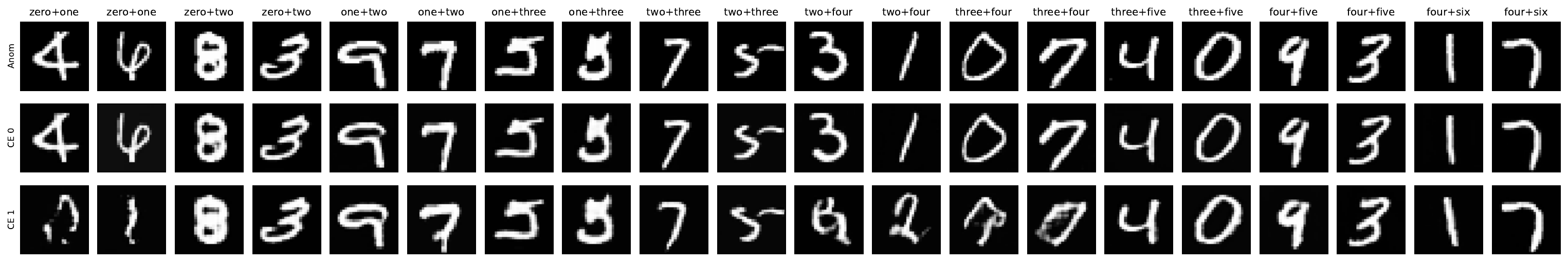}
                    \end{subfigure}

                    \begin{subfigure}{0.99\textwidth}
                        \includegraphics[width=0.99\textwidth]{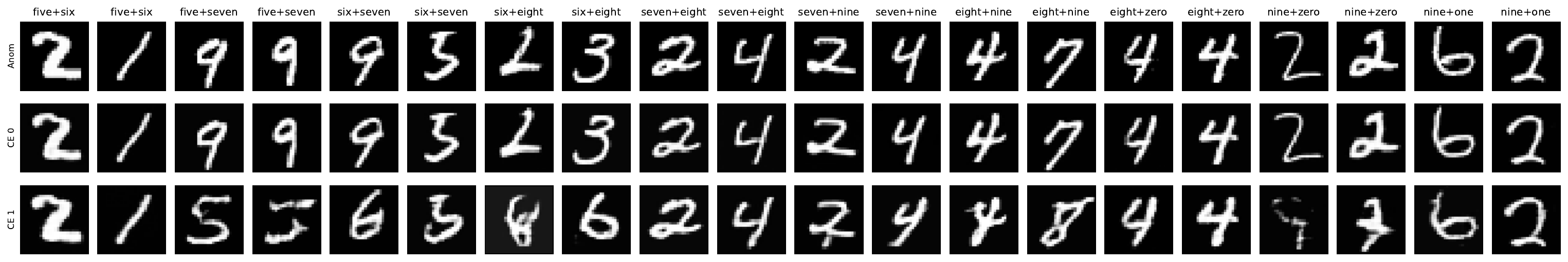}
                    \end{subfigure}
                    
                    \caption{
                    CEs for MNIST, diverse combined normal classes, and an anomaly detector trained with DSVDD. For each normal definition, a different detector and counterfactual generator was trained. 
                    In each subfigure, the first row shows anomalies, the other two corresponding counterfactuals for two different concepts. 
                    Each column is labeled with the corresponding combined normal class at the top.
                    }
                    \label{fig:appx_mnist_dsvdd_combined}
                    \end{figure}

\begin{figure}[h]
                    \centering
                    
                    \begin{subfigure}{0.99\textwidth}
                        \includegraphics[width=0.99\textwidth]{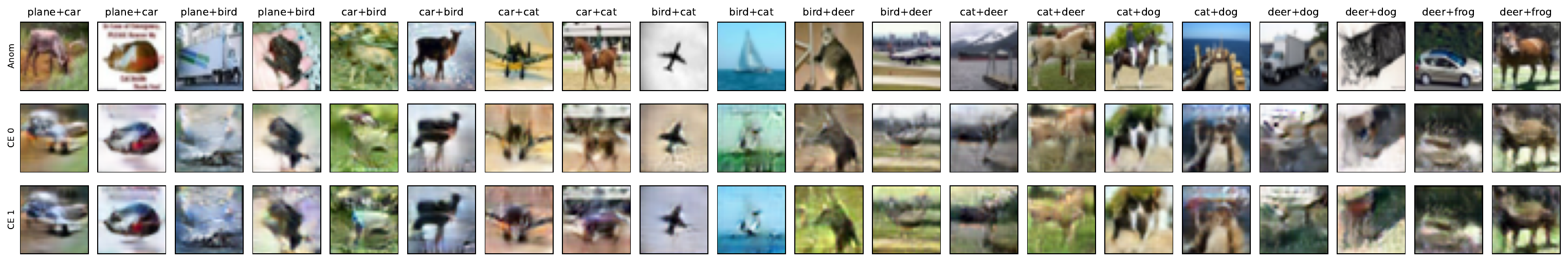}
                    \end{subfigure}

                    \begin{subfigure}{0.99\textwidth}
                        \includegraphics[width=0.99\textwidth]{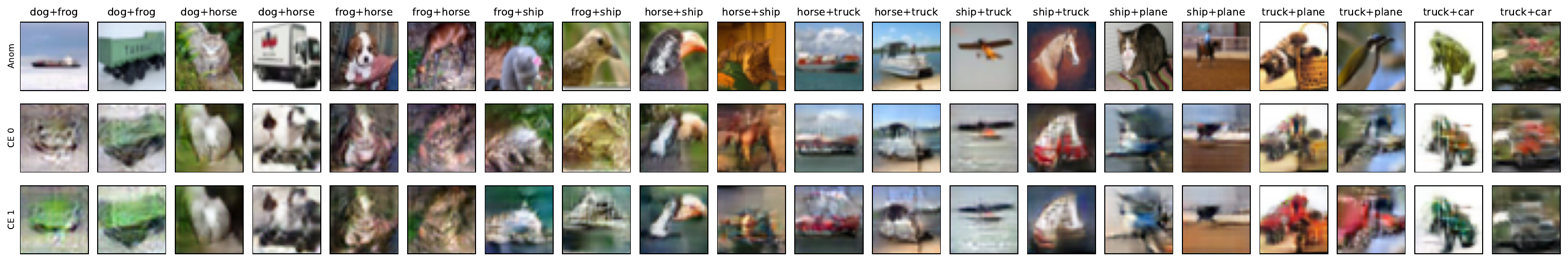}
                    \end{subfigure}
                    
                    \caption{CEs for CIFAR-10, diverse combined normal classes, and an anomaly detector trained with BCE (OE). For each normal definition, a different detector and generator was trained. In each subfigure, the first row shows anomalies, the other two corresponding counterfactuals for two different concepts. Each column is labeled with the corresponding combined normal class at the top.}
                    \label{fig:appx_cifar10_bce_combined}
                    \end{figure}

\begin{figure}[h]
                    \centering
                    
                    \begin{subfigure}{0.99\textwidth}
                        \includegraphics[width=0.99\textwidth]{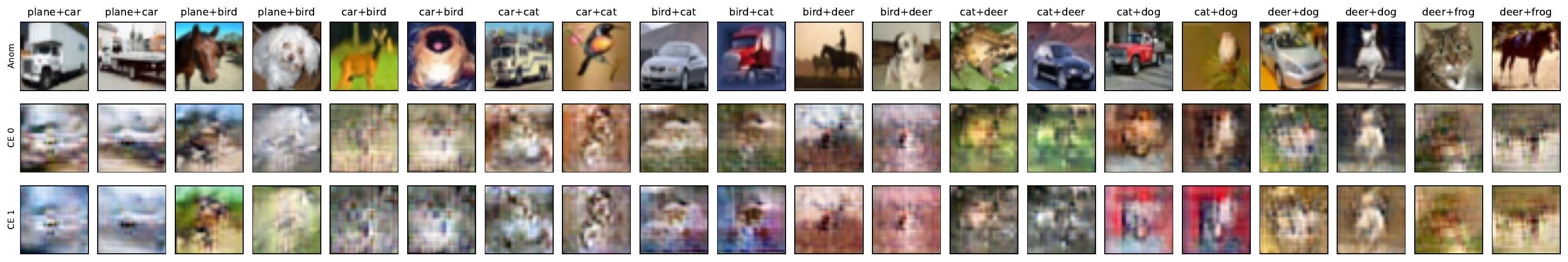}
                    \end{subfigure}

                    \begin{subfigure}{0.99\textwidth}
                        \includegraphics[width=0.99\textwidth]{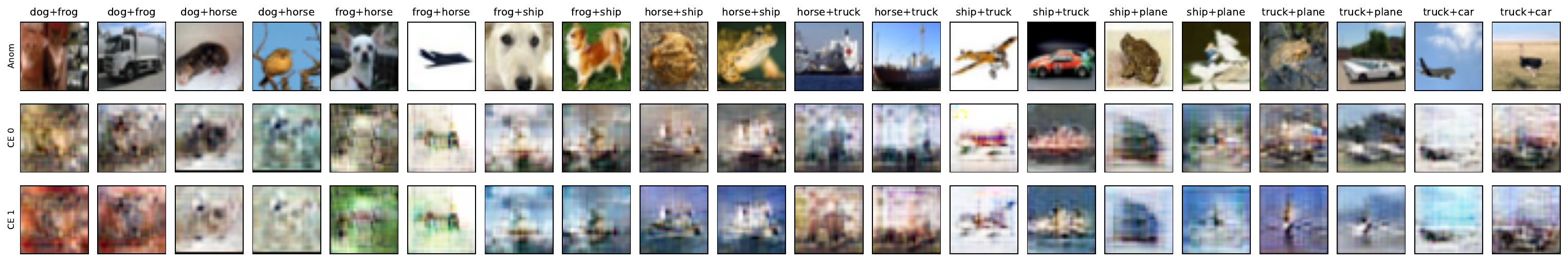}
                    \end{subfigure}
                    
                    \caption{CEs for CIFAR-10, diverse combined normal classes, and an anomaly detector trained with HSC (OE). For each normal definition, a different detector and generator was trained. In each subfigure, the first row shows anomalies, the other two corresponding counterfactuals for two different concepts. Each column is labeled with the corresponding combined normal class at the top.}
                    \label{fig:appx_cifar10_hsc_combined}
                    \end{figure}

\begin{figure}[h!]
                    \centering
                    
                    \begin{subfigure}{0.97\textwidth}
                        \includegraphics[width=0.99\textwidth]{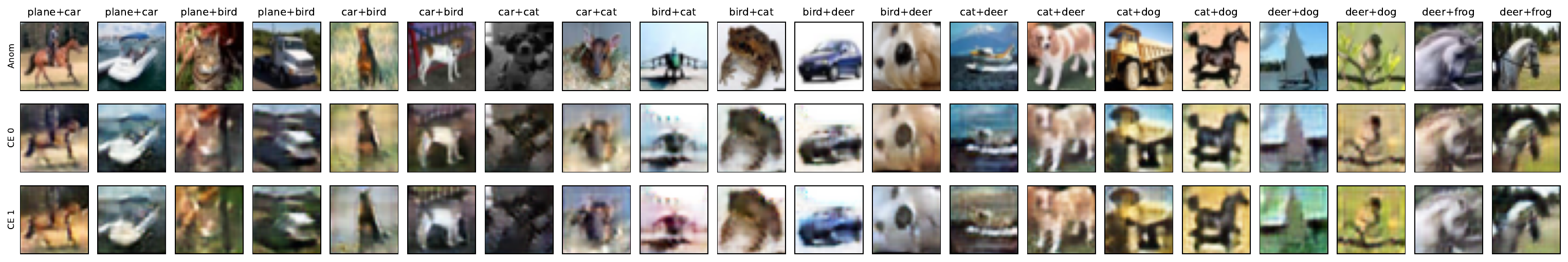}
                    \end{subfigure}

                    \begin{subfigure}{0.97\textwidth}
                        \includegraphics[width=0.99\textwidth]{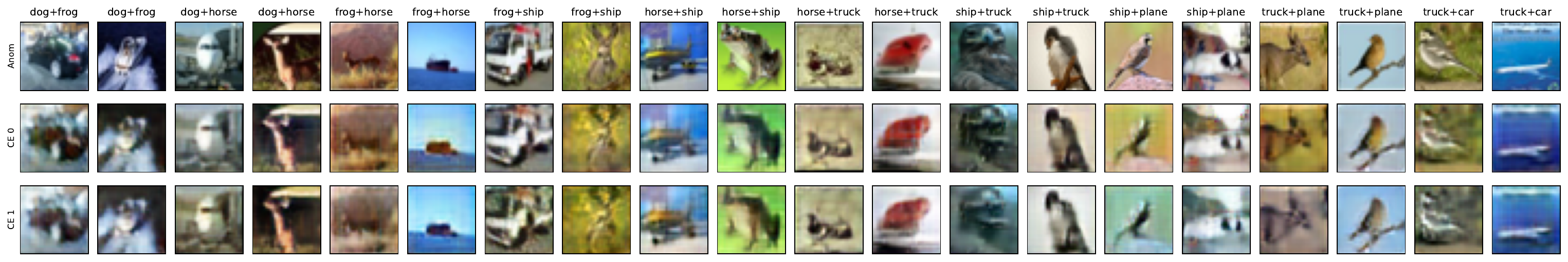}
                    \end{subfigure}
                    
                    \caption{CEs for CIFAR-10, diverse combined normal classes, and an anomaly detector trained with DSVDD. For each normal definition, a different detector and generator was trained. In each subfigure, the first row shows anomalies, the other two corresponding counterfactuals for two different concepts. Each column is labeled with the corresponding combined normal class at the top.}
                    \label{fig:appx_cifar10_dsvdd_combined}
                    \end{figure}

\clearpage
Figures \ref{fig:appx_mnist_bce_combined}, \ref{fig:appx_mnist_hsc_combined}, and \ref{fig:appx_mnist_dsvdd_combined} show CEs for MNIST, class combinations being normal, and an AD trained with BCE, HSC, and DSVDD, respectively.  
Figures \ref{fig:appx_cifar10_bce_combined}, \ref{fig:appx_cifar10_hsc_combined}, and \ref{fig:appx_cifar10_dsvdd_combined} show CEs for CIFAR-10, class combinations being normal, and an AD trained with BCE, HSC, and DSVDD, respectively.

\begin{figure}[h]
\centering

\includegraphics[width=0.97\textwidth]{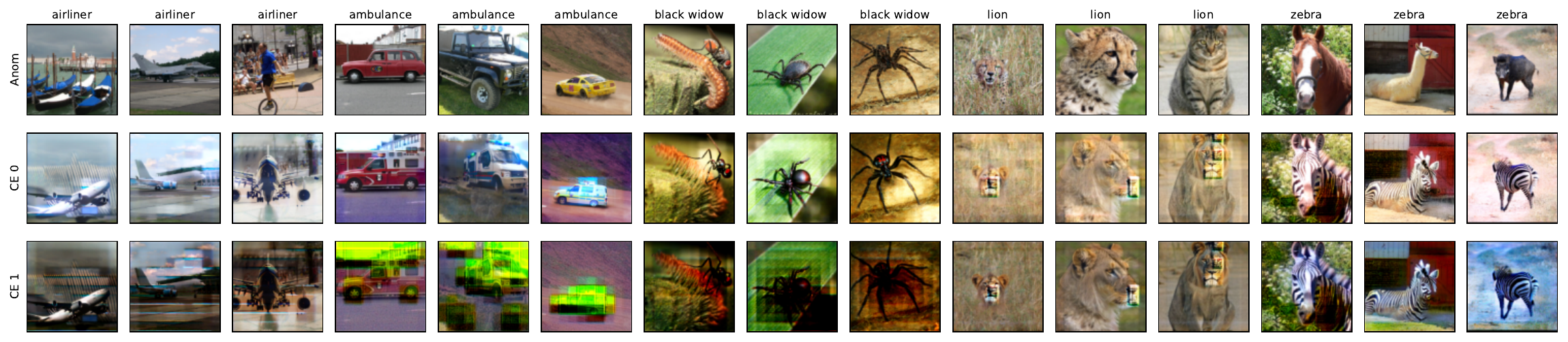}

\caption{CEs for ImageNet-Neighbors, single normal classes, and an anomaly detector trained with BCE (OE). For each normal definition, a different detector and counterfactual generator was trained. In each subfigure, the first row shows anomalies, the other two corresponding counterfactuals for two different concepts. Each column is labeled with the corresponding normal class at the top.}
\label{fig:appx_imagenet_bce_single}
\end{figure}

\begin{figure}[h]
\centering

\includegraphics[width=0.97\textwidth]{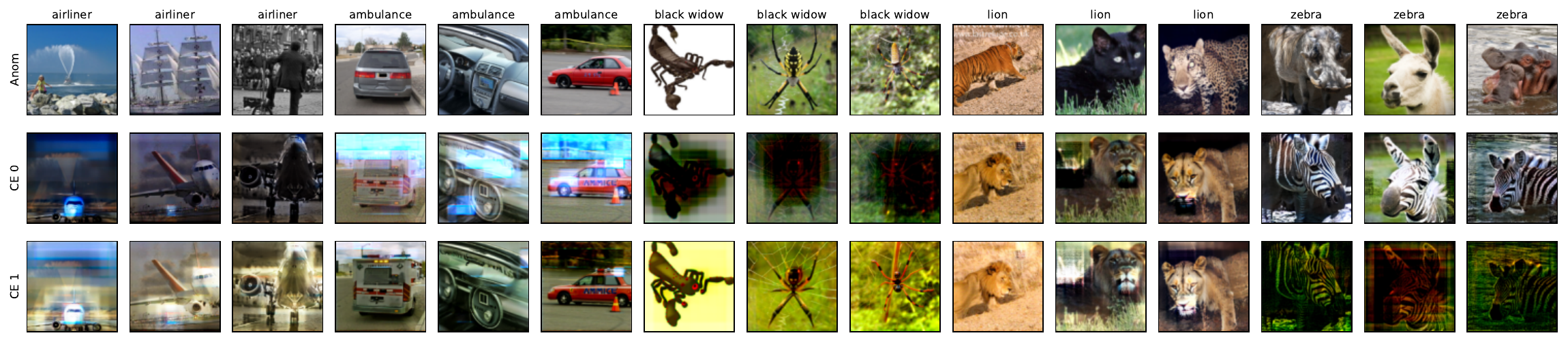}

\caption{CEs for ImageNet-Neighbors, single normal classes, and an anomaly detector trained with HSC (OE). For each normal definition, a different detector and counterfactual generator was trained. In each subfigure, the first row shows anomalies, the other two corresponding counterfactuals for two different concepts. Each column is labeled with the corresponding normal class at the top.}
\label{fig:appx_imagenet_hsc_single}
\end{figure}

\clearpage
Figures \ref{fig:appx_imagenet_bce_single} and \ref{fig:appx_imagenet_hsc_single} show the CEs for ImageNet-Neighbors, with single classes being normal, and an AD trained with BCE and HSC, respectively.
                    
\begin{figure}[h]
                \centering

                \begin{subfigure}{0.99\textwidth}
                    \includegraphics[width=0.99\textwidth]{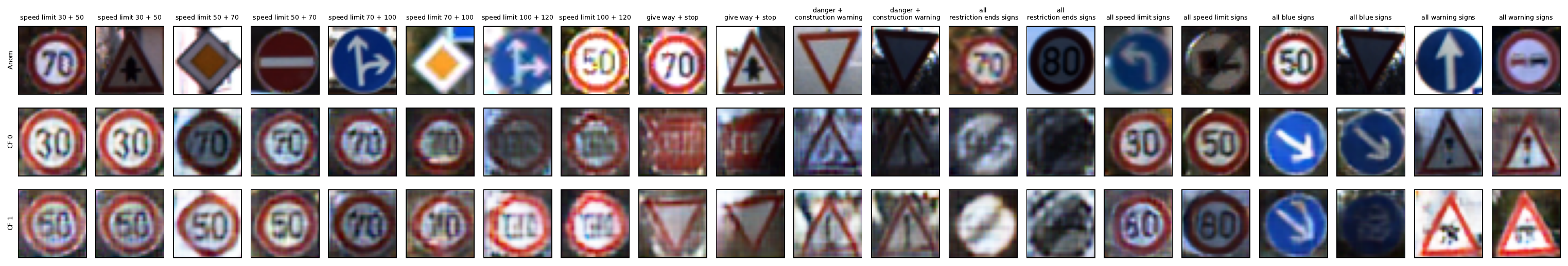}
                \end{subfigure}

                \caption{CEs for GTSDB and an anomaly detector trained with BCE OE. For each normal definition, a different detector and counterfactual generator was trained. In each subfigure, the first row shows anomalies, the other two corresponding counterfactuals for two different concepts. Each column is labeled with the corresponding combined normal class at the top.}
                \label{fig:appx_gtsdb_bce_combined}
                \end{figure}

\begin{figure}[h]
                \centering

                \begin{subfigure}{0.99\textwidth}
                    \includegraphics[width=0.99\textwidth]{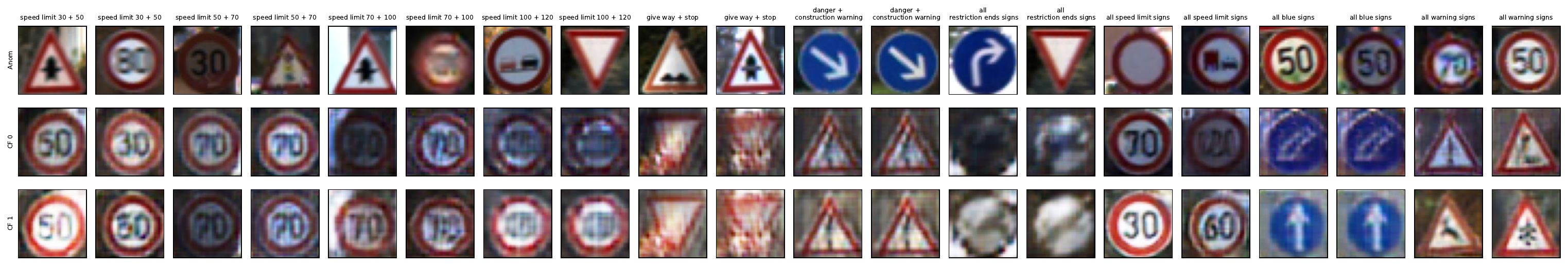}
                \end{subfigure}

                \caption{CEs for GTSDB and an anomaly detector trained with HSC OE. For each normal definition, a different detector and counterfactual generator was trained. In each subfigure, the first row shows anomalies, the other two corresponding counterfactuals for two different concepts. Each column is labeled with the corresponding combined normal class at the top.}
                \label{fig:appx_gtsdb_hsc_combined}
                \end{figure}

\begin{figure}[h!]
                \centering

                \begin{subfigure}{0.99\textwidth}
                    \includegraphics[width=0.99\textwidth]{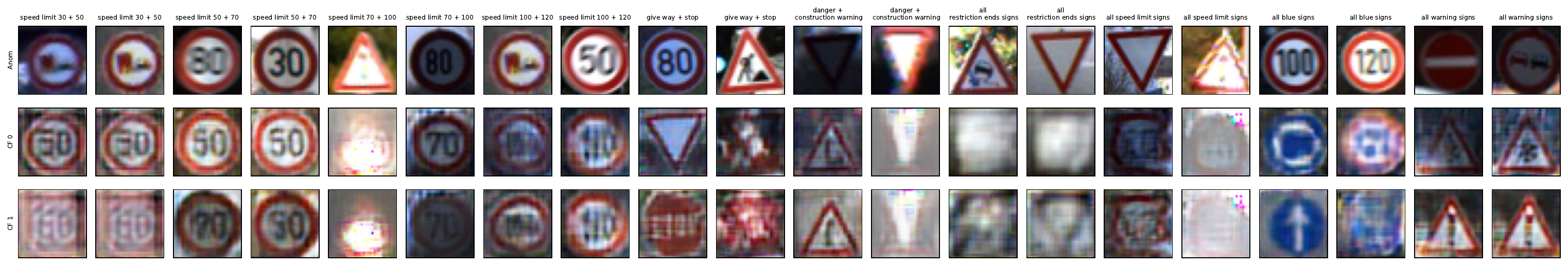}
                \end{subfigure}

                \caption{CEs for GTSDB and an anomaly detector trained with DSVDD. For each normal definition, a different detector and counterfactual generator was trained. In each subfigure, the first row shows anomalies, the other two corresponding counterfactuals for two different concepts. Each column is labeled with the corresponding combined normal class at the top.}
                \label{fig:appx_gtsdb_dsvdd_combined}
                \end{figure}

\clearpage
Figures \ref{fig:appx_gtsdb_bce_combined}, \ref{fig:appx_gtsdb_hsc_combined}, and \ref{fig:appx_gtsdb_dsvdd_combined} show CEs for GTSDB, class combinations being normal, and an AD trained with BCE, HSC, and DSVDD, respectively.

\vspace{-0em}

\vspace{-0em}

\end{document}